\documentclass{article}

\usepackage{longcat_style}
\usepackage{adjustbox}
\usepackage[utf8]{inputenc} %
\usepackage[T1]{fontenc}    %
\usepackage{newunicodechar}
\usepackage{hyperref}       %
\usepackage{xcolor}
\usepackage[normalem]{ulem} %
\hypersetup{
    colorlinks=true,      %
    linkcolor=blue,      %
    urlcolor=blue,       %
    citecolor=blue,      %
    linkbordercolor=blue, %
    urlbordercolor=blue,
    citebordercolor=blue,
    pdfborderstyle={/S/U/W 1}, %
}

\usepackage{graphicx}
\usepackage{booktabs}
\usepackage{caption} 

\usepackage{float}
\usepackage{url}            %
\usepackage{booktabs}       %
\usepackage{amsfonts}       %
\usepackage{nicefrac}       %
\usepackage{microtype}      %
\usepackage{lipsum}		%
\usepackage{graphicx}

\usepackage{CJKutf8}

\usepackage{multicol}
\usepackage[style=numeric-comp, backend=biber]{biblatex}
\usepackage{amsmath}
\usepackage{amssymb} %
\usepackage{xspace}
\usepackage{enumerate}
\usepackage{enumitem}
\usepackage{setspace}
\usepackage{multirow}
\usepackage{subcaption} 
\usepackage{makecell}
\usepackage{hyperref, cleveref}
\usepackage{pifont}
\usepackage[inkscapelatex=false]{svg}
\usepackage{subcaption}
\usepackage{wrapfig}
\usepackage[most]{tcolorbox} 
\usepackage{tabularx}
\usepackage{makecell}
\usepackage[table]{xcolor}
\newcommand{\cmark}{\ding{51}}
\newcommand{\xmark}{\ding{55}}

\newtcolorbox{summarybox}{
    colback=blue!5!white,    
    colframe=black!75,       
    arc=6pt,                 
    boxrule=1pt,             
    left=8pt,                
    right=8pt,               
    top=8pt,                 
    bottom=8pt,              
    enhanced,                
}

\newcommand{\longcat}{LongCat-Next\xspace}

\usepackage{enumitem}
\setlist{
    topsep=2pt,        
    itemsep=2pt,       
    parsep=0pt,        
    partopsep=0pt      
}

\usepackage[skip=4pt]{caption}
\usepackage{enumitem}
\setlist{topsep=1pt, itemsep=2pt, parsep=0pt, partopsep=0pt}

\usepackage{titlesec}
\titlespacing*{\section}{0pt}{10pt}{6pt}
\titlespacing*{\subsection}{0pt}{8pt}{4pt}
\titlespacing*{\subsubsection}{0pt}{6pt}{3pt}

\definecolor{midnightgreen}{rgb}{0.0, 0.29, 0.33}

\title{LongCat-Next: Lexicalizing Modalities as \\ Discrete Tokens}

\author{
\textbf{Meituan LongCat Team}
}

\renewcommand{\headeright}{\raisebox{-0.2\height}{\includegraphics[height=1.8em]{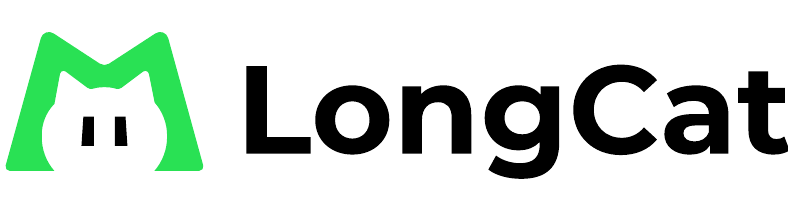}}}
\renewcommand{\shorttitle}{LongCat-Next: Lexicalizing Modalities as Discrete Tokens}

\begin{document}
\maketitle

\begin{abstract}

The prevailing Next-Token Prediction (NTP) paradigm has driven the success of large language models through discrete autoregressive modeling. However, contemporary multimodal systems remain language-centric, often treating non-linguistic modalities as external attachments, leading to fragmented architectures and suboptimal integration.
To transcend this limitation, we introduce Discrete Native Autoregressive (DiNA), a unified framework that represents multimodal information within a shared discrete space, enabling a consistent and principled autoregressive modeling across modalities.
A key innovation is the Discrete Native Any-resolution Visual Transformer (dNaViT), which performs tokenization and de-tokenization at arbitrary resolutions, transforming continuous visual signals into hierarchical discrete tokens. Building on this foundation, we develop LongCat-Next, a native multimodal model that processes text, vision, and audio under a single autoregressive objective with minimal modality-specific design.
As an industrial-strength foundation model, it excels at seeing, painting, and talking within a single framework, achieving strong performance across a wide range of multimodal benchmarks. 
In particular, LongCat-Next addresses the long-standing performance ceiling of discrete vision modeling on understanding tasks and provides a unified approach to effectively reconcile the conflict between understanding and generation.
Extensive experiments demonstrate that discrete tokens can universally represent multimodal signals and be deeply internalized within a single embedding space, offering interesting insights into this unified training paradigm.
As an attempt toward native multimodality, we open-source the LongCat-Next and its tokenizers, hoping to foster further research and development in the community.

\textbf{GitHub}: \href{https://github.com/meituan-longcat/LongCat-Next}{https://github.com/meituan-longcat/LongCat-Next}\\
\textbf{Hugging Face}: \href{https://huggingface.co/meituan-longcat/LongCat-Next}{https://huggingface.co/meituan-longcat/LongCat-Next}\\

\end{abstract}

\begin{figure}[h]
    \centering
    \includegraphics[width=0.76\linewidth]{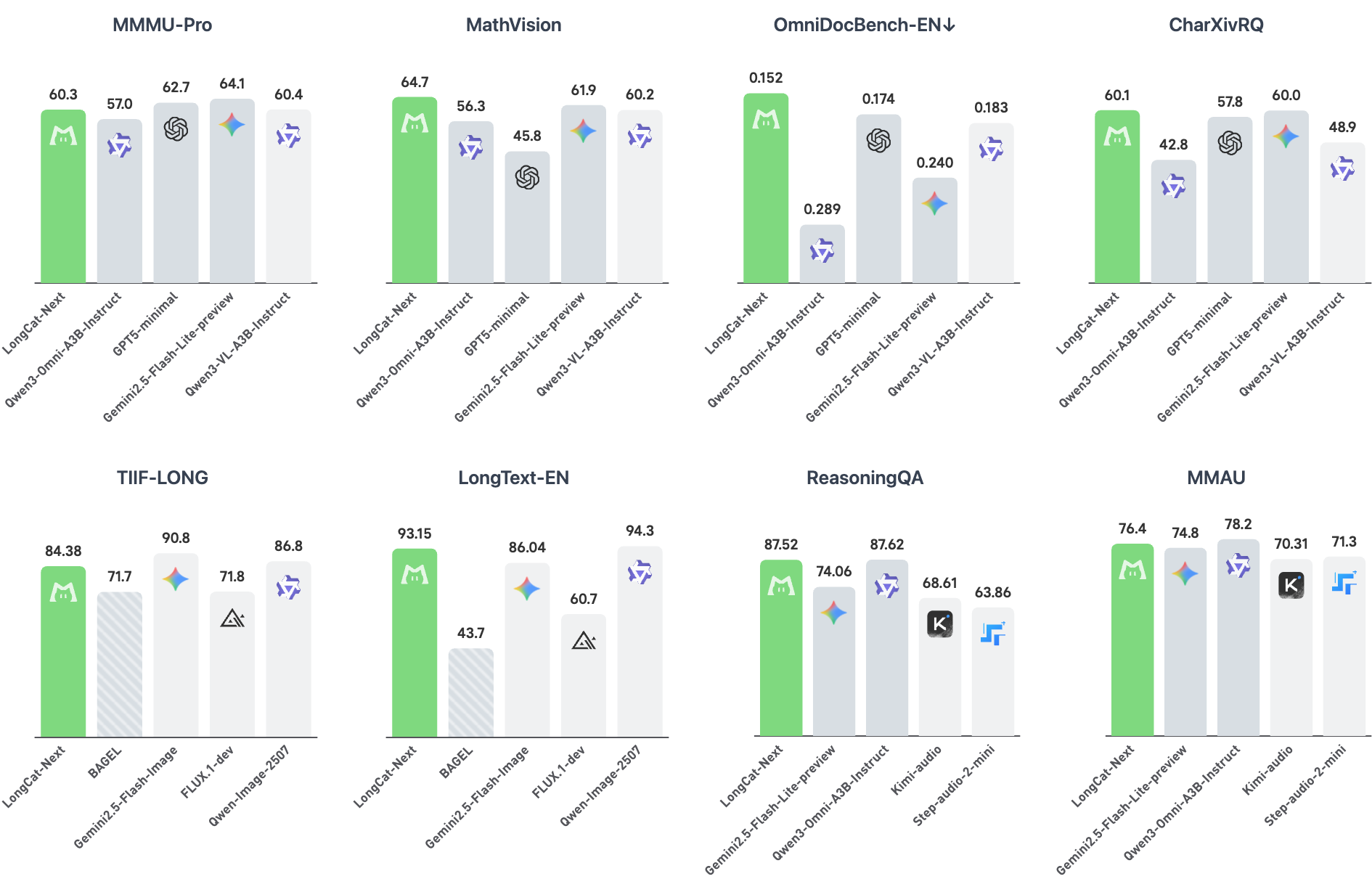}
    \caption{Benchmark performance of LongCat-Next.}
    \label{fig:benchmark}
\end{figure}

\clearpage
\tableofcontents
\clearpage

\section{Introduction}

Large Language Models (LLMs) have converged on the Next Token Prediction (NTP) paradigm \cite{gpt4o,comanici2025gemini25pushingfrontier,yang2025qwen3,meituanlongcatteam2025longcatflashtechnicalreport,team2026kimi,mimoflash}, where intelligence emerges from large-scale discrete autoregressive modeling~\cite{gpt2,gpt3}.
However, language captures only a limited portion of the rich perceptual information in the real world, which inherently spans multiple modalities, \textit{e.g,} text, vision, and audio.
Despite this, most prevailing multimodal systems still treat non-linguistic modalities as subordinate, bolt-on components that are loosely coupled with language modeling \cite{llava,wu2025qwenimagetechnicalreport}. 
This separation holds untapped potential to move beyond prevailing \textit{language-plus-auxiliary} paradigm toward native multimodal modeling.

When multimodality is conceptualized analogously as a native linguistic extension of language, the problem simplifies considerably, where all modalities are represented as interoperable token sequences governed by a single shared autoregressive objective. Despite these conceptual advantages, the field still lacks an industrial-strength training recipe for achieving a genuinely unified multimodal model at scale. At the core lies a fundamental question: how can non-linguistic modalities be effectively represented within a discrete token space? In essence, the pursuit of tokenizing all modalities into a universal interface lies at the heart of multimodal modeling~\cite{wang2024emu3,team2024chameleon}.

Since language is naturally expressed through speech, discrete autoregressive modeling has achieved remarkable progress in the audio area~\cite{defossez2024moshi,li2025baichuan,zhang2025mimo}, where discrete audio tokens capture not only text-aligned semantics but also rich paralinguistic information such as emotion, tone, and environmental context. However, extending discrete autoregressive modeling to vision is conceptually straightforward yet practically nontrivial. Unlike words, which are naturally compact and discrete units, visual signals are high-dimensional and continuous. There remains widespread doubt as to whether discrete visual modeling can achieve strong performance in both comprehension and autoregressive generation, as compressing rich visual information into a \textit{finite codebook} inevitably hinders representation capacity.

To address this challenge, we identify a fundamental dual bottleneck in discrete visual modeling: (i) capacity of visual representation, and (ii) information loss from discretization. For the former, we emphasize the importance of achieving \textit{semantic completeness} and highlight that a class of Semantic-and-Aligned Encoders (SAE) serves as a strong foundation. Interestingly, we discover that the encoder's residual architecture inherently preserves a latent pathway for low-level signal propagation, even without reconstruction supervision. 
For the information bottleneck of discretization, we leverage the hierarchical nature of visual signals by modeling the \textit{residual of the residual} via Residual Vector Quantization (RVQ)~\cite{rqvae}, effectively preserving information for both understanding and generation.

Building on these insights, we introduce the Discrete Native Resolution Vision Transformer (dNaViT), a unified visual tokenizer designed to function analogously to linguistic tokenizers.
Through a carefully designed training process, dNaViT can perform paired tokenization and de-tokenization, encoding images into discrete IDs with semantic completeness for understanding, and simultaneously decoding the token sequences back into images for reconstruction and generation, both at arbitrary resolution with up to 28 $\times$ compression ratio.
By treating multi-level residual tokens as a shared representational currency, dNaViT primarily enables bidirectional mapping between images and discrete IDs.
During autoregressive modeling, we employ additive encoding over multi-layer tokens and a DepthTransformer for efficient decoding, unlocking an exponential representation space for multi-level tokens, while maintaining the computational efficiency of a single autoregressive step. This design allows vision to be discretized into a unified token space akin to language, achieving an optimal balance between representation fidelity and compression rate.

The same design principle holds in audio modeling, where we employ an architecture based on RVQ for discrete representation. Utilizing a Whisper encoder~\cite{radford2023robust} to capture both semantic and paralinguistic features, our audio tokenizer compresses waveforms into discrete tokens at 12.5 Hz. The audio detokenizer uses a paired decoder and a refinement network based on flow matching to achieve high-fidelity reconstruction. For autoregressive audio modeling, we further introduce a unified training paradigm that aligns segment-level text and audio tokens with stochastic delays, enabling both parallel and serial text-guided speech generation. This approach enhances the linguistic quality of speech generation and facilitates seamless adaptation across diverse interaction scenarios.

\begin{figure}[t]
    \centering
    \includegraphics[width=1.0\linewidth]{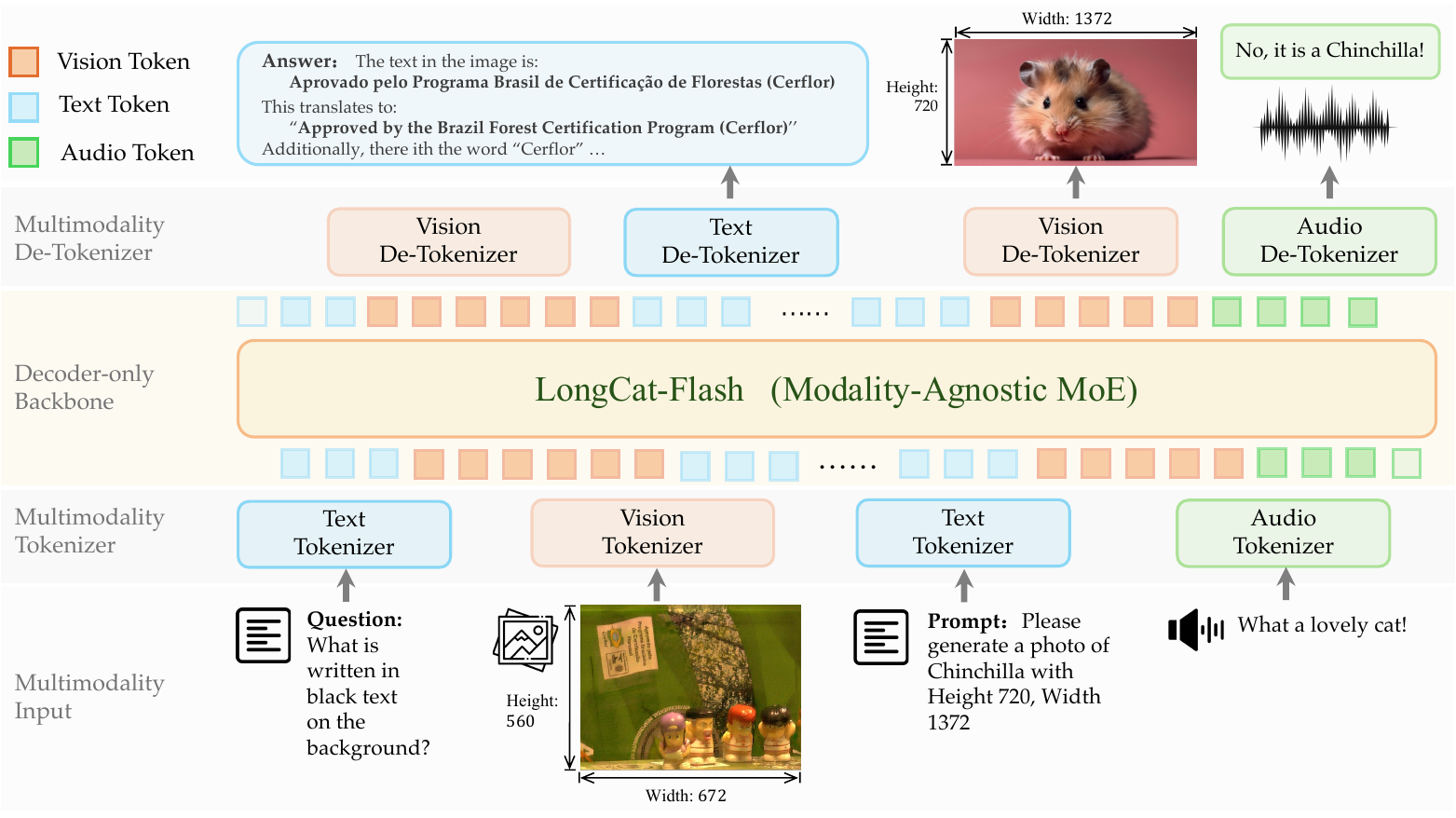}
    \caption{Overview of the LongCat-Next architecture, designed under a Discrete Native Autoregression (DiNA) paradigm that extends multimodality into a native language-style modeling framework via paired tokenizers.}
    \vspace{-4pt}
    \label{fig:overall}
\end{figure}

This work focuses on the fundamental challenge of native multimodality through a design philosophy that prioritizes simplicity, treating vision and audio as intrinsic extensions of language.
Building on Mixture-of-Experts (MoE) backbone \cite{meituanlongcatteam2025longcatflashtechnicalreport,longcat-flash-lite}, we instantiate this foundation to introduce LongCat-Next, a discrete native multimodal model that unifies language, vision, and audio within a single, cohesive framework, delivering industrial-strength performance and competitive results across diverse multimodal domains. The principal contributions of this work are listed as follows:
\begin{itemize}
\item 
\textbf{Discrete Native Autoregression Paradigm (DiNA).} 
We introduce DiNA, a unified paradigm that extends next-token prediction from language to native multimodality by representing all modalities within a shared discrete token space. By internalizing diverse modalities into this unified interface, DiNA aligns multimodal modeling with standard decoder-only architectures, enabling a single model to handle text, vision, and audio under a consistent autoregressive objective. Under this paradigm, the core challenge reduces to designing modality-specific tokenizer–detokenizer pairs, turning the model into a unified multi-task learner across modalities. This design preserves architectural simplicity while leveraging the mature training infrastructure of large language models, providing a unified multimodal foundation.
\item
\textbf{Discrete Native-Resolution Vision Transformer (dNaViT).} 
We propose dNaViT, a unified interface that represents visual inputs as discrete ``visual words'', guided by the principle of \emph{semantic completeness} to overcome the capability ceiling of discrete visual modeling. Concretely, we leverage Semantic-and-Aligned Encoders (SAE) to ensure semantically complete representations, and integrate them with Residual Vector Quantization (RVQ) to construct hierarchical discrete tokens that preserve both high-level semantics and fine-grained details. This design enables dynamic tokenization and de-tokenization across resolutions, supporting both any-resolution visual understanding and arbitrary-resolution image reconstruction. Moreover, dNaViT is plug-and-play compatible with existing large language models without performance degradation.

\item
\textbf{Exceling in Seeing, Painting, and Speaking in a Unified Model.}
LongCat-Next overcomes the longstanding bottleneck of discrete visual modeling, achieving competitive performance with specialized vision understanding models while maintaining strong any-resolution generative quality, even under a 28× compression ratio.
Within DiNA, visual understanding and generation are reformulated as two instances of the same predictive process, differing only in their conditional priors (e.g., image tokens for text generation and text tokens for image generation). This unified formulation effectively reconciles the traditionally competing objectives of understanding and generation, significantly mitigating their modeling conflict in practice.
This unified discrete modeling framework also empowers LongCat-Next with advanced audio comprehension capabilities, low-latency and accurate voice conversation, as well as customizable voice cloning features.

\end{itemize}

This concise architecture is driven by a design that treats vision and audio as intrinsic extensions of the language-centric autoregressive paradigm, rather than as external attachments. Such native integration gives rise to a naturally unified representation across modalities, where \textit{multimodal signals} are internalized in a manner analogous to \textit{linguistic tokens}, in contrast to loosely coupled hybrid approaches (Fig.~\ref{fig:tsne}).

Instantiated on LongCat-Flash-Lite~\cite{longcat-flash-lite} with an A3B (68.5B in total) model size and trained on over 2T tokens, extensive experiments demonstrate that LongCat-Next not only effectively reconciles traditionally competing multimodal objectives, but does so without compromising its foundational language capabilities. 
As a unified model, LongCat-Next excels at seeing, painting, and talking, breaking the performacne ceiling of discrete visual modeling. As as result, it surpasses existing unified frameworks like Qwen3-Omni, outperforms specialized models such as Qwen3VL-A3B on visual understanding benchmarks, and competes favorably with Flux-dev in high-fidelity image generation, particularly in text rendering.
In speech-related benchmarks, LongCat-Next outperforms that both omni and speech-specialized models with comparable parameter scales like Gemini 3.1 Flash-Lite preview and MiMo-Audio respectively. These results demonstrate that the natively discrete paradigm is not merely a conceptual alternative, but a scalable, industrial-strength foundation, one that might bring us closer to a truly unified model of generalist multimodal intelligence.

\section{Methodology}

While the discrete autoregressive paradigm has established a mature and scalable ecosystem for language modeling, approaches for other modalities remain fragmented and lack comparable system-level support. Conceptually, if multimodality is viewed as a form of linguistic modeling within a unified discrete framework, abstracting diverse multimodal signals into a shared discrete token space, this framework offers several key advantages, although this analogy primarily serves as a conceptual purpose.

In particular, the key advantages are as follows: (1) Architectural Synergy, where multimodal data can leverage the established optimization and scaling infrastructure of Large Language Models (LLMs), ensuring efficient training and deployment; (2) Unification of Understanding and Generation, where a single NTP objective merges discriminative understanding and high-fidelity generation, treating them as two aspects of the same underlying predictive logic; (3) Seamless Cross-Modal Interaction, enabling natural interactions between vision, language, audio, and other modalities without task-specific designs; and (4) Native Data Scaling and Unified Self-Supervision, where a universal discrete space flattens multimodal content into unified token sequences, allowing NTP objective to function as a self-supervised mechanism that learns structural and semantic priors directly from large-scale, uncurated \textit{in-the-wild} data.

Despite its conceptual appeal, the field still lacks an industrial-strength training recipe capable of scaling such unified systems. To move beyond conceptual demonstrations toward a production-ready alternative to specialized architectures, such a framework must satisfy the following criteria:

\begin{itemize}
\item \textbf{Performance Parity and Beyond:} The framework must match or surpass the state-of-the-art performance of specialized models in both comprehension and generation. A generalist paradigm is impractical if a substantial performance gap prevents it from replacing existing specialized systems.
\item \textbf{Modality Synergy Instead of Compromise:} Extending the model to encompass multimodality must not degrade its foundational language capabilities. Additional modalities should introduce complementary signals that foster cross-modal synergy, rather than creating optimization trade-offs.
\item \textbf{Infrastructure-Friendly Evolution:} The architecture should remain infrastructure-friendly, enabling a smooth transition from pure language models to native multimodal systems with minimal modality-specific inductive bias, all while preserving compatibility with existing large-scale frameworks.
\end{itemize}

To satisfy the aforementioned criteria, we design the approach entirely upon a discrete autoregressive foundation. Unlike the prevailing \textit{language-plus-auxiliary} paradigm, we eliminate the need to treat non-linguistic signals as continuous external inputs projected into language model. Instead, we unify the optimization objective itself with next-token prediction, internalizing vision, audio, and language within a single shared token representation. This conceptual unification translates the goals of native multimodality into a unified learning paradigm.

\subsection{Model Architecture}

To instantiate this discrete modeling approach, the system is built upon the LongCat-Flash Mixture-of-Experts (MoE) backbone~\cite{meituanlongcatteam2025longcatflashtechnicalreport,longcat-flash-lite}. As illustrated in Fig. \ref{fig:overall}, we adopt a structural decomposition: modality-specific tokenizer and detokenizer pairs are deployed to handle the conversion between raw signals and discrete IDs. Consequently, the decoder-only backbone remains modality-agnostic and serves as a multi-task learner. This design allows the model to natively execute language, visual understanding and generation, as well as audio comprehension and synthesis within a single predictive pipeline. In this section, we introduce the proposed methodology, with in-depth analysis and implementation details provided in Sec.~\ref{sec:model_analysis} and Sec.~\ref{sec:implementation_details}.

\subsection{Vision Tokenizer}
\label{sec:visual_tokenizer}

As the saying goes, \textit{a picture is worth a thousand words}. An image captures a vast spectrum of information, ranging from high-level semantic structures to fine-grained textures and visual details. Compressing high-dimensional visual signals into a finite discrete codebook inevitably introduces information loss, often leading to a performance gap between discrete modeling and continuous representations. Consequently, a prevailing view suggests that visual discretization imposes an intrinsic performance ceiling. This challenge is further compounded by the divergence between representations optimized for understanding versus those for generation, making a semantically complete and unified visual interface difficult to achieve.

To overcome this challenge, we introduce the Discrete Native Vision Transformer (dNaViT). Mirroring the role of language tokenizers, which provide a flexible, near-lossless foundation for unified autoregressive modeling, dNaViT serves as a unified tokenizer for both visual comprehension and generation at any-resolution. We address the limitations of discrete modeling by focusing on two core components: \textit{capacity of visual representation} and \textit{information loss from discretization}. In the following sections, we outline our solutions to ensure that the discrete space achieves the semantic completeness necessary for excelling in both visual understanding and generation.

\subsubsection{Design Motivation}

The success of language modeling is grounded in near-lossless discrete compression via subword tokenization, where language tokenizer encodes semantic content while preserving sufficient structure for faithful reconstruction in a discrete space. Built upon this foundation, the Next-Token Prediction (NTP) paradigm unifies comprehension and generation within a single autoregressive framework by operating directly on these token sequences.

\begin{figure}[t]
    \centering
    \includegraphics[width=1.0\linewidth]{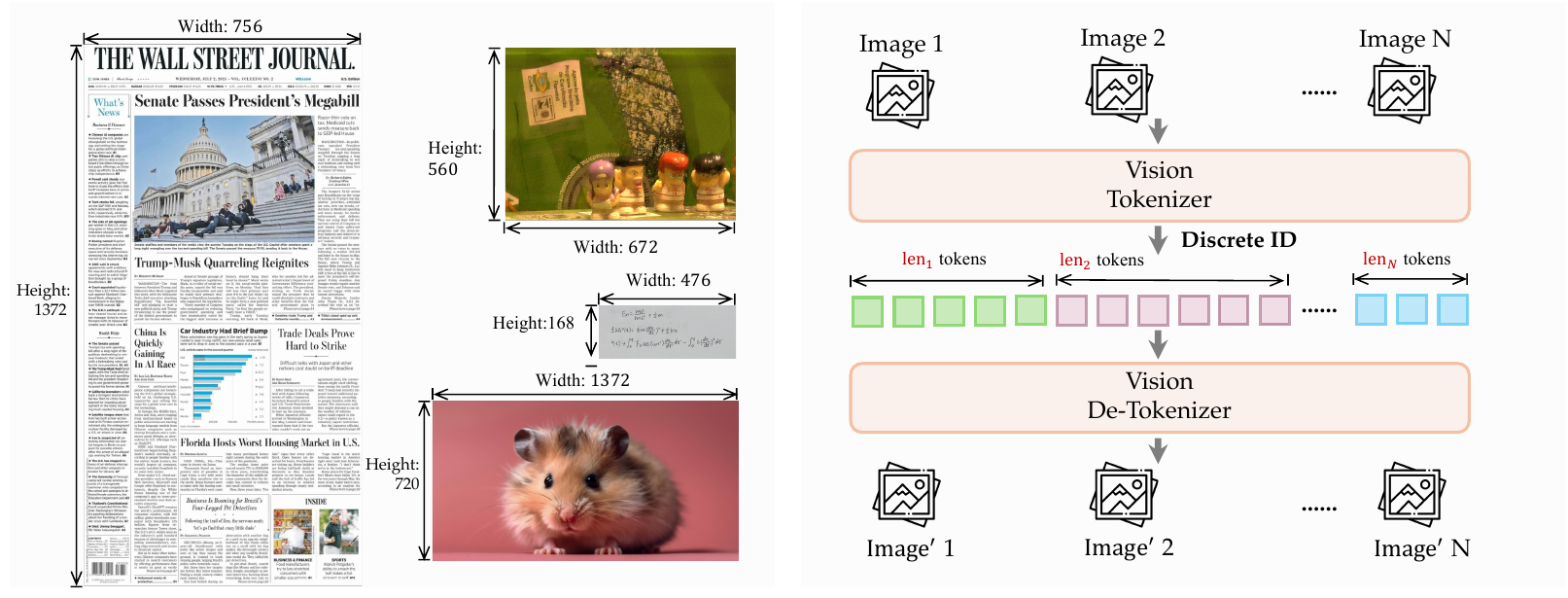}
    \caption{Overview of the discrete Native-Resolution Vision Transformer (dNaViT) design.}
    \vspace{-4pt}
    \label{fig:overall}
\end{figure}

Unlike words, visual information is inherently dense and continuous. Developing a comparable visual tokenizer, however, is hindered by the substantially higher information density of visual signals. To resolve this, we propose the principle of \textit{semantic completeness}: which requires a unified discrete representation to preserve sufficient information from the original visual signal to support both discriminative understanding and high-fidelity generation.

\textbf{Semantic Completeness}: Specifically, the semantic completeness of a discrete representation refers to its ability to serve as an approximately lossless proxy for the original visual signal across a wide range of downstream tasks. Formally, let $I$ denote an input image sampled from the continuous visual manifold $\mathcal{M}$, and let $z = \{idx_1, idx_2, \dots, idx_n\}$ denote the sequence of discrete indices produced by a quantization mapping $Q(\cdot)$. A discrete representation $z$ satisfies semantic completeness if, for any image-centric inquiry $\mathcal{Q}$ associated with task $\mathcal{T}$, the posterior distribution conditioned on $z$ approximates that conditioned on the original image $I$:

\begin{equation}
\mathcal{P}(A \mid z, \mathcal{Q}) \approx \mathcal{P}(A \mid I, \mathcal{Q}),
\end{equation}

where $A$ denotes the optimal response or latent output corresponding to the inquiry. This equivalence implies two fundamental properties:

\begin{itemize}

\item \textbf{Discriminative Invariance}: The discretization process should preserve the core semantic attributes of the original image. For tasks ranging from fine-grained recognition to semantic reasoning, the discrete representation $z$ must retain the critical information contained in the raw pixels $I$, ensuring that the quantization process does not degrade downstream discriminative performance.

\item \textbf{Generative Sufficiency}: Given the high redundancy in pixel space, the discrete codes $z$ should capture the essential visual semantics required for faithful image generation. In particular, the de-tokenizer $D(\cdot)$ should be able to reconstruct the structural and textural content of the image ($I' \approx D(z)$). More importantly, $z$ should function as a semantically sufficient descriptor that provides the language model with a compact yet information-preserving representation.

\end{itemize}

Because the tokenizer is typically fixed prior to large-scale autoregressive training, the induced representation capicity becomes the key factor determining the model’s performance ceiling.
Existing approaches generally fall into three categories: (i) Low-level Reconstructive Models (e.g., VAEs~\cite{kingma2013auto}, VQ-VAEs~\cite{van2017neural}), which have been successfully scaled in works like EMU series~\cite{wang2024emu3,cui2025emu3}, Chameleon~\cite{team2024chameleon}, LWM~\cite{liu2024world} and VILA-U~\cite{wu2024vila} to achieve exceptional pixel-level fidelity but struggle with high-level conceptual reasoning; (ii) Self-supervised Semantic Encoders (e.g., DINOv2~\cite{oquab2023dinov2}, SigLIP~\cite{zhai2023sigmoid}), which are widely adopted to capture structural or contrastive features in various works~\cite{jiao2025unitoken,fan2025unified,zou2025omnimamba,deng2025emergingpropertiesunifiedmultimodal}, exemplified by Janus series~\cite{wu2025janus,chen2025janus}, yet lack the explicit semantic grounding needed for generative reconstruction; and (iii) Encoder-free raw-pixel tokenization, which is championed by methods like EVE series~\cite{diao2024unveiling,diao2025evev2} and NEO~\cite{diao2025pixels} for offering simplicity and scalability, but suffers from pixel redundancy. In this section, we highlight another class of representations with significant potential that remains largely underexplored: \textit{Semantic-and-Aligned Encoders} (SAE). 
They provide a strong foundation for unified discrete visual modeling by enabling semantically rich representations that can support both visual understanding and generation.

\subsubsection{Semantic-and-Aligned Encoder (SAE)}

We argue that language-aligned semantic encoders are particularly well-suited as the pre-quantization space, due to their large-scale language-grounded supervision over diverse image-centric tasks. This endows the representation with two key properties: (i) \textit{semantic richness}, capturing both high-level concepts and fine-grained visual details (e.g., textual details), and (ii) \textit{affinity with language models}, enabling seamless integration into a unified discrete token space.
Formally, we define SAE as a mapping $\mathbf{z_p} = \mathcal{E}_{\text{sae}}(I)$ that projects an image $I$ into a pre-quantization representation $z_p$, which is required to preserve the information necessary for answering diverse image-centric queries $\mathcal{Q}$ (e.g., captioning, OCR, QA, and visual reasoning), formally: $\mathcal{P}(A \mid \mathbf{z_p}, \mathcal{Q}) \approx \mathcal{P}(A \mid I, \mathcal{Q})$,
where $A$ denotes the ground-truth response. To encourage this property, the SAE is trained with a large-scale multi-aspect alignment objective:
\begin{equation}
\mathcal{L}_{\text{SAE}} =
\mathbb{E}_{(I,\mathcal{Q},A)}
\left[
-\log P(A \mid \mathbf{z_p}, \mathcal{Q})
\right].
\end{equation}

In practice, semantic completeness is implicitly enforced through large-scale vision–language training.
This process can be constructed through different progressive stages:(i) \textit{Global Alignment}, where coarse cross-modal correspondence is established through CLIP-style contrastive learning and (ii) \textit{Detailed Learning}: where the model moves beyond coarse contrastive alignment (as in vanilla SigLIP) to capture fine-grained details and nuanced semantics via multi-aspect language-conditioned supervision (as in diverse QAs).

Importantly, this perspective allows us to leverage existing vision–language models as strong SAE approximations. The class of encoders, e.g., QwenViT \cite{qwenvl}, MoonViT \cite{kimivl}, and AIMv2 \cite{aimv2} can be directly adopted as candidates.
This avoids the need for expensive, dedicated SAE training from scratch, while still enabling discretized tokens to inherit rich semantic structure for unified multimodal modeling.

\subsubsection{Tokenization} 
\label{sec:tokenization}

While SAE provides a semantically complete manifold, the subsequent discretization must preserve this information density within a finite discrete space. To bridge the gap between continuous features and discrete tokens without the fidelity loss typical of single-stage quantization, we employ Residual Vector Quantization (RVQ) \cite{rqvae}, which decomposes the quantization of SAE features $\mathbf{z}$ into $L$ cascaded codebook levels. Instead of a single-step mapping, each stage $l \in \{1, \dots, L\}$ recursively encodes the residual error from the previous level. This hierarchical refinement allows quantization to approximate the high-dimensional, semantically rich SAE space with significantly higher fidelity, ensuring that the resulting tokens remain a sufficient proxy for both comprehension and generation.

\textbf{Semantic Tokenization at Native Resolution.} 
We adopt the RVQ approach applied to the latent representations produced by the SAE. This design yields two critical advantages:
(i) Semantic Structuredness: By discretizing features already refined through large-scale vision-language pre-training, the resulting tokens inherit a natural correlation with linguistic space. facilitating convergence in multimodal autoregressive training. (ii) Native Resolution: Instead of relying on a fixed-size bottleneck, we operate on the encoder’s native-resolution latent representations under variable lengths \cite{packnpack}. This allows the tokenizer to handle arbitrary input resolutions while maintaining architectural consistency with language models. The SAE output $\mathbf{z}$ is first projected through a learnable mapping $f_{\text{proj}}$ and then quantized through $L$ cascaded codebook levels:
\begin{equation}
\mathbf{r}_0 = f_{\text{proj}}(\mathbf{z}), \quad
\hat{\mathbf{q}}_l = \operatorname{VQ}(\mathbf{r}_{l-1}), \quad
\mathbf{r}_l = \mathbf{r}_{l-1} - \hat{\mathbf{q}}_l, \quad
\hat{\mathbf{z}} = \sum_{l=1}^{L} \hat{\mathbf{q}}_l,
\end{equation}

\begin{figure}[t]
    \centering
    \includegraphics[width=1.0\linewidth]{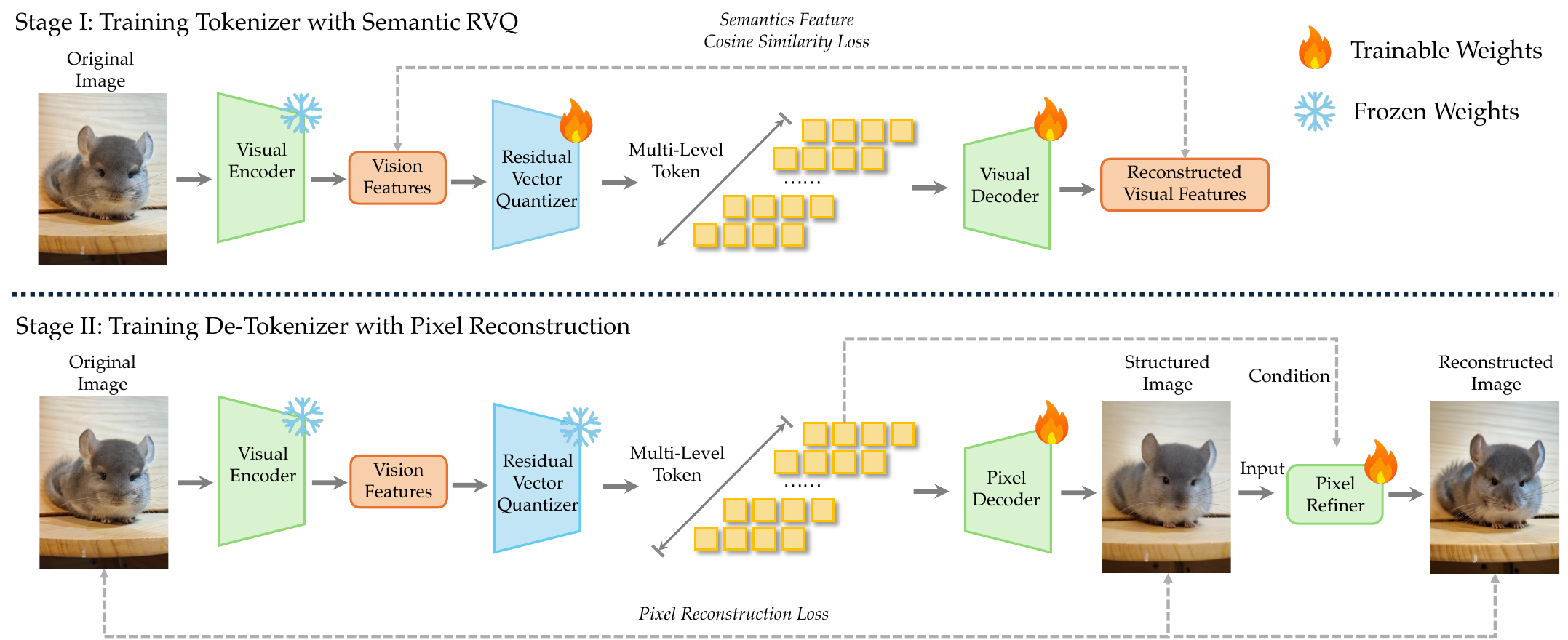}
    \caption{The tokenizer and de-tokenizer training pipeline of the proposed dNaViT, which encodes images into discrete tokens via RVQ and decodes them back into the image space through a pixel decoder at arbitrary resolutions.}
    \label{fig:dnavit}
\end{figure}

where \(\mathbf{r}_0 = f_{\text{proj}}(\mathbf{z})\) is the projected latent representation, \(\hat{\mathbf{q}}_l = \text{VQ}(\mathbf{r}_{l-1})\) is the quantized feature at level \(l\), \(\mathbf{r}_l = \mathbf{r}_{l-1} - \hat{\mathbf{q}}_l\) denotes the residual error, and \(\hat{\mathbf{z}} = \sum_{l=1}^{L} \hat{\mathbf{q}}_l\) is the final quantized representation.

Given a batch of continuous residual vectors $\{\mathbf{r}_j\}$ from the projection layer, we update the codebook entries using an exponential moving average (EMA) \cite{van2017neural,rqvae} rather than gradient descent. Specifically, for each codebook entry $\mathbf{e}_k$, we maintain a running cluster size $N_k \leftarrow \gamma N_k + (1-\gamma)|\mathcal{S}_k|$ and an embedding sum $\mathbf{m}_k \leftarrow \gamma \mathbf{m}_k + (1-\gamma)\sum_{j \in \mathcal{S}_k} \mathbf{r}_j$, where $\mathcal{S}_k = \{ \mathbf{r}_j : \arg\min_i \|\mathbf{r}_j - \mathbf{e}_i\| = k \}$ denotes the set of residual vectors assigned to entry $k$, and $\gamma$ is the decay factor. The codebook entry is then updated as:

\begin{equation}
\mathbf{e}_k \leftarrow \frac{\mathbf{m}_k}{N_k}.
\end{equation}

We apply Laplace smoothing to $N_k$ for numerical stability, and re-initialize inactive entries ($N_k < 1$) from current batch to maintain codebook utilization.
To provide supervision for $f_{\text{proj}}$, we introduce a lightweight decoder that reconstructs the pre-quantization semantic features from the quantized embeddings. The overall quantization objective is defined as:

\begin{equation}
\mathcal{L}_{\text{quant}} = \lambda_c \mathcal{L}_{\text{commit}} + \lambda_s \mathcal{L}_{\text{semantic}},
\end{equation}

where $\mathcal{L}_{\text{commit}}$ is the standard commitment loss that encourages the projected representation $\mathbf{r}_0$ to stay close to the assigned codebook entries, averaged across all $L$ residual levels, and $\mathcal{L}_{\text{semantic}}$ is the semantic reconstruction loss with feature cosine similarity.
As a result, the tokenization pipeline consists of three stages: (1) semantically aligned feature extraction via SAE, (2) hierarchical discretization using RVQ with EMA-updated codebooks, and (3) semantic tokens serving as input for multimodal understanding and generation via autoregressive modeling.

\subsubsection{De-tokenization}
\label{subsubsection:de-tokenization}

After tokenization training, both the SAE encoder and the codebook are frozen. This design is motivated by the premise that if the pre-quantization representation is semantically complete, then the resulting discrete tokens should inherently retain sufficient information, even without further adaptation. Specifically, the de-tokenizer is centered on a pixel decoder, a Vision Transformer~\cite{dosovitskiy2021imageworth16x16words} trained to reconstruct images from the discrete code embeddings alone. The pixel decoder recovers spatial layout, object structure, and identifiable content directly from the quantized representation, producing a structurally faithful reconstruction. We train the pixel decoder with the following objective:

\begin{equation}
\mathcal{L}_{\text{dec}} = \lambda_1 \mathcal{L}_{\text{pixel}} + \lambda_2 \mathcal{L}_{\text{percep}} + \lambda_3 \mathcal{L}_{\text{align}},
\end{equation}
With these objectives alone, the pixel decoder is already capable of recovering coherent spatial layouts, object structures, and semantic content directly from discrete tokens, demonstrating that the quantized representation retains sufficient information for faithful reconstruction. However, the resulting images tend to be overly smooth, with diminished high-frequency details and perceptual sharpness. To slightly improve visual fidelity, we introduce a lightweight image refiner trained with a flow-matching objective $\mathcal{L}_{\text{flow}}$, conditioned on the pixel decoder’s reconstruction and the discrete code embeddings. The reconstruction serves as a structural anchor, constraining the refinement process and allowing the model to focus on enhancing fine-grained textures without introducing noticeable artifacts.

\subsubsection*{Dicussion about the Information Recovery}

To achieve semantic completeness, a unified representation $\mathbf{z}_\mathbf{p}$ must not only support semantic comprehension but also possess \textit{generative sufficiency}, i.e., the ability to recover the original visual signal. While prior works~\cite{rae,zheng2025diffusion} (e.g., RAE) suggest reconstruction capability primarily to semantically enriched latent priors, we take a different perspective and argue that such recoverability also arises from intrinsic architectural properties of the encoder. We refer to this property as \textit{intrinsic information recovery}.

\begin{figure}[h]
    \centering
    \includegraphics[width=1.0\textwidth]{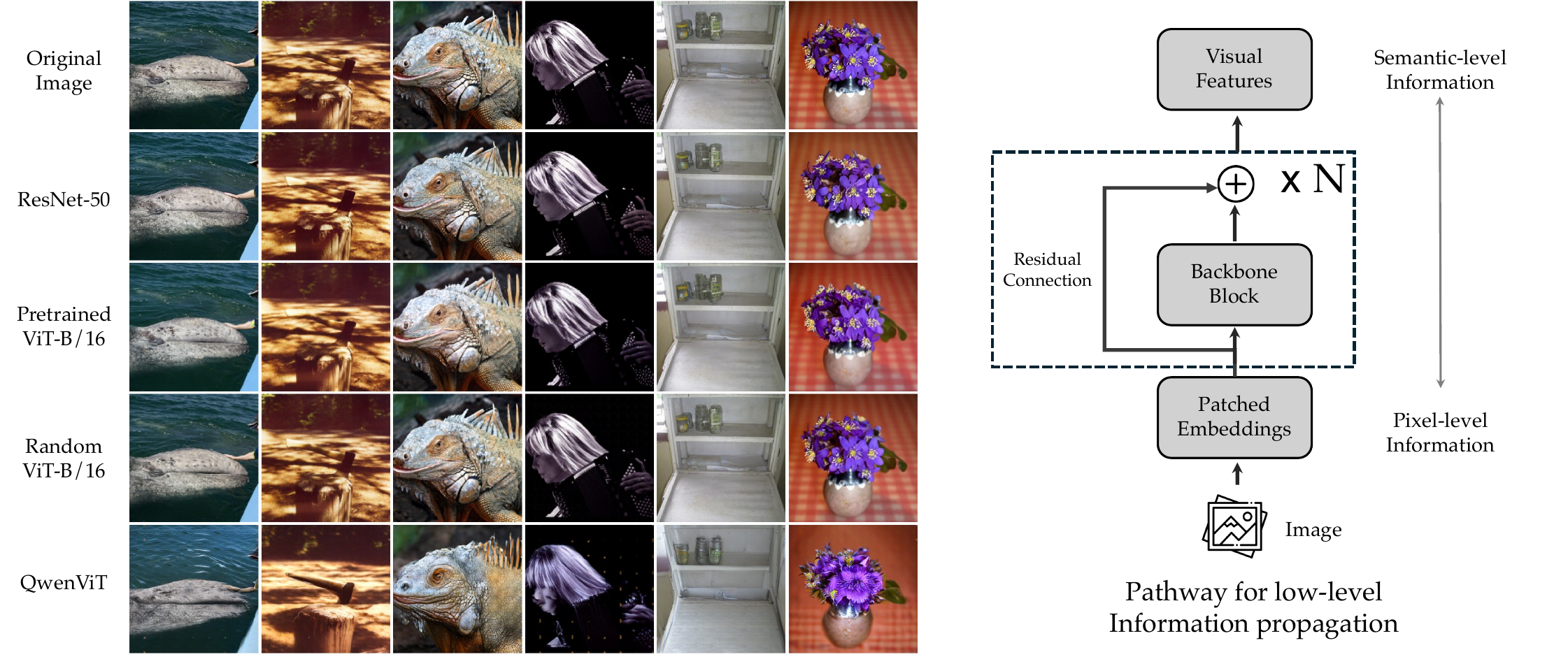} 
    \caption{Visual reconstruction from different frozen vision encoders, trained with a lightweight pixel decoder. The results suggest that residual connections inherently preserve a latent pathway for low-level signal propagation.}
    \label{fig:reconstruction_visuals}
\end{figure}

Empirically, we observe that even a randomly initialized ViT-Base \cite{dosovitskiy2020image} exhibits strong reconstruction capability, outperforming its pretrained counterpart in terms of reconstruction fidelity (see Sec.~\ref{information_recovery}). This suggests that effective information recovery is not solely a consequence of learned semantics, but is closely tied to the underlying network structure. This phenomenon can be understood through the residual architecture of modern vision encoders, which inherently preserves a latent pathway for low-level signal propagation, allowing effective reconstruction even without supervision. Formally, we formulate the encoder as a sequence of $L$ residual blocks. The final latent representation $\mathbf{z}_p$ can be expressed as the additive accumulation of the initial projection and subsequent feature increments:

\begin{equation}
\mathbf{z}_p = \mathbf{x}_0 + \sum_{l=1}^{L} \mathcal{F}_l( \mathbf{x}_{l-1}) = \underbrace{\mathbf{x}_0 + \mathcal{F}_1(\mathbf{x}_0) + \dots + \mathcal{F}_L(\mathbf{x}_{L-1})}_{\textit{pixel-level} \xrightarrow{\hspace{2cm}} \textit{semantic-level}}, \quad \text{where} \quad \mathbf{x}_l = \underbrace{\mathbf{x}_{l-1} + \mathcal{F}_l(\mathbf{x}_{l-1})}_{\text{residual connection}},
\end{equation}

where the identity mapping $\mathbf{x}_{l-1}$ ensures that fine-grained visual signals from earlier layers are not overwritten by higher-level semantic abstractions. Instead, they are propagated through the network $\mathcal{F}_L$ and progressively integrated into $\mathbf{z}_p$, resulting in a representation that retains high mutual information $\mathcal{I}(I; \mathbf{z}_p)$ with the original image. Surprisingly, this analysis suggests that modern encoders, even an encoder optimized primarily for semantic alignment (such as SAE),  naturally retain the structural elements required for image recovery, providing a structural foundation for semantic completeness prior to discretization.

\subsubsection{dNaViT: Discrete Native Resolution Vision Transformer}

Building upon the principles of semantic completeness and discrete quantization approach, we introduce the Discrete Native-Resolution Vision Transformer (dNaViT), a vision tokenizer designed to play a role analogous to language tokenizers in large language models. The central objective of dNaViT is to establish a discrete visual interface where images can be represented, processed, and generated entirely through sequences of semantically meaningful tokens.

Unlike conventional vision encoders that rely on fixed-resolution inputs, dNaViT operates directly on images at their \emph{native resolution}, avoiding information loss and spatial distortions introduced by resizing, cropping, or padding. This design allows the model to faithfully preserve visual details across arbitrary image scales.
More importantly, dNaViT constructs a hierarchy of semantically aligned discrete tokens that serve as a structured proxy for visual content. These tokens move beyond low-level quantization by encoding information across multiple semantic levels, preserving both global structure and local details within a unified discrete representation.

Built upon this hierarchical representation, dNaViT effectively induces a language-like visual vocabulary, where images are expressed as sequences of discrete symbols that are directly compatible with autoregressive modeling.
Within this symbolic space, image understanding and generation are unified under a single formulation, where captioning and synthesis correspond to inverse sequence modeling over shared token representations, enabling a single autoregressive model to jointly handle visual perception and generation.

\textbf{Native Visual Representation.} 
dNaViT provides a topological interface that maps images to discrete token sequences, enabling a unified symbolic representation for multimodal modeling. Beyond this interface, the semantic content is not fixed within the tokenizer, but is instead internalized in the language model’s visual embedding space.
Specifically, visual codebook embeddings are randomly initialized and then co-evolve with linguistic tokens under a shared autoregressive objective. Through this joint optimization process, the model learns \emph{native visual representations} that are fully adapted to the LLM’s internal representation space.
This mechanism enables the model to develop a "native language" for vision, shifting the representation paradigm from externally defined quantized IDs to learned, model-internal visual semantics, enabling a truly native multimodal modeling framework.

\subsection{Audio Tokenizer}

\begin{figure}[h]
    \centering
    \includegraphics[width=0.8\linewidth]{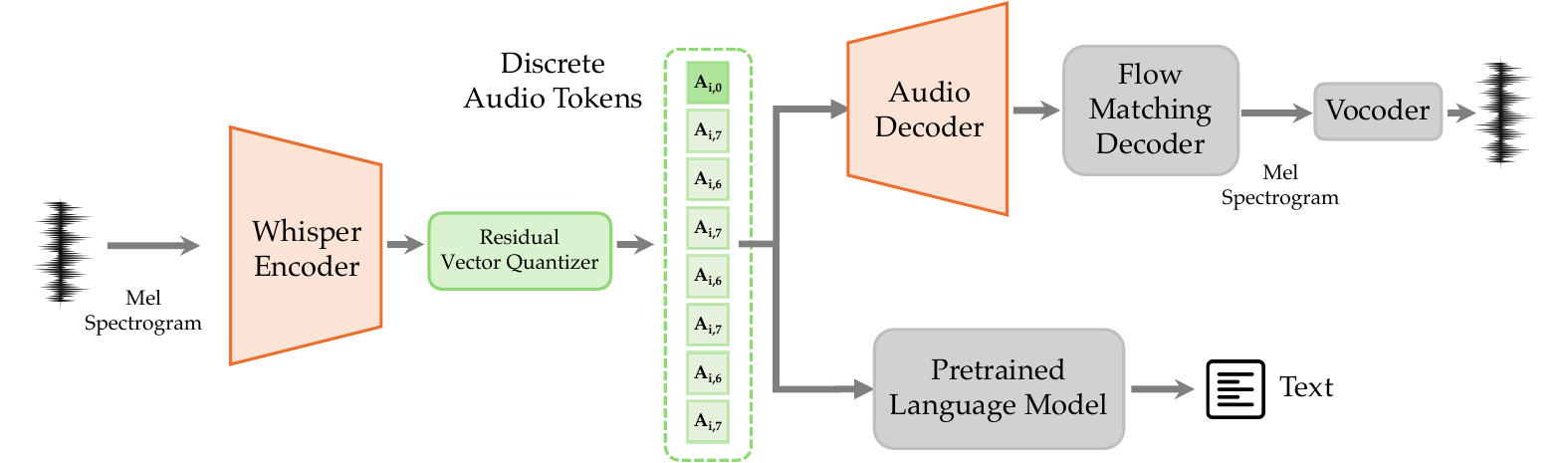}
    \caption{Illustration of the audio tokenizer framework.}
    \label{fig:audio_tokenizer}
    \vspace{-2pt}
\end{figure}

We design an audio tokenizer that transforms continuous speech into discrete tokens, preserving both semantic and acoustic information. 
As shown in Fig.~\ref{fig:audio_tokenizer}, the input audio is first passed to a Whisper encoder~\cite{radford2023robust} for audio feature extraction. The features are then downsampled by a factor of 4 before being quantized into discrete tokens by an 8-layer residual vector quantizer (RVQ).

The resulting discrete audio tokens are forwarded along two parallel branches. 
In the first branch, they are fed into a frozen pretrained large language model (LLM). Through extensive training on diverse audio understanding tasks, these tokens encode both semantic and acoustic information necessary for downstream tasks. They simultaneously align with the LLM's textual embedding space. 
The audio representations are transferable across LLM architectures and vocabularies. Based on this finding, we adopt a smaller LLM, Qwen3-1.7B~\cite{yang2025qwen3}, in the tokenizer training phase to improve efficiency and discard it in the later stages.

In the second branch, the audio tokens are passed to a decoder whose architecture is symmetric to that of the encoder. It reconstructs coarse Mel spectrograms from the input tokens. 
To further enhance reconstruction fidelity, we introduce a flow-matching model~\cite{lipman2022flow} after the decoder, which refines the coarse Mel spectrograms. The resulting refined Mel spectrograms are then converted into audio waveforms by a vocoder~\cite{kong2020hifi}. The training objective of audio tokenizer includes Mel spectrogram reconstruction loss, RVQ commit loss, and LLM loss for audio understanding:
\begin{equation}
\mathcal{L}_{\text{audio}} = \lambda_1 \mathcal{L}_{\text{recon}} + \lambda_2 \mathcal{L}_{\text{commit}} + \lambda_3 \mathcal{L}_{\text{llm}}.
\end{equation}

\subsection{Language Model Backbone}

Under the DiNA framework, we observe that a language backbone equipped with a modality-agnostic MoE inherently functions as a \textit{multi-task learner} across different modalities, dynamically allocating capacity to multiple objectives without requiring modality-specific design. We adopt LongCat-Flash-Lite A3B~\cite{longcat-flash-lite} as the decoder-only backbone, trained from scratch with 68.5B total parameters and an average of 3B activated parameters (ranging from 2.9B to 4.5B depending on the context), the design of Zero-Expert and Shortcut MoE is adopted in architecture~\cite{meituanlongcatteam2025longcatflashtechnicalreport}. Unlike conventional approaches to multimodal integration, which often introduce modality-specific branching~\cite{deng2025emergingpropertiesunifiedmultimodal,qwenvl} (e.g., modality-aware MoE, 3D RoPE, or bidirectional attention), our framework enforces a unified latent space in which all tokens, textual, visual, and acoustic, are processed through a single modality-agnostic pathway. We provide further analysis and empirical findings in Sec.~\ref{moe_label}.

\begin{figure}[t]
    \centering
    \includegraphics[width=0.7\linewidth]{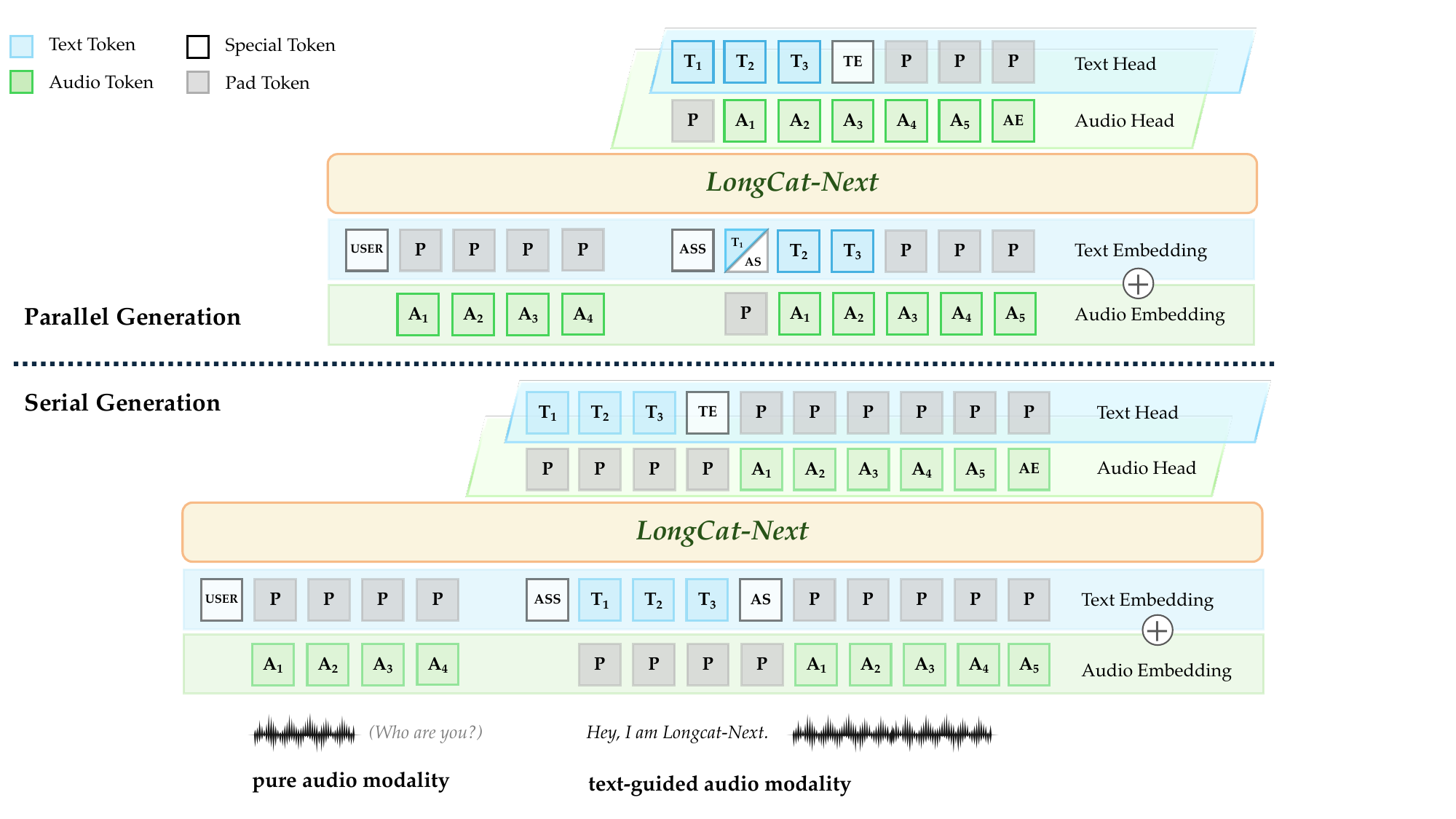}
    \caption{Two speech generation strategies with internal linguistic guidance. The user's input audio is treated as a pure audio modality, whereas the output generated by Longcat-Next (or the assistant) is regarded as a text-guided audio modality. To indicate modality conversion and facilitate segment alignment, we introduce three special tokens: AS (Audio Start), AE (Audio End), and TE (Text End), which respectively denote the beginning and end of an audio segment, and the end of a text segment.}
    \vspace{-4pt}
    \label{fig:audiotext}
\end{figure}

\subsection{Multimodality Component}

Unlike conventional multimodal approaches that project continuous visual features into embedding space of the language model. All multimodal embeddings are randomly initialized and jointly trained with the native model, ensuring that vision, audio, and language share a unified token-level representation paradigm. 

\subsubsection{End-to-End Multimodal Embedding}

Specifically, the visual embedding uses a codebook of size (8 $\times$ 16,384), with multi-level tokens combined through a multi-level summation for efficient representation. These embeddings are not shared across levels, allowing each level to capture complementary information. 
The audio component follows the same design but with decreasing codebook sizes. Its multi-level embeddings are also randomly initialized and learned during training. 
Importantly, these embeddings are learned end-to-end to support both multimodal understanding and generation. The pre-quantization features are used solely for establishing the discrete clustering assignments in the Residual Vector Quantization (RVQ) process, rather than directly dictating the embedding values themselves.

\subsubsection{Multimodality Head}

In this architecture, the native model is designed to learn semantic-level compression, while modality-specific signals are compacted through modality-aware tokenizers. Visual and audio inputs are encoded into multi-level discrete tokens, enabling efficient representation of rich perceptual information. During training, we adopt a multi-level supervision scheme: the language model performs single-step autoregressive prediction, while multi-level tokens are decoded in parallel to reconstruct modality-specific details. Specifically, the LLM head is a standard MLP, while multimodal tokens are decoded by a task-aware DepthTransformer~\cite{rqvae}, which generates multi-level outputs to recover structured visual and audio information.

\subsubsection{Internal Linguistic Guidance}

For recent large speech language models, whether speech interaction can inherit the same linguistic capabilities as text is an important topic.
Following the Moshi~\cite{defossez2024moshi} approach, we provide internal linguistic guidance by explicitly modeling text as part of the speech generation process. In our method, segment-level aligned text tokens and audio tokens are first embedded with dedicated input embedding layers and then fused via element-wise summation. We refer to this format as the text-guided audio modality. In contrast, for real-time user input scenarios where textual guidance is often unavailable, we define these instances as the pure audio modality. 
As shown in Fig.~\ref{fig:audiotext}, within the text-guided audio modality, LongCat-Next is capable of two generation strategies:
\textbf{(1) Parallel Generation}: The model generates text and audio tokens simultaneously at each step but intentionally delays the generation of the first audio token by a specified number of steps to maintain alignment with the text. We illustrate the case with a delay of one in Fig.~\ref{fig:audiotext}.
This strategy eliminates response latency and is more suitable for full-duplex modeling. \textbf{(2) Serial Generation}: The model first generates the guided text segment, followed by the corresponding audio segment. Within this strategy, the model only needs to predict tokens from a single modality at each step, which simplifies the process and avoids conflicts between modality representations. As a result, this approach ensures high linguistic quality in the generated speech.

To unify these strategies, we propose a general training paradigm for the text-guided audio modality: for each aligned text-audio segment, the delay is randomly selected from one up to the length of the text segment. This approach encourages the model to learn robust semantic alignment between text and audio representations, enabling it to generate text-guided speech at arbitrary delay steps. Consequently, serial generation and parallel generation can be regarded as two extreme cases within this unified framework.

\section{Main Experiments}
Our model is a unified omni-modal system capable of processing text, vision, and speech simultaneously. Accordingly,
we compare against Qwen3-Omni-A3B-Instruct, the current state-of-the-art omni modal, as the primary baseline to
situate our model within the competitive landscape of full-modality systems. In addition, to rigorously assess the visual
understanding capability of our model in isolation, we further include Qwen3-VL-A3B, a dedicated vision-language
model optimized purely for visual comprehension, as a complementary reference. In the audio domain, we also compare with audio-specialized models including MiMo-Audio~\cite{zhang2025mimo}, Kimi-Audio~\cite{ding2025kimi}, and Step-Audio-2-mini~\cite{wu2025step}. We conduct a comprehensive evaluation across the following dimensions:

\subsection{Main Results}

\subsubsection{Visual Understanding}
\textbf{STEM.}

We conduct a comprehensive evaluation of our model across a diverse suite of multimodal reasoning benchmarks to rigorously assess its reasoning capabilities under complex logic. 
The evaluated datasets encompass multi-disciplinary queries (MMMU~\cite{yue2024mmmu}, MMMU-Pro~\cite{yue2025mmmu}) and mathematics-focused tasks (MathVista~\cite{lu2023mathvista}, MathVision~\cite{wang2024measuring}).
On mathematical reasoning benchmarks, our model demonstrates superior performance, achieving the highest scores among all compared models on MathVista (83.1) and MathVision (64.7). Notably, it outperforms specialist MLLMs such as InternVL3.5-A3B-Flash and Qwen3-VL-A3B-Instruct in these domains. In comprehensive multi-discipline evaluations, our model exhibits competitive performance; it surpasses Qwen3-Omni-A3B-Instruct on both MMMU (70.6) and MMMU-Pro (60.3).

We extend our evaluation to VisuLogic~\cite{DBLP:journals/corr/abs-2504-15279} and BabyVision~\cite{DBLP:journals/corr/abs-2601-06521}, which serve as rigorous testbeds for vision-centric reasoning. Although we did not specifically optimize towards these datasets, the empirical results are surprisingly strong. On the VisuLogic benchmark, our model secures the top position with a score of 29.4, surpassing baselines including InternVL3.5-A3B-Flash (28.4) and Gemini2.5-Flash-Lite (26.1). Similarly, on BabyVision, the model maintains a competitive score of 14.4, outperforming InternVL3.5-A3B-Flash and Qwen3-Omni-A3B-Instruct, while standing moderately behind Gemini2.5-Flash-Lite (19.6) and Qwen3-VL-A3B-Instruct (16.0). These unexpected gains suggest that our model develops robust, emergent generalization capabilities for decoding complex visual logic puzzles.

\textbf{OCR.} To comprehensively evaluate the model's capability in document understanding 
and information extraction across diverse document types, we conduct assessments on OCRBench~\cite{ocrbench}, OCRBenchV2~\cite{ocrbenchv2}, 
DocVQA~\cite{docvqa}, and OmniDocBench~\cite{omnidocbench}. OmniDocBench is a comprehensive and challenging benchmark encompassing a wide variety of document types — including academic papers, financial reports, and administrative forms. As shown in Table \ref{tab:academic_results}, LongCat-Next-A3B scores 0.152 on $\text{OmniDocBench}_{\text{en}}$ and 0.226 on $\text{OmniDocBench}_{\text{zh}}$, performing better than the other baselines listed and demonstrating that discrete modeling need not compromise fine-grained textual perception even under high-complexity, text-dense conditions. On the OCRBench benchmark, our model achieves a score of 865, outperforming baselines such as Qwen3-Omni-A3B-Instruct, GPT5-minimal, and Gemini2.5-Flash-Lite. 

Furthermore, to 
evaluate the model's ability to process structured visual information 
in charts and figures, we additionally benchmark on ChartQA~\cite{chartqa}, 
CharXiv~\cite{charxiv}, and InfoVQA~\cite{infovqa}. As shown in Table \ref{tab:academic_results}, our model achieves the best performance among all models  on $\text{CharXiv}_{\text{RQ}}$ and ChartQA, with scores of 60.1 and 88.0, respectively. On $\text{CharXiv}_{\text{DQ}}$, it obtains a score of 89.9, outperforming other baselines and ranking second only to Gemini2.5-Flash-Lite.

\begin{figure}[t]
    \centering
    \includegraphics[width=1.0\linewidth]{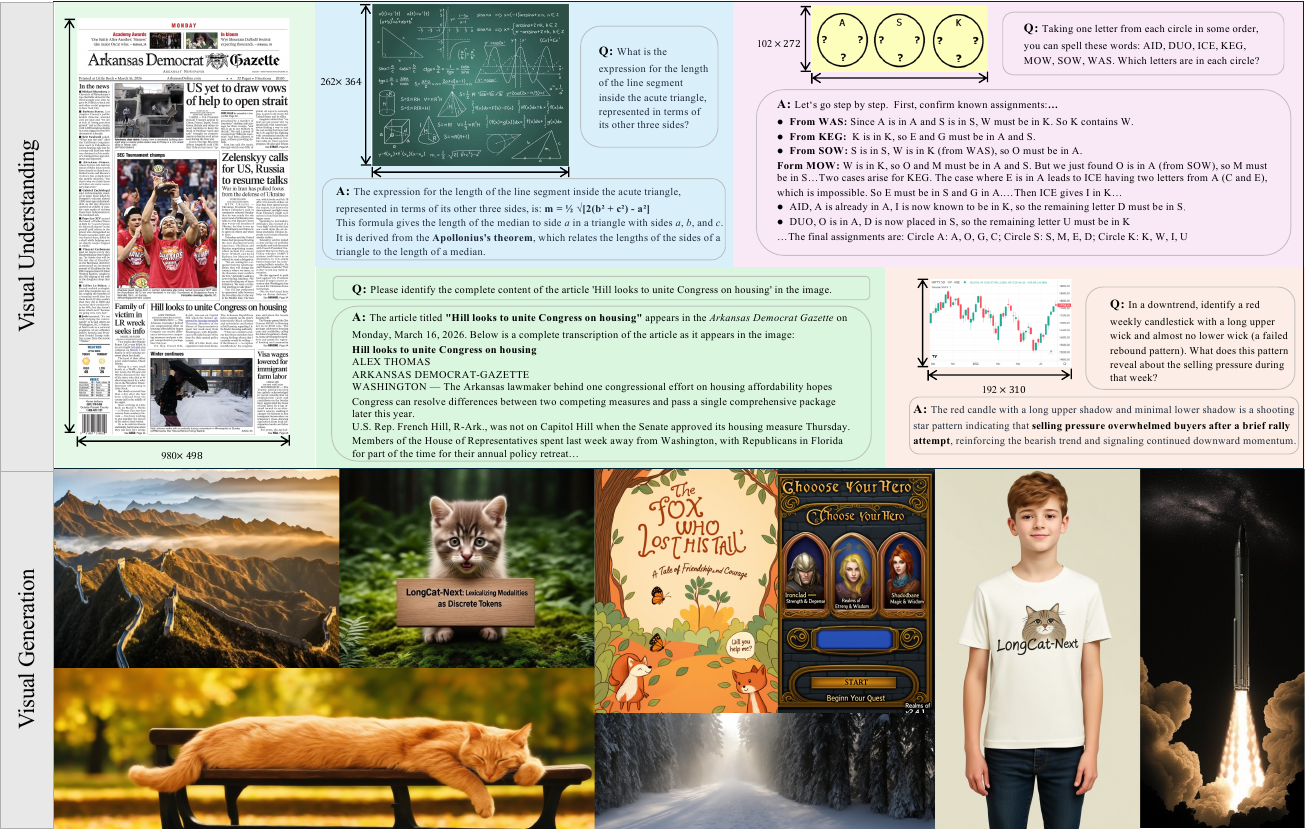}
    \caption{The understanding and generation cases of LongCat-Next at arbitrary resolution.}
    \label{fig:case_und_gen}
\end{figure}

\paragraph{General Domains.} 
We conduct a comprehensive evaluation across multiple mainstream benchmarks, such as MMBench~\cite{DBLP:conf/eccv/LiuDZLZZYWHLCL24}, RealWorldQA~\cite{realworldqa2024}, and MMStar~\cite{DBLP:conf/nips/ChenLDZZCDWQLZ24}, to systematically measure the general VQA capabilities. Across general-domain evaluations, our proposed model demonstrates highly competitive performance. On the MMStar benchmark, our model achieves a score of 69.3, outperforming baselines such as Qwen3-Omni-A3B-Instruct (68.5) and GPT5-minimal (65.2). Notably, it maintains performance levels that are closely comparable to specialist multimodal large language models (MLLMs), including Qwen3-VL-A3B-Instruct (72.1). Similarly, on RealWorldQA (72.0), our model yields robust results, successfully surpassing Gemini2.5-Flash-Lite (70.5). For CountBench, our model secures a commendable score of 82.5, performing comparably to Qwen3-Omni-A3B-Instruct (83.8), although it trails slightly behind the top-performing models in this specific category, such as Gemini2.5-Flash-Lite (90.0).


\paragraph{Graphical User Interface (GUI).}
We evaluate UI perception capabilities across diverse digital environments through GUI-grounding tasks like OSWorld-G~\cite{xie2025scalingcomputerusegroundinguser}, and ScreenSpot-V2~\cite{wu2024atlas}. Our model demonstrates highly competitive performance, maintaining parity with Qwen3-Omni-A3B-Instruct and Qwen3-VL-A3B-Instruct. This validates that our architecture effectively captures the visual granularity required for complex interfaces. While the current data mixture contains a limited proportion of high-resolution samples, optimizing for higher-density modeling remains a key objective for our next iteration.

\begin{table}[t]
\centering
\small 
\caption{Comparison of LongCat-Next and top-tier models on vision benchmarks. Values marked with * are sourced from public reports.}
\label{tab:academic_results}
\vspace{1em}
\setlength{\tabcolsep}{3pt}
\begin{tabularx}{\textwidth}{c*{7}{>{\raggedright\arraybackslash}X}}
\toprule
\multirow{3}{*}{\textbf{Benchmark}} & \multirow{3}{=}{\textbf{LongCat-Next}} & \multirow{3}{=}{\textbf{Qwen3-Omni-A3B-Instruct}} & \multirow{3}{=}{\textbf{GPT5-minimal}} & \multirow{3}{=}{\textbf{Gemini2.5-Flash-Lite}} & \multicolumn{2}{c}{\textbf{Specialist MLLMs}} \\
 \cmidrule(r){6-7}
  & &&&&\textbf{InternVL3.5-A3B-Flash} & \textbf{Qwen3-VL-A3B-Instruct}  \\
\midrule
\multicolumn{7}{c}{\textbf{STEM \& Reasoning}} \\
\midrule
MMMU-Pro & 60.3 & 57.0$^*$ &62.7$^*$ & 64.1 & -- & 60.4$^*$ \\ 
MMMU$_{\text{val}}$ & 70.6 & 69.1$^*$ &74.4$^*$ & 74.9 & 75.6$^*$ &  74.2$^*$ \\ 
MathVista$_{\text{mini}}$ & 83.1 & 75.9$^*$ & 50.9$^*$ & 78.2 &80.9$^*$& 80.1$^*$\\
MathVision & 64.7 & 56.3$^*$ & 45.8$^*$ & 61.9 &55.7$^*$& 60.2$^*$\\
VisuLogic & 29.4 & 20.0 & -- & 26.1 & 28.4 & 23.0$^*$  \\
BabyVision & 14.4 & 11.9 & -- & 19.6 & 11.3 & 16.0 \\
\midrule
\multicolumn{7}{c}{\textbf{OCR \& Doc \& Chart}} \\
\midrule
 OmniDocBench$_{\text{en}}$ $\downarrow$ & 0.152 & 0.289 & 0.174$^*$ & 0.240 & 0.246 & 0.183$^*$\\
OmniDocBench$_{\text{zh}}$ $\downarrow$ & 0.226 & 0.406 & 0.389$^*$ & 0.312 &  0.359 & 0.253$^*$ \\
CharXiv$_{\text{DQ}}$ & 89.9 & 78.8 & 79.5$^*$ & 91.2 & 81.8$^*$ & 85.5$^*$ \\
CharXiv$_{\text{RQ}}$ & 60.1 & 42.8 & 57.8$^*$ & 60.0 & 48.0$^*$ & 48.9$^*$ \\
ChartQA & 88.0 & 86.8$^*$ & 59.1$^*$ & 79.0 & 87.4$^*$ & 86.8$^*$\\
InfoVQA$_{\text{test}}$ & 83.3 & 84.4 & 69.9$^*$ & 82.2 & 81.4$^*$& 81.8$^*$\\
DocVQA$_{\text{test}}$ & 94.2 & 94.1 & 89.6$^*$ & 91.8 & 94.2$^*$ & 95.0$^*$\\
OCRBench & 86.5 & 85.4$^*$ & 78.7$^*$ & 84.8 & 88.0$^*$ & 90.3$^*$ \\
OCRBenchV2$_{\text{en}}$ & 58.5 & 62.0 & 48.2$^*$ & 51.5 & 47.0 & 63.2$^*$ \\
OCRBenchV2$_{\text{zh}}$ & 59.3 & 60.5 & 37.7$^*$ & 39.5 & 48.2 & 57.8$^*$ \\
\midrule
\multicolumn{7}{c}{\textbf{Agent}} \\
\midrule
OSWorld-G & 58.3 & 60.3 & -- & -- & 42.4$^*$ & 60.5$^*$ \\
ScreenSpot-V2 & 88.3 & 91.8 & -- & -- & 87.3$^*$ & 87.7 \\
\midrule
\multicolumn{7}{c}{\textbf{General}} \\
\midrule
MMStar & 69.3 & 68.5$^*$ & 65.2$^*$ & 74.93 & 72.0 & 72.1\\
RealWorldQA & 72.0 & 72.9 & 77.3$^*$ & 70.5 & 72.3 & 73.7\\
CountBench & 82.1 & 83.8 & 87.8$^*$ & 90.0 & 84.2 & 90.6\\

\bottomrule
\end{tabularx}

\end{table}

\subsubsection{Visual Generation}

We evaluate our model's text-to-image (T2I) capability across a diverse set of public benchmarks, covering compositional reasoning, prompt fidelity, long-text understanding, world knowledge, and text rendering.
We evaluate on GenEval \cite{ghosh2023genevalobjectfocusedframeworkevaluating} (compositional alignment), DPG-Bench \cite{hu2024ella} (prompt following),  WISE \cite{niu2025wise} (world knowledge and reasoning), and LongText-Bench \cite{geng2025xomni} / TIFF \cite{wei2025tiif} / CVTG-2K\cite{du2025textcrafter} (text rendering).
Baselines are grouped into (i) unified multimodal models and (ii) specialized T2I models.

\begin{table}[t]
\centering
\caption{Comparison with specialized T2I models. Despite significantly smaller model, our model achieves competitive performance across multiple benchmarks, demonstrating strong industrial-level generation capability. Values marked with * are sourced from public reports.}
\vspace{1em}
\label{tab:t2i_specialized}
\resizebox{1.0\textwidth}{!}{
\begin{tabular}{lllllllll}
\toprule
\textbf{Model} & \multicolumn{1}{c}{\textbf{GenEval}} & \multicolumn{1}{c}{\textbf{DPG}} & \multicolumn{1}{c}{\textbf{LongText-EN}} & \multicolumn{1}{c}{\textbf{LongText-ZH}} & \multicolumn{1}{c}{\textbf{WISE}} & \multicolumn{1}{c}{\textbf{TIFF}} & \multicolumn{1}{c}{\textbf{CVTG}}  \\
\midrule
Emu-3.5 \cite{cui2025emu3}              & 72.67 & ${89.42}^{*}$ & ${97.60}^{*}$ & ${92.80}^{*}$ & 57.64 & ${89.48 / 88.18}^{*}$ & ${91.23}^{*}$ \\

Qwen-Image 2507  \cite{wu2025qwenimagetechnicalreport}      & ${87.00}^{*}$ & ${88.32}^{*}$ & ${94.30}^{*}$ & ${94.60}^{*}$ & ${63.00}^{*}$ & ${86.10}^{*}$ / ${86.80}^{*}$ & ${82.88}^{*}$ \\
Gemini 2.5 Flash Image \cite{comanici2025gemini25pushingfrontier} & 79.67 & 85.82 & 86.04 & - & 76.27  & 90.53 / 90.80 & ${73.64}^{*}$ \\
FLUX.1-dev  \cite{labs2025flux1dev}        & ${66.00}^{*}$ & ${84.00}^{*}$ & ${60.70}^{*}$ & ${0.50}^{*}$  & ${50.00}^{*}$ & ${71.10 / 71.80}^{*}$ & ${49.65}^{*}$ \\
Seeddream 3.0  \cite{gao2025seedream} & - & ${94.31}^{*}$ & ${89.60}^{*}$ & ${87.80}^{*}$  & - & ${86.02 / 84.31}^{*}$ & ${59.24}^{*}$ \\

\midrule
LongCat-Next  & 84.44 & 84.66 & 93.15 & 89.08 & 57.00 & 82.85  / 84.38 & 76.36 \\
\bottomrule
\end{tabular}
}
\end{table}

\paragraph{Analysis of Results.} \textbf{(1) Clear Advantage over Unified Multimodal Models.}
Our model outperforms prior unified approaches across most benchmarks, establishing a new performance level for generation within unified architectures. The gains are particularly significant on long-text understanding and text rendering (LongText, TIFF, CVTG-2K), where our model shows clear margins over existing methods.
We attribute this improvement to our \textbf{Unified architecture}, which tightly integrates language understanding with image generation. By enabling stronger semantic planning before synthesis, the model better preserves textual intent in complex scenarios such as multi-object composition and text rendering, where prior unified models often struggle. \textbf{(2) Competitive Performance with Specialized T2I Models.}
Despite being a unified model, our approach achieves competitive performance compared to specialized T2I systems. Notably, it matches or exceeds larger models on GenEval and LongText-Bench, and remains competitive on text rendering benchmarks.
This suggests that strong generation capability can be achieved without dedicated task-specific architectures, and that integrating understanding and generation can provide complementary benefits beyond pure scaling. Overall, our results indicate that unified multimodal models can surpass prior limitations in generation quality when equipped with stronger language-guided generation mechanisms. This enables a favorable balance between capability, efficiency, and deployment practicality.

\begin{table*}[t]
\centering
\small
\caption{Comparison with unified multimodal models. We evaluate across both understanding and generation benchmarks. Our model (LongCat-Next) consistently outperforms prior unified approaches across most generation benchmarks while maintaining highly competitive understanding capabilities, establishing a new state-of-the-art for unified generation-understanding models. All results are from public reports.}
\label{tab:omni_unified_results}
\vspace{1em}
\resizebox{\textwidth}{!}{
\begin{tabular}{l ccccc | cccccc}
\toprule
\multirow{2}{*}{\textbf{Model}} & \multicolumn{5}{c|}{\textbf{Understanding}} & \multicolumn{6}{c}{\textbf{Generation}} \\
\cmidrule(lr){2-6} \cmidrule(lr){7-12}
& \textbf{MMMU} & \textbf{MathVista} & \textbf{OCRBench}  & \textbf{DocVQA} & \textbf{MMStar} & \textbf{GenEval} & \textbf{DPG} & \textbf{LT-EN/ZH} & \textbf{WISE} & \textbf{TIFF} & \textbf{CVTG} \\
\midrule
Janus-Pro \cite{chen2025janus} & 41.0 & -- & -- & -- & -- & 80 & 84.19 & 1.90 / 0.60 & 35 & 65.50 / 68.80 & -- \\
Show-o2\cite{xie2025showo2improvednativeunified}  & 48.9 & -- & --  & -- & 56.6 & 76 & 86.14 & -- / -- & -- & 65.80 / 65.00 & -- \\
OneCAT \cite{li2025onecatdecoderonlyautoregressivemodel} & 41.9 & 61.7 & --  & 91.2 & -- & 90 & 84.53 & -- / -- & -- & -- & -- \\
Mogao \cite{liao2025mogaoomnifoundationmodel} & 44.2 & --  & -- & -- & -- & 89 & 84.33 & -- / -- & -- & -- & -- \\
BAGEL \cite{deng2025emergingpropertiesunifiedmultimodal} & 55.3 & 73.1 & 80.9  & 92.2 & 69.3 & 82 & 85.07 & 43.70 & 4.70 / 49 & 71.50 / 71.70 & 35.60 \\
NEO-unify \cite{diao2026pixelswordsnative} & 68.9 & -- & 81.5  & 91.6 & 65.5 & 85 & 86.71 & 91.40 / 75.50 & 42 & -- & -- \\
Ovis-U1\cite{wang2025ovisu1technicalreport} & 51.1 & 69.4 & 88.3  & -- & -- & 89 & 83.72 & 3.00 / 5.10 & 42 & 66.70 / 68.20 & 9.30 \\
Lumina\cite{xin2025luminadimooomnidiffusionlarge} & 58.6 & -- & -- & --  & -- & 88 & 86.04 & 43.70 / 4.70 & 40 & 74.70 / 72.00 & 59.00 \\
OmniGen2\cite{wu2025omnigen2explorationadvancedmultimodal} & 53.1 & -- & -- & --  & -- & 80 & 83.57 & 56.10 / 5.90 & -- & -- & -- \\
UniWorld-V1\cite{lin2025uniworldv1highresolutionsemanticencoders} &  58.6 & -- & --  & -- & -- & 80 & 81.38 & -- / -- & 55 & -- & -- \\
X-Omni \cite{geng2025xomni}& -- & -- & 70.4  & -- & -- & 83 & 87.65 & 90.00 / 81.40 & -- & -- & -- \\
InternVL-U\cite{tian2026internvludemocratizingunifiedmultimodal} & 54.7  & -- & 83.9 & -- & --  & 85 & 85.18 & 73.80 / 86.00 & 46 & 74.90 / 73.90 & 62.30 \\
BLIP3-o \cite{blip3oNext2025} & 50.6   & -- & -- & -- & -- & 84 & 81.60 & -- / -- & 62 & -- & -- \\
\midrule
LongCat-Next & 70.6 & 83.1 & 86.5  & 94.2 & 69.3 & 84 & 84.66 & 93.15 / 89.08 & 57 & 82.85  / 84.38 & 76.36  \\
\bottomrule
\end{tabular}
}
\end{table*}

\subsubsection{Audio}

We conduct a comprehensive evaluation of the audio capability of LongCat-Next, covering automatic speech recognition (ASR), text-to-speech (TTS), audio understanding, and audio-to-text chat.
ASR and TTS serve as foundational tasks for evaluating speech understanding and generation. To assess ASR performance, we utilize a range of benchmark datasets, including LibriSpeech~\cite{panayotov2015librispeech}, AISHELL-1~\cite{bu2017aishell}, AISHELL-2~\cite{du2018aishell}, FLEURS~\cite{conneau2023fleurs}, and WenetSpeech~\cite{zhang2022wenetspeech}. Meanwhile, for evaluation of TTS, both the Chinese and English versions of SeedTTS~\cite{anastassiou2024seed} are used to ensure comprehensive coverage. For ASR and TTS task, we report the WER metric for evaluation.
Beyond these tasks, audio understanding focuses on the model’s ability to perceive and reason about fine-grained acoustic information within input audio, such as speech, environmental sounds, music, speaker gender, age, accent, etc. Consequently, we adopt MMAU~\cite{sakshi2024mmau}~\footnote{MMAU v05.15.25 test-mini}, VocalSound~\cite{gong2022vocalsound}, TUT2017~\cite{mesaros2016tut},  and ClothoAQA~\cite{lipping2022clotho}as standard benchmarks for this purpose. Furthermore, to measure LongCat-Next’s performance in audio-instructed conversations, OpenAudioBench~\cite{li2025baichuan} is used as the evaluation benchmark, examining the model’s capabilities in world knowledge, mathematical proficiency, and reasoning ability. To ensure standardized and reproducible results, we employ the evaluation method released by Kimi-Audio-Evalkit~\footnote{\url{https://github.com/MoonshotAI/Kimi-Audio-Evalkit}}. We use \textit{gpt-4o-2024-08-06} as the judge model for tasks like audio question answering. Through multi-faceted evaluation approaches, we provide a thorough assessment of the model’s performance across diverse audio tasks.

\begin{table}[t]
\centering
\footnotesize
\setlength{\tabcolsep}{2pt}
\renewcommand{\arraystretch}{0.9}
\caption{Comparison of LongCat-Next and top-tier models on audio benchmarks.Values marked with * are sourced from public reports.}
\vspace{1em}
\label{tab:audio_res}
\begin{tabularx}{\textwidth}{l*{7}{>{\centering\arraybackslash}X}}
\toprule
\multirow{3}{*}{\textbf{Benchmark}} & \textbf{LongCat-Next} & \textbf{Gemini-3.1-Flash-Lite-preview} & \textbf{Gemini-2.5-Flash-Lite-preview} & \textbf{Qwen3-Omni-A3B-Instruct} & \textbf{MiMo-Audio} & \textbf{Kimi-Audio} & \textbf{Step-Audio-2-mini}\\
\midrule
\multicolumn{8}{c}{\textbf{ASR}} \\
\midrule
LibriSpeech$_\text{test-clean}$ $\downarrow$ &1.63&2.38&3.14&1.22$^*$&2.47&1.28$^*$&1.33$^*$ \\
LibriSpeech$_\text{test-other}$ $\downarrow$ &3.42&5.68&7.19&2.48$^*$&6.13&2.42$^*$& 2.86$^*$\\
AISHELL-1  $\downarrow$  & 1.47 & 6.00 & 11.64 &  0.84$^*$ & 1.69 & 0.60$^*$ & 0.78$^*$ \\
AISHELL-2  $\downarrow$     & 2.82 & 8.00 & 15.33 & 2.34$^*$ & 3.02 & 2.56$^*$ & 2.16$^*$ \\

Fleurs$_\text{zh}$ $\downarrow$ &3.24&3.88&7.77&2.20$^*$&9.29&2.69$^*$&2.53$^*$ \\
Fleurs$_\text{en}$ $\downarrow$ &5.24&5.96&6.84&2.72$^*$&6.33&4.44$^*$& 3.05$^*$\\

WenetSpeech$_\text{test-meeting}$ $\downarrow$ &8.19 &20.37 &23.04&5.89$^*$&9.49&6.28$^*$& 4.87$^*$\\
WenetSpeech$_\text{test-net}$ $\downarrow$ &5.98&16.15&24.83&4.69$^*$&7.64&5.37$^*$&4.82$^*$ \\
\midrule
\multicolumn{8}{c}{\textbf{TTS}} \\
\midrule

SeedTTS$_\text{zh}$ $\downarrow$& 1.90 & - & - & 1.07$^*$ & 1.96$^*$ & 13.46 & 2.13$^*$ \\
SeedTTS$_\text{en}$ $\downarrow$& 1.89 & - & - & 1.39$^*$ & 5.37$^*$ & 29.45 & 3.18$^*$ \\
\midrule

\multicolumn{8}{c}{\textbf{Audio Understanding}} \\
\midrule
MMAU         & 76.40 & 71.70 & 74.80 & 78.20 & 75.80 & 70.31 & 71.30 \\
ClothoAQA    & 73.45 & 60.53 & 63.27 & 75.16$^*$ & 68.85 & 72.21$^*$ & 68.39$^*$ \\
TUT2017      & 43.09 & 23.15 & 22.47 & 40.74$^*$ & 15.06 & 65.25$^*$ & 30.67$^*$ \\
VocalSound   & 85.91 & 69.81 & 57.67 & 91.59$^*$ & 87.94 & 94.85$^*$ & 87.58$^*$ \\
\midrule
\multicolumn{8}{c}{\textbf{Audio-to-Text Chat}} \\
\midrule
AlpacaEval     & 86.83 & 62.56 & 89.05 & 90.10 & 85.67 & 78.74 & 53.06 \\
LlamaQuestions & 79.67 & 82.67 & 83.33 & 82.33 & 75.00 & 79.33 & 69.70 \\
ReasoningQA    & 87.52 & 65.64 & 74.06 & 87.62 & 75.34 & 68.61 & 63.86 \\
TriviaQA       & 67.60 & 68.40 & 66.90 & 76.60 & 50.30 & 62.10 & 45.30\\
Webquestions   & 69.10 & 69.10 & 70.00 & 75.90 & 61.30 & 70.30 & 56.90 \\
\bottomrule
\end{tabularx}
\end{table}

Table~\ref{tab:audio_res} summarizes the evaluation results of audio capabilities in comparison with other competitors, including Gemini-3.1-Flash-Lite-preview, Gemini-2.5-Flash-Lite-preview~\cite{comanici2025gemini25pushingfrontier}, Qwen3-Omni-A3B-Instruct~\cite{xu2025qwen3}, MiMo-Audio~\cite{zhang2025mimo}, Kimi-Audio~\cite{ding2025kimi}, and Step-Audio-2-mini~\cite{wu2025step}. LongCat-Next demonstrates impressive fundamental audio capabilities, excelling at both audio recognition and synthesis. Notably, its performance in text-to-speech synthesis surpasses most compared models in terms of accuracy. Moreover, LongCat-Next achieves state-of-the-art performance across a wide range of audio comprehension tasks. These findings suggest that LongCat-Next possesses advanced capabilities in processing and understanding both general and complex acoustic information, far surpassing traditional speech recognition systems. Additionally, LongCat-Next shows strong results on audio-instructed question-answering benchmarks, highlighting its extensive world knowledge and advanced reasoning abilities.

\subsubsection{Text}
\begin{table}[t]
    \centering
    \small
    \caption{Comparison between \longcat and other models. Values marked with * are sourced from public reports.}
    \label{tab:chat_model_results}
    \vspace{1em}
    \begin{tabularx}{\textwidth}{l c*{5}{>{\raggedright\arraybackslash}X}}
    \toprule
    \multirow{2}{*}{\textbf{Benchmark}} & \textbf{\longcat} & \textbf{Kimi-Linear-48B-A3B} & \textbf{Qwen3-Next-80B-A3B-Instruct} & \textbf{Qwen3-Omni-A3B-Instruct} \\
        \midrule
        \multicolumn{5}{c}{\textbf{Agentic Tool Use}} \\
        \midrule
        Tau2-Airline(avg@8) & {56.50}& 44.00 & 45.5* & 27.00  \\
        Tau2-Retail(avg@8) & {73.68}& 18.86 & 57.3* & 40.80  \\
        Tau2-Telecom(avg@8) & {62.06} & 15.68 & 13.2* & 4.39  \\
        VitaBench(avg@4) & {5.80} & - & {5.80} & - \\
        \midrule
        \multicolumn{5}{c}{\textbf{Agentic Coding}} \\
        \midrule
        SWE-Bench(acc) & {43.00} & 32.80 & 37.60 & - \\
        TerminalBench(acc) & {18.75} & {20.00} & 15.19 & - \\
        \midrule
        \multicolumn{5}{c}{\textbf{General Domains}} \\
        \midrule
        MMLU(acc) & 83.95 & 79.91 & {89.28} & 87.10 \\
        MMLU-Pro(acc) & 77.02 & 67.22 & {82.93} & 79.89 \\
        CEval(acc) & 86.80 & 78.48 & {90.91} & 88.50   \\
        CMMLU(acc) & 82.13 & 76.26 & {86.50} & 85.76   \\
        \bottomrule
    \end{tabularx}
    
\end{table}

Table~\ref{tab:chat_model_results} presents the foundational text-modality evaluation of \longcat. To comprehensively assess its pure-language cognitive abilities—which are crucial for supporting complex cross-modal reasoning—our evaluation framework is structured around three pivotal dimensions: agentic tool use, coding capabilities, and general-domain knowledge. The results demonstrate that \longcat establishes a formidable baseline, particularly excelling in practical execution. \textbf{(1) Excellence in Agentic Workflows and Coding.} We measure the model's capacity for environmental interaction and complex software engineering via Vita Bench~\cite{he2025vitabench}, a curated noise-reduced $\tau^2$-Bench\footnote{\url{https://github.com/AGI-Eval-Official/tau2-bench-revised}}, SWE-Bench~\cite{jimenez2023swe}, and TerminalBench~\cite{merrill2026terminalbenchbenchmarkingagentshard}. \longcat exhibits a distinct and overwhelming advantage in these scenarios. In tool-use evaluations, it establishes a clear lead across all $\tau^2$-Bench sub-scenarios, most notably achieving an exceptional 62.06 in the Telecom domain, completely eclipsing Qwen3-Omni-A3B-Instruct (4.39) and Kimi-Linear-48B (15.68). This proficiency extends to coding, where it achieves a 43.0 accuracy on the highly challenging SWE-Bench, significantly outperforming both Kimi-Linear-48B and Qwen3-Next. These results underscore its superior capability in dynamic environment navigation, complex tool dependency resolution, and real-world codebase manipulation. \textbf{(2)Robust Foundational Knowledge and Reasoning.} Beyond execution, \longcat successfully mitigates the ``multimodal tax''—a common performance degradation in text capabilities when scaling non-linguistic modalities. Broad multi-disciplinary knowledge and reasoning are tested across standard encyclopedic and mathematical benchmarks, including MMLU~\cite{hendrycks2021measuringmassivemultitasklanguage}, MMLU-Pro~\cite{wang2024mmluprorobustchallengingmultitask}, C-Eval~\cite{huang2023ceval}, CMMLU~\cite{li2023cmmlu}. \longcat consistently outperforms Kimi-Linear-48B and remains highly competitive with the multimodal Qwen3-Omni-A3B-Instruct. While the strictly text-optimized Qwen3-Next-80B-Instruct naturally sets the upper bound in these traditional academic tests, \longcat maintains a highly resilient cognitive baseline, ensuring its multimodal outputs are anchored by deep logical reasoning.

\subsection{Experimental Analysis of Methodology}
\label{sec:model_analysis}

We conduct ablation studies to analyze the key components of our method and the motivations behind their design. Due to the computational cost of full-scale experiments, some studies are conducted under a reduced setting using Qwen-7B as the language backbone. The detailed experiments are described below.

\subsubsection{Bridging the Understanding Gap Between Discrete and Continuous Modeling}

\begin{figure}[t]
     \centering
    \begin{minipage}[c]{0.38\textwidth}
        \centering
        \includegraphics[width=\linewidth]{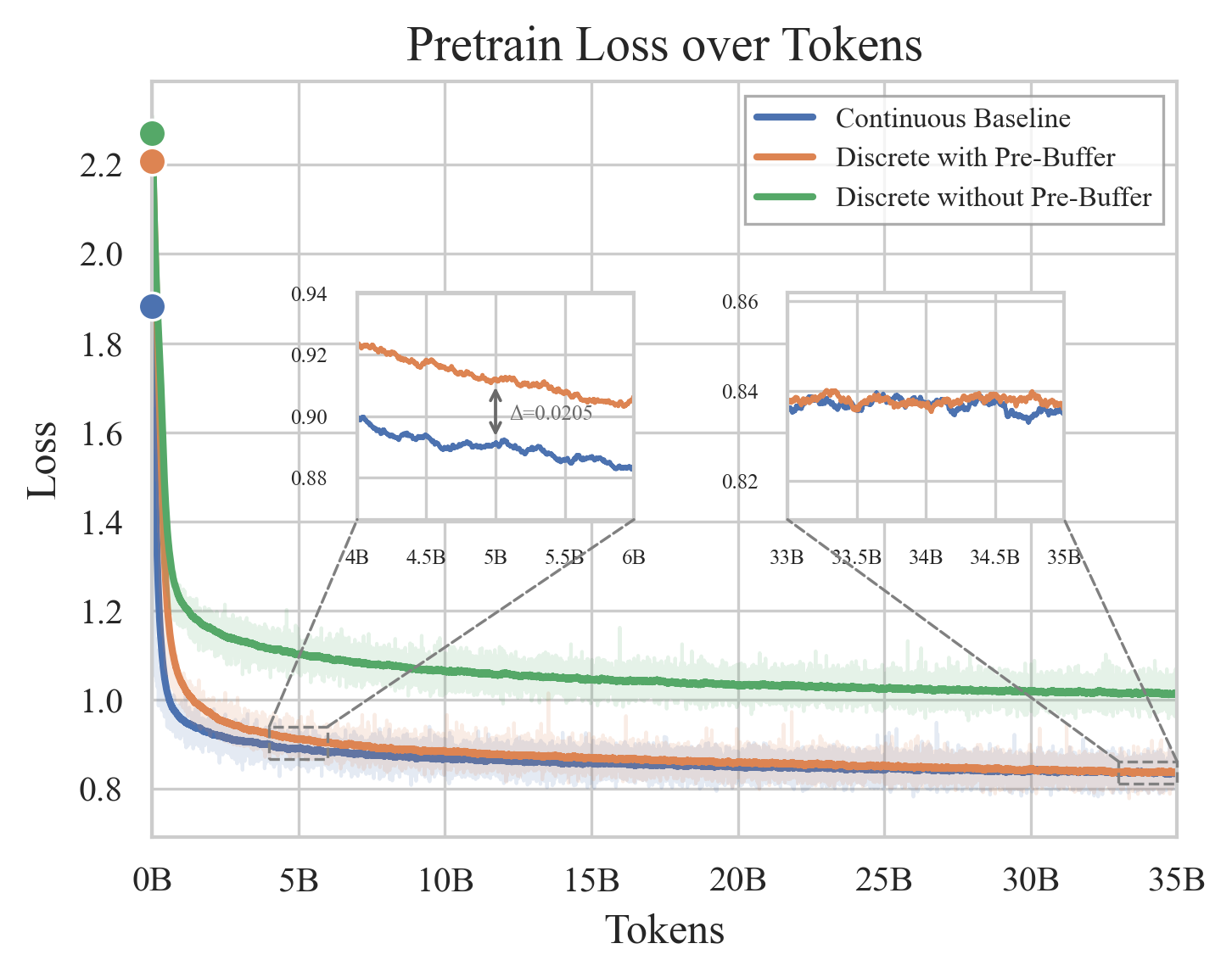}
    \end{minipage}
    \hfill
    \begin{minipage}[c]{0.6\textwidth}
        \centering
        \scriptsize
        \renewcommand{\arraystretch}{1.0}
        \setlength{\tabcolsep}{3pt}
        \begin{tabular}{l|ccccc|cc}
        \toprule
        & \multicolumn{5}{c|}{\textbf{Partial Data}} & \multicolumn{2}{c}{\textbf{Scale Up Data}} \\
        \midrule
        \textbf{Experiment ID} & \textbf{I} & \textbf{II} & \textbf{III} & \textbf{IV} & \textbf{V} & \textbf{VI} & \textbf{VII} \\
        \midrule
        Rep. Type & Continuous & Discrete & Continuous & Discrete & Discrete & Continuous & Discrete \\
        PT Tokens & 0.1B & 0.1B & 5B & 5B & 5B & 5B & 5B \\
        MT\&SFT Tokens & 4B & 4B & 4B & 4B & 4B & 300B & 300B \\
        Pre-Buffer & - & \cmark & - & \cmark & \xmark & - & \cmark \\
        \midrule
        \multicolumn{8}{c}{\textbf{OCR \& Doc \& Chart}} \\
        \midrule
        OCRBench & 779 & 598 & 776 & 720 & 705 & 858 & 844\\
        DocVQA & 88.2 & 78.0 & 88.9 & 86.7 & 85.2 & 96.0 & 96.0 \\
        ChartQA & 80.1 & 71.6 & 80.6 & 76.8 & 75.4 & 84.0 & 84.0 \\
        \midrule
        \multicolumn{8}{c}{\textbf{STEM \& Reasoning}} \\
        \midrule
        MMMU$_{\text{val}}$ & 49.8 & 44.8 & 49.1 & 48.0 & 49.8 & 58.0 & 60.0 \\
        MathVista & 59.6 & 47.3 & 56.1 & 54.2 & 56.7 & 75.0 & 74.0 \\
        \bottomrule
        \end{tabular}
    \end{minipage}
    \vspace{1em}
    \caption{L: Loss comparison on pre-align stage. R: Performance comparison of discrete and continuous versions.}
    \label{fig_tab:combined_optimized_layout}
    \vspace{-5pt}
\end{figure}

\textbf{Modeling Analysis.} As discussed in Sec.\ref{sec:visual_tokenizer}, the SAE coupled with RVQ formulation effectively addresses the challenge of semantic completeness. Building upon this foundation, we further investigate the extent to which discrete representations can approximate their continuous counterparts, and accordingly design a pre-alignment experiment. In this setting, all modules, except for the frozen language backbone and vision encoders (i.e., discrete dNaViT vs.\ continuous NaViT), are optimized. During this stage, we employ a captioning loss as a proxy to quantify the information loss introduced by discretization relative to the continuous version.

As depicted in Fig. \ref{fig_tab:combined_optimized_layout}, our empirical analysis yields three key insights: 
(1) the initial training loss is significantly higher than that of the continuous baseline. This indicates that continuous features are inherently easier to align at the early stage. Although the gap narrows as training progresses, a noticeable performance discrepancy (Exp I and II) remains under the vanilla discrete formulation. 
(2) Pre-Buffer. We hypothesize that this gap stems from insufficient re-encoding after the sum-up operation on multi-level embeddings. To address this, we introduce a lightweight \textit{Pre-Buffer} module (implemented as a single-layer FFN) to remaps the visual representations after codebook lookup and re-encodes the multi-level summed features. Despite its simplicity, this module substantially accelerates convergence and improves the expressiveness of the discrete tokens.
(3) Longer-Training. Unlike the continuous setting, discrete visual embeddings are learned entirely from scratch. As a result, they require more data to reach comparable performance. Comparisons across experiments (e.g., Exp I vs. II and Exp IV vs. VII) show that the performance gap can be largely reduced through extended training and increased data scale.

\textbf{Closing the Understanding Gap with Continuous Models.} Tab. \ref{fig_tab:combined_optimized_layout} reveals that dNaViT not only achieves comparable training loss but also maintains downstream performance within approximately a 1\% margin of the continuous baseline. Although the discrete model initially underperforms under limited training data, scaling the data progressively closes this gap, with its loss asymptotically approaching that of the continuous counterpart. For instance, Exp VI and VII demonstrate that the discrete formulation achieves near-parity with the continuous model. This conclusion is further supported by large-scale experiments on LongCat-Next, which achieve performance competitive with specialized continuous models such as Qwen3-VL-A3B-Instruct. These results indicate that discrete modeling does not have an inherent performance ceiling. Instead, its performance is primarily influenced by the training data. With the appropriate data scaling and improved quality, the full potential of discrete representations can be unlocked.

\subsubsection{Information Recovery Analysis} 
\label{information_recovery}

We provide a previously underexplored perspective: the reconstruction capability of semantic encoders is not solely determined by supervision, but is also intrinsically linked to architectural properties such as residual connections. As evidenced in Fig.~\ref{fig:reconstruction_visuals} and Tab.~\ref{tab:reconstruction_metrics}, all decoders are paired with a lightweight ViT-based decoder to isolate encoder characteristics. Semantic encoders with residual pathways exhibit non-trivial reconstruction capability despite the absence of explicit reconstruction supervision (e.g., ResNet50). In contrast, QwenViT without the merger module achieves comparable reconstruction performance, while introducing the merger leads to noticeable degradation due to the aggressive downsampling ($14\times \rightarrow 28\times$), as shown in Fig.~\ref{fig:reconstruction_visuals}. Qualitative visualizations further suggest that such SAE-style encoders (e.g., QwenViT) still retain the ability to recover coarse image-level structures, but are only less effective at reconstructing fine-grained, high-frequency details. This provides an insightful perspective for rethinking the trade-off between pixel-level and semantic-level information.

\begin{table}[htbp]
    \centering
    \footnotesize 
    \caption{Quantitative reconstruction performance across various visual encoder architectures. PSNR and SSIM represent reconstruction fidelity ($\uparrow$), while rFID measures perceptual discrepancy ($\downarrow$).}
    \vspace{1em}
    \label{tab:reconstruction_metrics}
    \setlength{\tabcolsep}{6pt} 
    \renewcommand{\arraystretch}{1.05} 
    \begin{tabular}{l cccc}
        \toprule
        \textbf{Metric} & ResNet50 & {ViT-B/16 (Pretrained)} & {ViT-B/16 (Random)} & {QwenViT (w/o merger)}  \\
        \midrule
        PSNR ($\uparrow$) & $20.88 \pm 3.44$ & $21.86 \pm 3.14$ & $30.52 \pm 3.42$ & $18.16 \pm 2.61$ \\
        
        SSIM ($\uparrow$) & $0.509 \pm 0.174$ & $0.581 \pm 0.139$ & $0.887 \pm 0.051$ & $0.46 \pm 0.14$  \\
        
        rFID ($\downarrow$) & 0.4619 & 0.8850 & 0.5847 & 0.987 \\

        \bottomrule
    \end{tabular}
\end{table}

Interestingly, the vanilla ViT with randomly initialized weights achieves the best reconstruction performance, yielding the highest PSNR among the compared models. A possible explanation is that the outputs of a randomly initialized ViT tend to resemble noise-like signals, which are easier for the decoder to denoise during reconstruction. Meanwhile, the residual pathways in the architecture may still preserve a portion of the original pixel-level information.

\begin{figure}[h]
    \centering
    \includegraphics[width=1.0\linewidth]{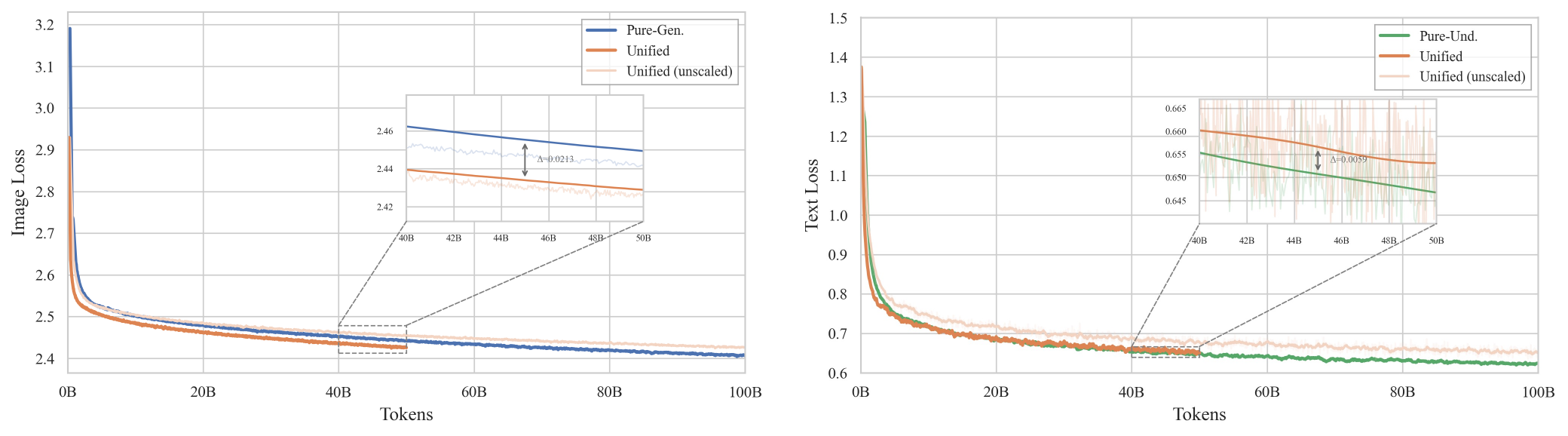}
    \caption{Visual understanding and generation interaction under DiNA framework.}
    \label{fig:und_and_gen}
\end{figure}

\subsubsection{Conflict between Understanding and Genearation}

Given that DiNA integrates understanding and generation through a unified autoregressive objective, we designed a set of illuminating experiments to investigate the conflicting or synergistic relationship between these two tasks. All training was conducted under identical experimental settings using the same model checkpoint. Specifically, we trained Pure-Und. and Pure-Gen. models using 100B tokens of understanding and generation data, respectively. Additionally, we trained a Unified model on a 1:1 mixture comprising 50B tokens from each dataset. Since the Unified model receives only half the task-specific data compared to the pure models under the same token budget, we proportionally scale its loss curve while retaining the original unscaled version for reference. Fig.~\ref{fig:und_and_gen} demonstrates that under comparable token counts, the Unified model exhibits a marginal loss difference of 0.006 compared to the Pure-Und. model, while achieving a loss of 0.02 lower than that of the Pure-Gen. model. This suggests that generation does not compromise understanding, whereas understanding actively enhances generation.

\subsubsection{Semantic Comparison of Parallel and Serial Audio Generation}
In parallel internal text-guided generation, each decoding step simultaneously generates both text and audio outputs. This mixed multimodal representation introduces considerable challenges in achieving semantically accurate responses.  In contrast, serial generation avoids this issue, inherently offering stronger semantic consistency by sequentially generating outputs.
To this end, we propose a random delay-based unified modeling paradigm for internal language guidance. Leveraging this approach, LongCat-Next adaptively learns to align audio and text semantics within the context, significantly reducing discrepancies between the two generation strategies.

To validate the effectiveness of our method, we systematically evaluate the text guidance accuracy on the LlamaQuestions and ReasoningQA benchmarks in an audio-to-audio manner. The results show that parallel generation achieves performance comparable to serial generation, with 79.33 vs. 81.67 on LlamaQuestions and 74.95 vs. 80.30 on ReasoningQA. Despite a slight gap, the parallel approach maintains strong semantic fidelity, demonstrating that it can preserve response quality while offering improved efficiency.

\subsubsection{On the Training Dynamics of Modality-Agnostic MoE}
\label{moe_label}

\begin{figure}[h]
    \centering
    \includegraphics[width=0.8\linewidth]{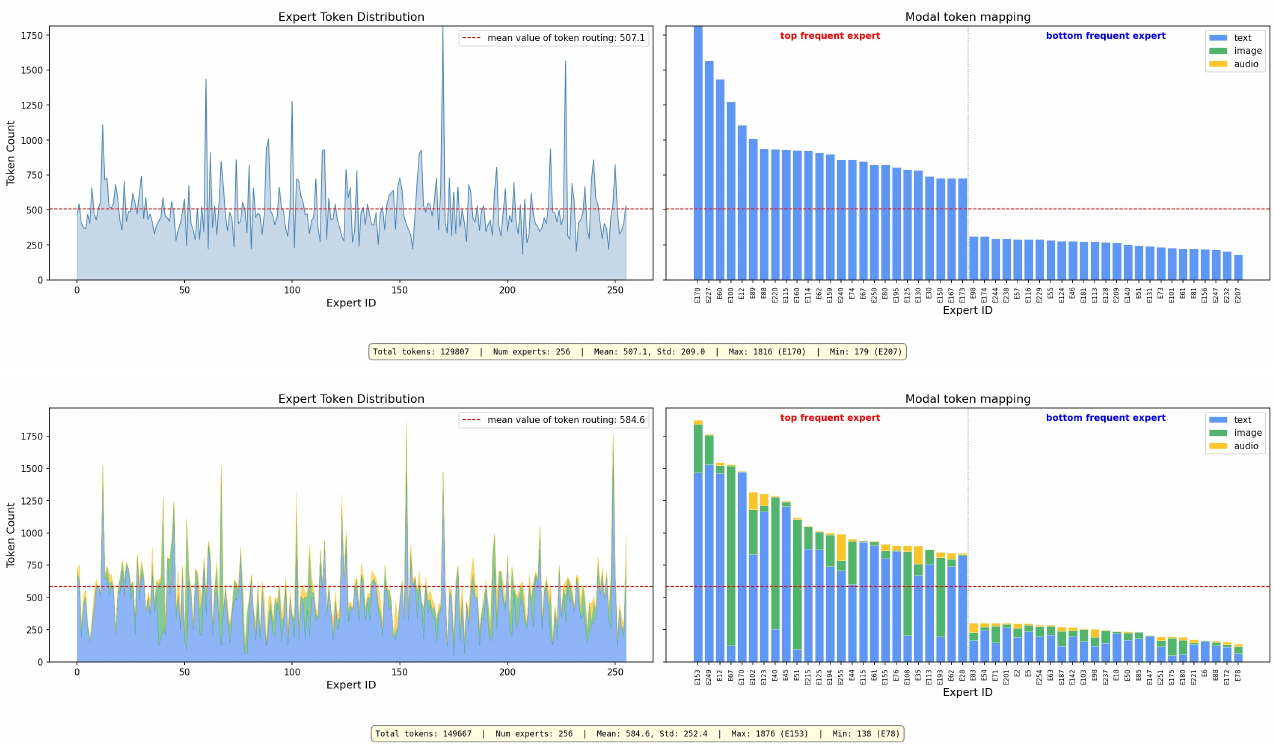}
    \caption{Training dynamics of modality-agnostic MoE before and after multimodal training. Left: token distribution across experts. Right: the emergence of modality-specialized roles within an initially modality-agnostic MoE.}
    \label{fig:moe_dynamic}
\end{figure}

We observe that as a language model evolves into a native multimodal architecture, the internal dynamics of the Mixture-of-Experts (MoE) layers undergo notable shifts, as illustrated in Fig.~\ref{fig:moe_dynamic}. To better understand this transition, we analyze the intermediate MoE layers of LongCat-Next, comparing the pure text model with its natively trained multimodal counterpart.

First, even though the MoE design is modality-agnostic, a subset of experts gradually becomes functionally specialized, exhibiting clear preferences for specific modalities such as vision or audio. Second, the routing mechanism becomes more structured, with routers developing increasingly distinct and stable selection patterns over experts. Moreover, after introducing multimodal training, the average number of routed tokens per expert increases from 507.1 to 584.6, indicating that a larger portion of the expert capacity is being utilized. This suggests that multimodal learning not only induces functional differentiation but also effectively expands the model’s capacity usage. Together, these results suggest that native multimodal training induces both functional specialization and more efficient use of model capacity within the MoE architecture.

\subsubsection{Platonic Representation Hypothesis}

\begin{figure}[t]
    \centering
    \includegraphics[width=0.9\linewidth]{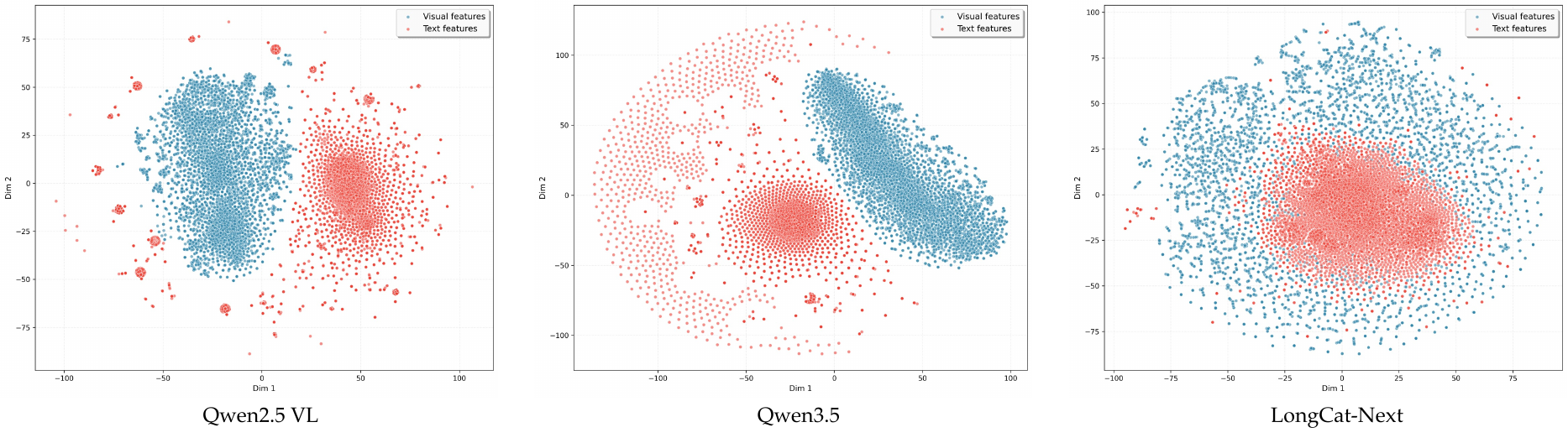}
    \caption{Modality-aware feature distribution divergence: a comparison of traditional Qwen2.5-VL, data-native-training Qwen3.5, and our architectural native modality with LongCat-Next.}
    \label{fig:tsne}
\end{figure}

We hypothesize that, under such a native discrete token space, modality tokens should complement textual tokens and form an interlaced distribution in the embedding space, rather than modality-specific clusters. This intuition is related to a \textit{Platonic representation perspective}~\cite{huh2024platonic}, where text and vision can be viewed as different expressions of the same underlying reality, and thus should ideally share a common semantic space.

Empirically, analysis of token-level representations after training shows that LongCat-Next, trained under the DiNA paradigm, naturally exhibits interwoven embeddings across visual and textual tokens. Specifically, we use image–text inputs and visualize the joint embedding space of vision and text tokens via t-SNE with 50,000 sample points, as shown in Fig.~\ref{fig:tsne}. In contrast, the non-native Qwen2.5-VL produces largely separated modality clusters, while the natively trained Qwen3.5 exhibits only partial cross-modal alignment. Notably, LongCat-Next demonstrates a stronger ability to internalize different modalities and align them within a shared space.

These results suggest that a native discrete multimodal design more effectively promotes a unified semantic space, where features from different modalities behave like \textit{multilingual} expressions of the same underlying concepts. An additional observation is that the quantized mapping of the visual tokenizer, although entirely frozen prior to integration with the language backbone, exhibits negligible performance degradation when adapted to the LongCat architecture. This indicates that the discrete semantic space is inherently aligned with the language representation space, enabling seamless cross-modal integration.

\section{Implementation Details}
\label{sec:implementation_details}

\subsection{Training Stage}

\begin{figure}[htbp] 
    \centering
    
    \begin{minipage}[c]{0.4\textwidth} 
        \centering
        \includegraphics[width=\textwidth]{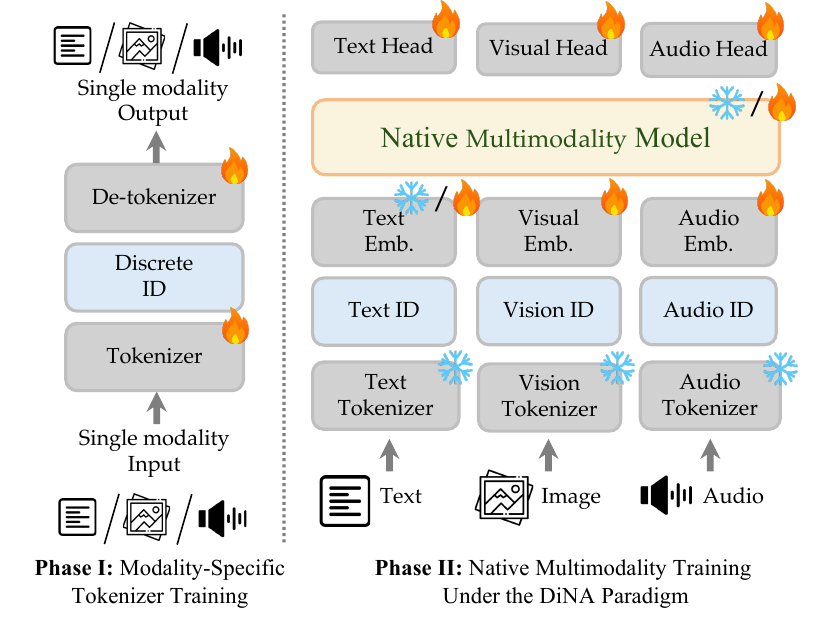} 
        \caption{Training phases of LongCat-Next.}
        \label{fig:training_stage}
    \end{minipage}%
    \hfill 
    \begin{minipage}[c]{0.57\textwidth} 
        \centering
        \footnotesize
        \captionof{table}{Detailed training stages, configurations, and data for \textbf{V (Visual)}, \textbf{A (Audio)}, and \textbf{T (Text)} Modalities.}
        \label{tab:training_details}
        
        \renewcommand{\arraystretch}{1.2} 
        \setlength{\tabcolsep}{3pt}
        
        \resizebox{\textwidth}{!}{%
        \begin{tabular}{@{}l | c c c c@{}}
            \toprule
            \textbf{Stages} & \textbf{Stage 1} & \textbf{Stage 2} & \textbf{Stage 3} & \textbf{Stage 4} \\
            \midrule
            
            \textbf{Purpose} & 
            \makecell[c]{Pre-Align} & 
            \makecell[c]{Pre-training} & 
            \makecell[c]{Mid-training} & 
            \makecell[c]{SFT} \\
            \midrule
            
            \textbf{Dataset} & 
            \makecell[c]{\textbf{V:} Image-Caption Pairs, \\ \textbf{A:} Pure Audio, ASR, \\ TTS, Interleaved Data} & 
            \makecell[c]{+\\ \textbf{V:} Interleaved Data,\\ OCR, Grounding, \\ FixRes. Generation \\ \textbf{A:} Same as Last Stage\\ \textbf{T:} Pure Text} & 
            \makecell[c]{+\\ \textbf{V:} Long CoT,\\Genaral QA,\\Video, GUI\\AnyRes. Generation\\ \textbf{A:} High-quality Data \\ Multi-turn Dialogue \\ \textbf{T:} Same as Last Stage} & 
            \makecell[c]{\\ \textbf{V:} High-quality\\Instruct,\\ Reasnoing,\\AnyRes. Generation \\ 
            \textbf{A:} Instruction Data\\
            \textbf{T:} Long Pure Text} \\
            \midrule
            
            \textbf{Batch Size} & 8192 & 8192 & 1024 & 128 \\
            
            \textbf{Seq. Len} & 8K & 8K & 32K & 64K \\
            
            \textbf{Trainable} & \makecell[c]{Embed.,\\DepthT.} & All & All & All \\
            
            \bottomrule
        \end{tabular}%
        } 
    \end{minipage}
    
\end{figure}

The optimization of DiNA can be concisely divided into two phases: modality-specific tokenizer training and unified multimodal training, as illustrated in Fig.\ref{fig:training_stage}. (1) \textit{Tokenizer Training:} The process begins with the independent training of modality-specific tokenizers and detokenizers to establish a well-structured discrete representation space. (2) \textit{Native Multimodality Training:} This phase starts with a \textit{Pre-align} stage, in which the codebook embeddings and DepthTransformer decoders are warmed up while the language backbone remains frozen, facilitating alignment between discretized tokens and the backbone model. Subsequently, the entire framework, comprising the language backbone, all modality embeddings, and decoders, is unfrozen for end-to-end training, while the tokenizers remain fixed. This stage encompasses \textit{Pre-train}, \textit{Mid-training}, and \textit{SFT}. Such a unified training paradigm enables seamless integration of multimodal understanding and generation, allowing the model to function as a single, coherent system across modalities.

\subsubsection{Visual Tokenizer}

The methodology for constructing a semantically complete dNaViT tokenizer is described in Sec.~\ref{sec:visual_tokenizer}. In this work, we focus on the development of the dNaViT tokenizer itself, bypassing the computationally demanding SAE training phase, which typically involves large-scale vision–language pretraining. Instead, we directly adopt Qwen2.5-ViT~\cite{qwenvl} (with a 28× spatial compression ratio) as the encoder for our experiments, concentrating on the tokenization and de-tokenization processes. While a better optimized SAE could potentially yield further improvements, we find this choice sufficient for the target of current version. In the following, we detail how dNaViT is trained to support tokenization and de-tokenization at arbitrary resolutions.

\begin{itemize}
\item
\textbf{Stage 1: Visual Tokenization} As described in Sec~\ref{sec:tokenization}, this stage maps the continuous manifold of SAE features into a discrete latent space. We employ RVQ to minimize quantization error for converting dense visual signals into discrete token IDs.
To support arbitrary-resolution encoding and decoding, we adopt a flexible sequence-based modeling paradigm~\cite{packnpack}, where images of varying resolutions are flattened into a single sequence and efficiently processed using variable-length FlashAttention~\cite{dao2022flashattention,dao2023flashattention}. Training could be conducted in two phases: an initial fixed-resolution stage for fast convergence, followed by the any-resolution training with RVQ to adapt the quantization process to variable token lengths, where the maximum training sequence length is set to 8192. The training corpus is designed to support both understanding and generation, comprising a diverse collection of images from LAION~\cite{schuhmann2022laion}, COYO~\cite{kakaobrain2022coyo-700m}, DataComp~\cite{gadre2023datacomp}, and TextAtlas~\cite{wang2025textatlas5m}, along with a subset of in-house visual understanding dataset. To further enhance generation quality, we incorporate high-fidelity synthetic data (e.g., MidJourney). Training is conducted on approximately 50M images at arbitrary resolutions until convergence, with the maximum image resolution set to 1,736$\times$ 1,736. Empirically, we find that the resulting discrete representations effectively preserve both discriminative and generative information, making them well-suited for autoregressive modeling.
\item
\textbf{Stage 2: Visual De-tokenization}. Once the discrete codebook is established, we proceed to train the de-tokenizer, as described in \Cref{subsubsection:de-tokenization}, to reconstruct pixel-level images from the discrete token IDs. The pixel decoder is a 400M-parameter Vision Transformer~\cite{dosovitskiy2020image, QLIP} trained from scratch. It receives the projected discrete code embeddings, which are initially processed by a learnable MLP-based patch unmerger that reverses the spatial merging applied by the SAE encoder, thereby restoring the original patch sequence. The unmerged features are subsequently passed through a stack of transformer layers with 2D RoPE positional embeddings, and a linear head is employed to project the final hidden states to pixel space.
To further enhance perceptual sharpness and high-frequency detail, we train an image refiner initialized from OmniGen2~\cite{wu2025omnigen2explorationadvancedmultimodal} with continued training using flow-matching loss. The refiner receives two forms of conditioning: (1) the pixel decoder's reconstruction, concatenated with the noise latent along the channel dimension for spatial guidance, and (2) the projected discrete code embeddings for semantic conditioning. 
Overall, the training data for this stage reuses the same image corpus as Stage 1, supplemented with the data from SAM-1B~\cite{kirillov2023segment}, RenderedText~\cite{wendler2024renderedtext}, IDL~\cite{biten2022ocr} and a collection of high-resolution in-house images. Both the SAE encoder and the codebook remain frozen throughout this stage, and the de-tokenizer is trained at native resolution until convergence.
\end{itemize}

\subsubsection{Audio Tokenizer}

Leveraging the scaling laws~\cite{parker2024scaling} observed in audio tokenizers, we initialize our encoder from the pre-trained Whisper-large-v3~\cite{radford2022whisper} and further train it on our tasks. The resulting model demonstrates strong semantic comprehension of speech, environmental sounds, and music, while maintaining robust acoustic reconstruction fidelity.

The training corpus consists of two parts: (i) a large-scale collection of web-collected Chinese and English speech data, which has been cleaned and automatically transcribed using open-source ASR models; and (ii) a curated dataset consisting of high-quality multilingual and dialectal ASR data, in-house synthetic speech, and music/sound captioning datasets. The training corpus totals approximately 2.5 million hours.

The audio tokenizer is trained in three stages:

\begin{itemize}

\item \textbf{Stage 1: Decoder Warm-up.} The encoder and LLM are initialized with Whisper-large-v3 and Qwen3-1.7B, respectively, and the decoder is randomly initialized. During this stage, the encoder and LLM remain frozen, and the decoder is trained on the Mel spectrogram reconstruction task.

    \item \textbf{Stage 2: Semantic-Acoustic Joint Training.} In this stage, all modules are updated except the LLM and flow-matching module. The training objective combines all three losses described above. The RVQ module is also enabled. It consists of 8 layers, with codebook sizes of 8k, 4k, 2k, 1k, 1k, 1k, 1k, and 1k, respectively.

    \item \textbf{Stage 3: Decoder Fine-tuning.}.This stage aims to further improve the decoder's audio reconstruction capability using high-quality data. We collect a substantial volume of 24kHz high-quality audio. The decoder is reconfigured as a Diffusion Transformer (DiT) module to denoise the artifacts introduced by the RVQ. This refinement yields 24kHz Mel-spectrogram representations that are well-adapted for downstream vocoders.
\end{itemize}

\subsubsection{Native Multimodality Model}

The training pipeline consists of four stages: warm-up, full-modality pre-training, mid-training, and SFT (with $\sim$2 trillion training tokens in total). Pre-training leverages diverse multimodal data sources. During mid-training, we introduce additional synthetic data and high-quality curated datasets; the understanding branch is enhanced with long chain-of-thought reasoning, while the visual generation branch incorporates arbitrary-resolution feature training. Finally, the SFT stage further improves the model’s instruction-following capability.

\subsection{Visual Understanding}

We collect a large-scale and diverse multimodal corpus  for visual-language understanding, which primarily consists of the following components: Image Caption Data, Interleaved Image-Text Data, OCR data, Grounding data, STEM data, and GUI data. Each component is carefully curated to ensure broad coverage and high data quality, enabling the model to develop robust multimodal comprehension capabilities across a wide range of tasks and domains.

\subsubsection{Data Collection}

\paragraph{Image Caption Data} 
To facilitate robust vision-language alignment, we curate a large-scale, high-quality image-text dataset through a three-stage cleaning pipeline. First, we apply heuristic filtering to remove corrupted files, abnormal resolutions, and low-quality images. Second, to mitigate web data noise, we leverage multiple open-source LVLMs for recaption, transforming sparse labels into detailed, semantically rich descriptions. Finally, we use a SigLIP-based similarity threshold to prune misaligned pairs, retaining only highly accurate matches. This pipeline yields a balanced and diverse dataset, providing a semantically grounded foundation for dNaViT’s unified discrete modeling.

\paragraph{Interleaved Image-Text Data}
The interleaved image-text data consists of two components to enhance multimodal understanding. The first includes image-text pairs from open web pages, filtered using CLIP scores to remove noisy and low-alignment pairs. The second comes from video data, segmented into scenes with key frames and textual content extracted via ASR and OCR. These are organized into interleaved token sequences for fine-grained temporal and cross-modal learning. Together, they provide diverse data, strengthening vision-language alignment and reasoning.

\paragraph{Optical Character Recognition (OCR)}

Fine-grained visual perception is fundamental to multimodal large language models(MLLMs), enabling complex tasks like STEM reasoning and GUI interaction.

To this end, we constructed a massive OCR dataset comprising 75\% in-house synthetic data and 25\% filtered real-world data from over 90 open-source datasets. This hybrid approach ensures cross-scenario robustness and instruction-following proficiency. Domain-specific OCR data construction strategies are as follows:

\begin{itemize}

\item \textbf{{Document.}} Utilizing in-house data synthesis tools, paragraphs, formulas, and diverse text-image layouts were converted into structured Markdown formats, while tabular data was meticulously annotated in HTML format.

\item \textbf{{Chart \& Infographic.}} 

A dataset of 4.2M chart-to-code/table/JSON tuples was synthesized from arXiv to improve structural and numerical extraction. Additionally, a heterogeneous corpus of web-crawled infographics, presentations, and flowcharts was annotated to facilitate the parsing of intricate visual structures.

\item \textbf{{Scene \& Synthetic Text.}} 

Natural scene images from ST-VQA~\cite{biten2019scene} and filtered Common Crawl were rigorously decontaminated and re-annotated. To mitigate linguistic priors and hallucinations, we introduced out-of-vocabulary (OOV) text combinations. Furthermore, specialized tools (e.g., SynthDoG~\cite{kim2022ocr}, \LaTeX) were leveraged to generate challenging cases—including artistic fonts, handwriting, and distorted surfaces—to bolster OCR robustness.
\end{itemize}

This data construction pipeline ensures precise character-level alignment across diverse domains. While pre-training establishes foundational perception, the SFT stage refines the model into a functional OCR agent by enhancing instruction-following and structured output capabilities. We curated task-oriented queries from high-quality datasets (e.g., IDL-WDS~\cite{biten2022ocr}, ChartGalaxy~\cite{li2025chartgalaxy}, WildReceipt~\cite{sun2021spatial}, Hiertext~\cite{long2022towards}, and CORD~\cite{park2019cord}) across three difficulty levels: \textit{(1) Simple} (direct extraction), \textit{(2) Intermediate} (standardized formatting), and \textit{(3) Complex} (relational reasoning and inference). To ensure robust training signals, data quality is further optimized through \textit{rule-based filtering}, \textit{LLM-as-a-judge evaluation}, and \textit{rejection sampling}.

\paragraph{STEM} 

Robust STEM reasoning serves as the cognitive cornerstone of next-generation artificial intelligence, directly empowering advanced analytical tasks such as algorithmic problem-solving and automated scientific discovery.
To enhance the model's logical reasoning, we construct a high-quality dataset comprising 60\% open-source and 40\% proprietary/synthetic data. This composition provides a solid training foundation, ensuring both extensive knowledge breadth and deep reasoning structures. The specific construction strategies are as follows:
\begin{itemize}
\item \textbf{Public Data Curation:} We systematically collect and integrate over 70 open-source datasets (such as MAVIS~\cite{DBLP:conf/iclr/ZhangWJGZT0ZZG025}, AI2D~\cite{DBLP:conf/eccv/KembhaviSKSHF16}, 
MMK12~\cite{DBLP:journals/corr/abs-2503-07365}, and ViRL39K~\cite{DBLP:journals/corr/abs-2504-08837}). To improve data quality and representativeness, we perform deep-level deduplication and resampling on the original data, ensuring that the model is exposed to diverse STEM knowledge content spanning from basic education to higher education levels.
\item \textbf{Private Data Synthesis:} We focus on strengthening the quality control of synthetic data and the deep modeling of domain-specific knowledge. 

This involves the large-scale, in-house collection and structured processing of K12 and STEM content.
This curated data not only contains expert-annotated logical reasoning chains but also introduces complex multi-step reasoning synthetic samples, significantly improving the dataset's logical rigor and expressive precision, thereby addressing the shortcomings of open-source data in supporting deep reasoning capabilities. Additionally, for relatively specialized fields such as art, we synthesize high-quality image-text samples to supplement the model's knowledge representation in interdisciplinary contexts.
\end{itemize}

We restructure the dataset to enhance the model's multi-step reasoning in complex STEM tasks. First, we clean and rewrite problem statements to remove noise and improve clarity. Next, we construct structured reasoning chains for each instance, breaking complex problems into traceable, step-by-step nodes. This approach strengthens the model's multi-step reasoning, long-range dependency handling, and implicit logic modeling.

\paragraph{GUI} 
The training data for our GUI agent combines real-world collections with open-source datasets, using tailored processing pipelines to ensure high data quality and cross-platform adaptability. For real-world data, building on recent automated synthesis methods (e.g., UI-TARS~\cite{qin2025ui}, EvoCUA~\cite{xue2026evocua}), we deploy tens of thousands of concurrent sessions to collect raw interaction traces. We then apply a strict refinement pipeline: first, we remove visual noise to isolate fundamental interactive elements. Next, we align the collected instructions with our model's input format to guarantee precise coordinate prediction. Finally, we translate raw metadata into natural language intents and carefully fine-tune the spatial coordinates. For open-source data, we aggregate established datasets (e.g., AgentNet~\cite{wang2025opencua}, ScaleCUA~\cite{liu2025scalecua}) and immediately filter out abnormal samples. To ensure the model handles various screen layouts robustly, we balance this data across different operating systems and applications. We also rewrite unclear or fragmented instructions to accurately reflect the user's true intent. To salvage complex boundary cases, we introduce a dual-model recovery process: a lightweight model filters out spatial noise, while a stronger reasoning model repairs semantic errors and recalibrates coordinates, allowing us to retain valuable long-tail data.

\paragraph{Grounding and Counting}
Visual grounding connects semantic recognition with spatial localization to identify both \textit{what} is present and \textit{where} it is located. We enhance this by using bounding boxes for region-level reasoning and extend it to support quantitative counting. \textbf{(1) Grounding:}
We create a large-scale multi-source dataset by aggregating over ten public sources, including GRiT~\cite{peng2023kosmos}, Visual Genome~\cite{krishna2017visual}, RefCOCO/+/g~\cite{kazemzadeh2014referitgame, nagaraja2016modeling, yu2016modeling}, Objects365~\cite{shao2019objects365}, and OpenImages~\cite{kuznetsova2020open}. To ensure data quality, we apply bounding boxes and filter misaligned annotations using a pre-trained VLM. This pipeline yields a curated dataset of approximately 60M samples, ensuring robust grounding and high-precision visual prompting. \textbf{(2) Counting:}
Expanding on grounding, we develop a counting dataset with 8M samples, based on PixMo-Points~\cite{deitke2025molmo} for point-based counting and TallyQA~\cite{acharya2019tallyqa} for complex reasoning. A VLM-based filtering process further improves annotation quality. These datasets support direct and point-based counting, using a normalized coordinate system [0, 1,000] for better resolution robustness and simpler downstream application.

\subsubsection{Data Cleaning, Filtering, and Sampling}

\paragraph{Decontamination and Deduplication}
To ensure data quality and evaluation fairness, we adopt a multi-stage decontamination pipeline that filters training samples at both the image and text levels. Specifically, we apply pHash-based matching for images and N-gram overlap detection for text to remove samples that are highly similar to public benchmarks, mitigating memorization effects. In addition, we perform semantic-level deduplication via clustering to eliminate redundant or overly similar queries, thereby improving data diversity and coverage while reducing the risk of overfitting.

\paragraph{Quality Filtering}
We design a multi-faceted filtering strategy to improve data quality in logical structure and knowledge representation:
\textit{(1) Rule-Based Filtering}: We apply heuristic rules to eliminate anomalies like web links, formatting errors, and redundant text in the CoT. This enhances the information density and efficiency of the reasoning chains.
\textit{(2) Model-Based Filtering}: We use automated evaluation to compare CoT reasoning against ground-truth answers, removing flawed derivation paths and incorrect results. Additionally, we score image captions for knowledge density to prioritize professional, high-quality samples with strong image-text alignment and rich semantic content.

\paragraph{Data Sampling} 
We establish a comprehensive metadata management system to ensure the SFT data distribution supports model generalization and robustness. We annotate each entry with its source, difficulty, domain, and quality scores to implement differentiated sampling strategies:
\textit{(1) Difficulty Proportional Control}: We quantify sample difficulty using the model's correct response rate over multiple rollouts. We dynamically adjust the ratio of easy, medium, and hard samples, enabling the model to consolidate foundational knowledge while mastering complex reasoning.
\textit{(2) Reasoning Length Balancing}: We balance the length distribution of CoT reasoning chains based on task complexity. This ensures diverse step coverage and guides the model to generate logically compact, efficient paths without being excessively verbose or overly brief.
\textit{(3) Diversity Preservation Mechanism}: We set a minimum retention ratio for each data source to ensure diverse domain representation. This prevents capability degradation on specialized tasks caused by over-filtering and guarantees balanced, multi-source learning.

\subsection{Visual Generation}

Foundation text-to-image models typically acquire their generative capabilities from Web-scale image–text data, where a large fraction of samples concentrate on high-frequency patterns (e.g., generic portraits or common objects), leading to severe long-tail distributions and low semantic density. This causes the model to spend much of its capacity memorizing repetitive patterns rather than learning the broad and diverse visual concepts it needs.

\begin{figure}
    \centering
    \includegraphics[width=0.8\linewidth]{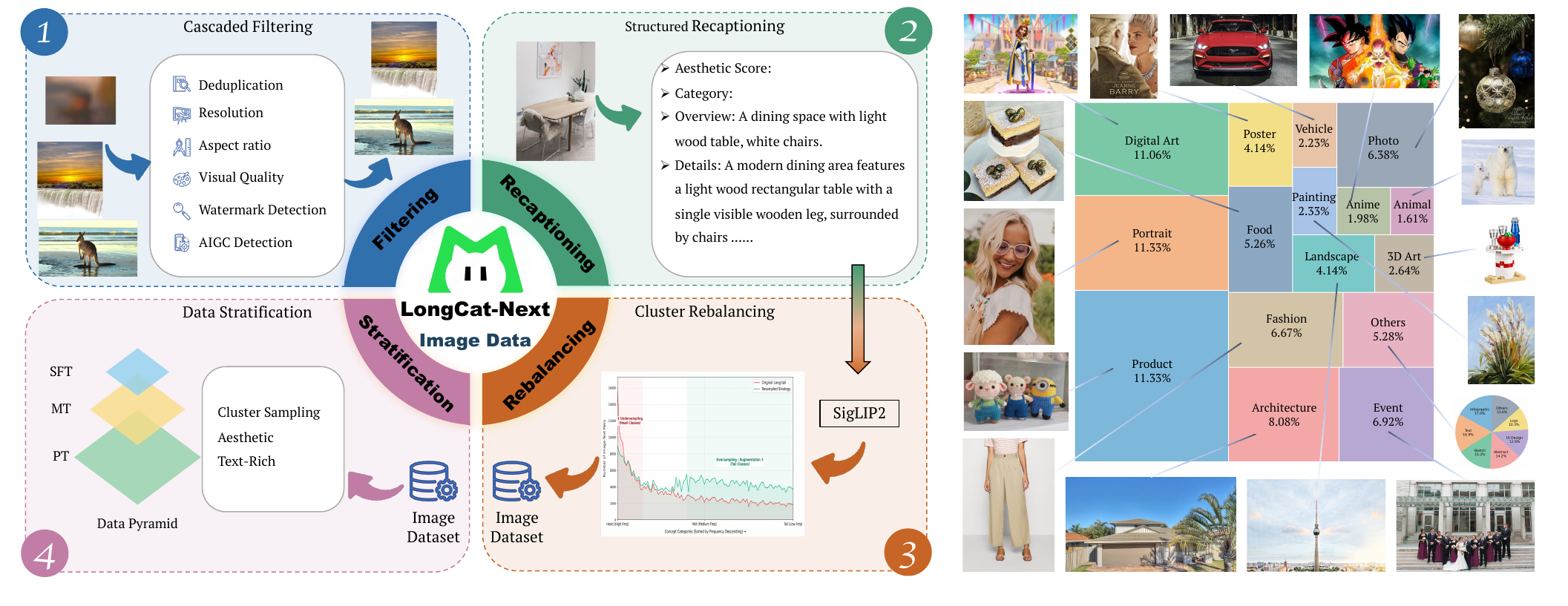}
    \caption{Left: The data curation and training process consists of three stages: (I) Pre-training with filtering and recaptioning for basic alignment; (II) Mid-training with semantic clustering and rebalancing to address data imbalance; (III) Supervised Fine-Tuning on high-quality, instruction-following data for improved aesthetics and text rendering. Right: The distribution of image sources.}
    \vspace{-3pt}
    \label{fig:placeholder}
\end{figure}

To move beyond purely scale-driven approaches, we design our data strategy around \textit{maximizing effective semantic coverage and density}. Our curation pipeline follows a multi-stage progression (see Fig.~\ref{fig:placeholder}): starting from broad foundational alignment, advancing to distributional rebalancing, and culminating in instruction-driven refinement. Following this strategy, we curated an in-house dataset of approximately 300 million image–text pairs from diverse public sources and web platforms~\cite{team2025longcat}. Compared to single-source datasets, this multi-source approach significantly enhances both semantic diversity and visual coverage.
In addition, we place particular emphasis on \textit{text-rich visual content}, as precise text rendering remains a critical challenge for real-world applications. Our text-centric corpus includes: (1) Synthetic typographic images. (2) Natural scenes featuring embedded text. (3) Professional poster and advertisement layouts. (4) Information-dense infographics and chart-style content. By providing stronger supervision for complex layouts and vision–language interactions, this strategy maximizes concept coverage while keeping dataset size manageable and training efficient.

\subsubsection{Stage I: Pre-training}
We first establish robust visual–text alignment through filtering and recaptioning.
 
\textit{Data Filtering.}
We apply multi-stage filtering to ensure data quality:
(1) \textbf{Deduplication}: SigLIP-based embeddings with cosine similarity to remove near-duplicates;
(2) \textbf{Quality Control}: resolution $\geq 384\times384$, aspect ratio constraints, and HPS v3 scoring~\cite{ma2025hpsv3};
(3) \textbf{Content Filtering}: watermark detection (Qwen3-VL-8B) and AIGC filtering to remove synthetic or corrupted samples. \textit{Structured Recaptioning.}
We perform recaptioning using Qwen3-VL-8B to generate structured descriptions (overview, detailed attributes, and tags), improving semantic richness and alignment. For text-rich images, OCR signals (PaddleOCR) are integrated to enhance text–image consistency.

\subsubsection{Stage II: Mid-training — Cluster-based Rebalancing}
Once the model has learned basic visual–textual alignment, the training bottleneck shifts to \textbf{distributional imbalance}. Aggressive aesthetic filtering can narrow the model toward a homogeneous ``AI-style'' aesthetic at the expense of diversity. To address this, we reorganize the dataset through a cluster-based rebalancing procedure:

\textit{Semantic Clustering}: We encode all images using SigLIP2 and perform large-scale distributed K-Means clustering via FAISS, partitioning the dataset into millions of semantic clusters. \textit{Intra-cluster Deduplication}: Within high-density clusters, we apply aggressive deduplication to remove near-redundant samples.
\textit{Inter-cluster Reweighting}: We adopt a power-law rebalancing strategy that increases the sampling probability of sparse clusters (e.g., rare flora or specialized scientific instruments), effectively flattening the distribution so that long-tail concepts receive sufficient gradient updates during training.

\subsubsection{Stage III: Supervised Fine-Tuning}
The final stage performs supervised fine-tuning (SFT) to align the model with complex human instructions. The SFT dataset is constructed from three complementary sources:

\textit{Cluster-Representative Sampling}: From the semantic clusters obtained in Stage~II, we select the highest-quality exemplar near each centroid, maintaining broad diversity while eliminating noisy samples. \textit{High-Aesthetic Data}: A curated set of professional photography and digital art is incorporated to improve lighting, composition, and fine-grained detail. \textit{Text-Rich Data}: To address the text-rendering bottleneck, we include a dedicated subset of synthetic typographic layouts and real-world OCR-grounded scenes (e.g., posters and infographics), providing dense supervision for precise character generation.

By progressively evolving the training data—from large-scale noisy pairs, to semantically rebalanced clusters, and finally to high-fidelity instruction-tuning sets—our model achieves broad concept coverage and strong aesthetic quality within an efficient computational budget.

\definecolor{boxgray}{RGB}{250, 250, 250} 
\definecolor{bordergray}{RGB}{160, 160, 160} 

\newtcolorbox{qwenbox}[1]{
    enhanced,
    colback=boxgray,
    colframe=bordergray,
    arc=6pt,                  
    boxrule=0.8pt,            
    top=12pt, bottom=10pt, left=15pt, right=15pt,
    fontupper=\small\sffamily\linespread{1.2}\selectfont, 
    attach boxed title to top left={xshift=8mm, yshift=-3mm},
    boxed title style={
        colback=black, 
        colframe=black, 
        arc=4pt, 
        boxrule=0pt,
        top=3pt, bottom=3pt, left=10pt, right=10pt
    },
    title={\bfseries #1} 
}

\subsection{Audio}
\subsubsection{Data Collection and Processing}
Audio pretraining at scale depends heavily on web-crawled recordings, but raw audio is often much less uniform than text-image data, with noisy supervision, duplication, background interference, and highly imbalanced speaker distributions. To move beyond purely scale-driven collection, we design our audio data strategy around broad acoustic coverage, controllable paired supervision, and targeted support for specialized perception tasks. As summarized in Fig.~\ref{fig:audio-data-pipeline}, this strategy is realized through three complementary data sources: large-scale web audio, synthetic speech-text data, and curated task-specific datasets for abilities such as paralinguistic perception and audio-event understanding that are less accessible from naturally collected web audio.

\noindent
\begin{minipage}[t]{0.56\textwidth}
\vspace{0pt}

\textbf{Web-Audio Curation.}
This data recipe is implemented through a multi-stage curation pipeline. Starting from 19.9 million hours of open-web audio, we apply three major curation steps:
(1) Filtering and Alignment: VAD, forced alignment, deduplication, dual-ASR consistency filtering, DNSMOS-based quality filtering, and background-noise filtering are used to remove noisy or unreliable samples and improve supervision fidelity;
(2) Speaker Distribution Control: speaker-embedding-based clustering is used to retain medium-frequency speaker clusters while suppressing both overrepresented synthetic voice patterns and extremely sparse outlier categories, thereby reducing distributional skew and improving speaker diversity;
(3) Re-segmentation: utterances are re-segmented so that their length distribution better aligns with the power-law structure of natural speech, improving corpus consistency and training usability at scale.
Collectively, these stages transform raw web audio into a cleaner, more balanced speech corpus while preserving acoustic diversity. After curation, 3.2 million hours of processed audio are retained, corresponding to 16.2\% of the original collection.

\end{minipage}\hfill
\begin{minipage}[t]{0.42\textwidth}
\vspace{0pt}
\centering
\includegraphics[width=\linewidth]{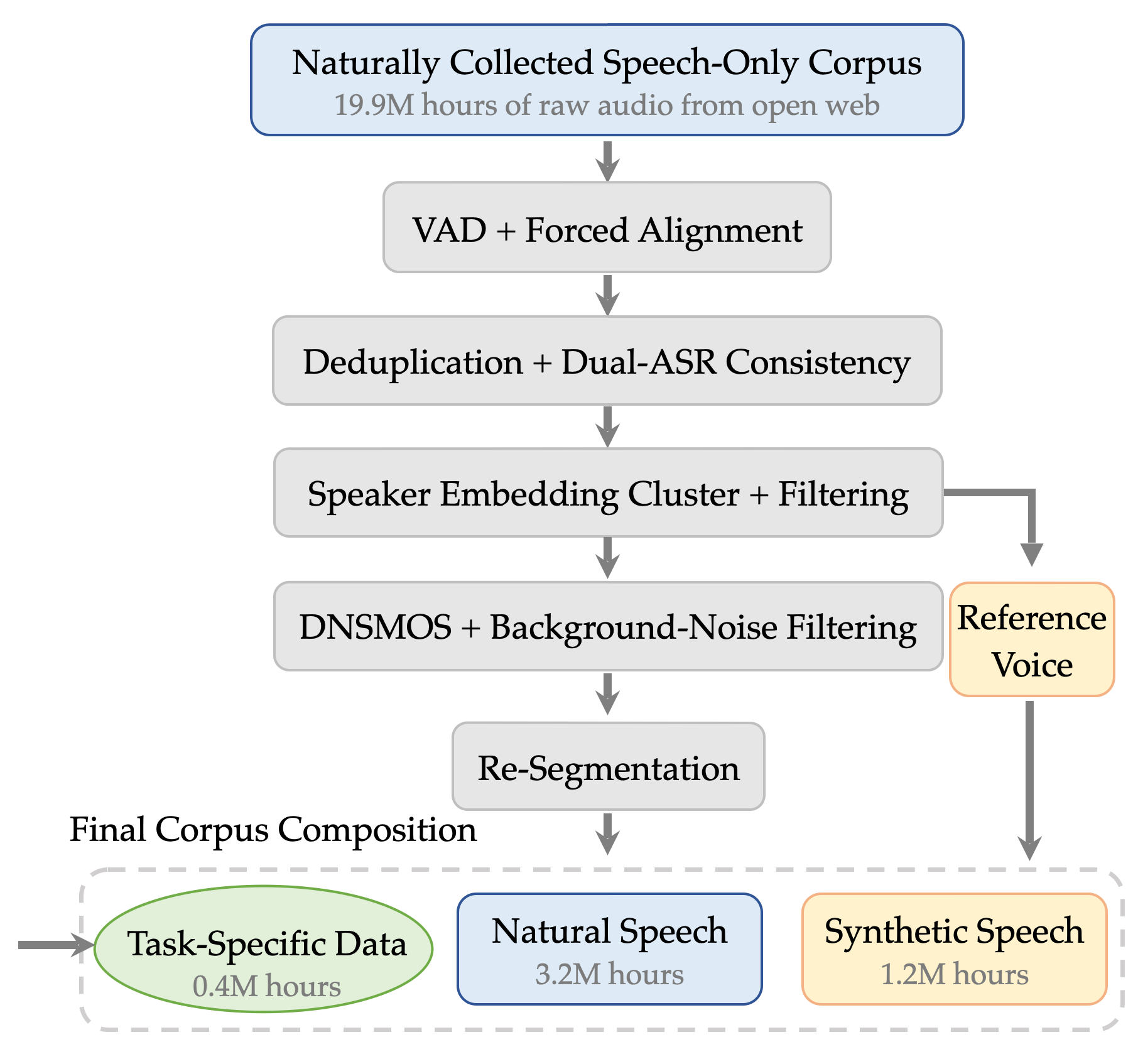}
\vspace{-6pt}
\captionof{figure}{Audio data pipeline.}
\label{fig:audio-data-pipeline}
\vspace{-8pt}
\end{minipage}

\paragraph{Synthetic Speech-Text Data.}
Based on representative speaker clusters derived from the curated web-audio corpus, reference voices are constructed to synthesize 1.2 million hours of speech-text data. This component serves as a source of controllable paired supervision, particularly when target text content or supervision structure cannot be reliably obtained from raw web audio. During synthesis, model-based controls on speaker similarity and ASR consistency are applied to improve the quality and reliability of the generated speech-text pairs.

\paragraph{Task-Specific Data.}
In addition, 0.4 million hours of task-specific datasets are collected from both open-source and internal resources to supplement capabilities that remain underrepresented in the main corpus, particularly paralinguistic perception and audio-event understanding. These datasets provide targeted supervision for specialized perception and understanding tasks that are not well covered by curated web audio or synthetic speech-text data.

\begin{table}[htbp]
\centering
\small
\setlength{\tabcolsep}{1em}
\caption{Detailed Statistics of Audio Pre-training Data. Text-Guided Audio is abbreviated as TextAudio here.}
\label{tab:audiodata}
\vspace{1em}
\begin{tabularx}{0.9\textwidth}{llc}
\toprule
\textbf{Task} &\textbf{ Data Format} & \textbf{Prop. (\%)} \\
\hline
Automatic Speech Recognition (ASR) & <prompt, audio, transcript> & 11 \\
Contiguous TextAudio (TTS) & <textaudio\_1, textaudio\_2, textaudio\_3, ...> & 40\\
Interleaved Audio and Text (INTLV) & <audio\_1, text\_2, audio\_3, text\_4, ...> & 22 \\
Interleaved Audio and TextAudio (INTLV-TA) & <audio\_1, textaudio\_2, audio\_3, textaudio\_4, ...> & 22 \\
Pure Audio & <audio> & 2 \\
others (AQA, S2TT, etc.) & <prompt, audio, response> & 3 \\
\bottomrule
\end{tabularx}
\end{table}
\subsubsection{Data Typology and Statistics}
The audio data utilized in pre-training encompasses several formats: Automatic Speech Recognition (ASR), Pure Audio, Contiguous Text-Guided Audio, Interleaved Audio and Text (INTLV), Interleaved Audio and Text-Guided Audio (INTLV-TA), as well as small amounts of Audio Query Answer (AQA), Speech-to-Text Translation (S2TT), and other task types.
The data formats and proportions for these task types are summarized in Table~\ref{tab:audiodata}.

ASR and S2TT data ensure that the model possesses fundamental semantic alignment capabilities between text and audio. Notably, Contiguous Text-Guided Audio data consists of single or multiple text-guided audio spans separated by special marks. This format essentially corresponds to a Text-To-Speech (TTS) task: serial generation is analogous to non-streaming TTS, while parallel generation aligns with streaming TTS. By segmenting Contiguous Text-Guided Audio, the long-range correspondence between audio and text tokens is reduced, thereby simplifying the model’s learning process for generating audio under text supervision.
The interleaved data is divided into two types: INTLV and INTLV-TA. INTLV comprises alternating pure audio and text modalities, which enhance the model’s ability to complete text given audio context. INTLV-TA, on the other hand, consists of alternating pure audio and text-guided audio modalities, aiming to strengthen the model’s capability for audio continuation. These interleaved formats are beneficial for cross-modal knowledge transfer learning.
We utilize a small amount of pure audio data during the pre-training stage. We believe this is advantageous for training the depth transformer-based audio head, and it also helps the model to preserve timbre when generating audio.

Since unconditional audio generation is extremely challenging and can degrade the model’s ability for cross-modal continuation, we omit loss computation for the pure audio modality in ASR, INTLV, and INTLV-TA data following \cite{li2025baichuan}. In contrast, for the text-audio modality in TTS and INTLV-TA, we compute the loss jointly for both text tokens and audio tokens.

\section{Infrastructure}

The computational workload in multimodal models is inherently heterogeneous. The execution time of the embedding layer and modality-specific loss modules (\textit{e.g.}, DepthTransformer) significantly differs from that of the core LLM transformer layers. Furthermore, this latency fluctuates dynamically based on the specific token distribution of each modality within a given sample. Under a naive linear and uniformly partitioned pipeline parallel configuration, this heterogeneity induces pronounced inter-stage load imbalance. It also exacerbates peer-to-peer (P2P) communication overhead, leading to substantial pipeline bubbles and reduced hardware utilization.

\subsection{VHalf-based Pipeline Parallelism}
To address this, we propose a profile-guided load-balancing pipeline schedule based on the V-Half~\cite{qi2024pipelineparallelismcontrollablememory} pipeline parallel architecture. It consists of two core architectural design choices: \textbf{(1) V-shaped Schedule:} Instead of a linear assignment, we fold the pipeline into a V-shape by co-locating the first computational stage (the embedding layer) and the final computational stage (the modality-specific loss modules) onto the same physical device. \textbf{(2) Shared Buffer:} We construct a shared buffer at the first stage of the V-shaped pipeline, enabling the modality loss module to directly access the hidden states corresponding to RVQ multi-level tokens after embedding lookup for loss computation.

\begin{figure}[t]
    \centering
    \includegraphics[width=1.0\linewidth]{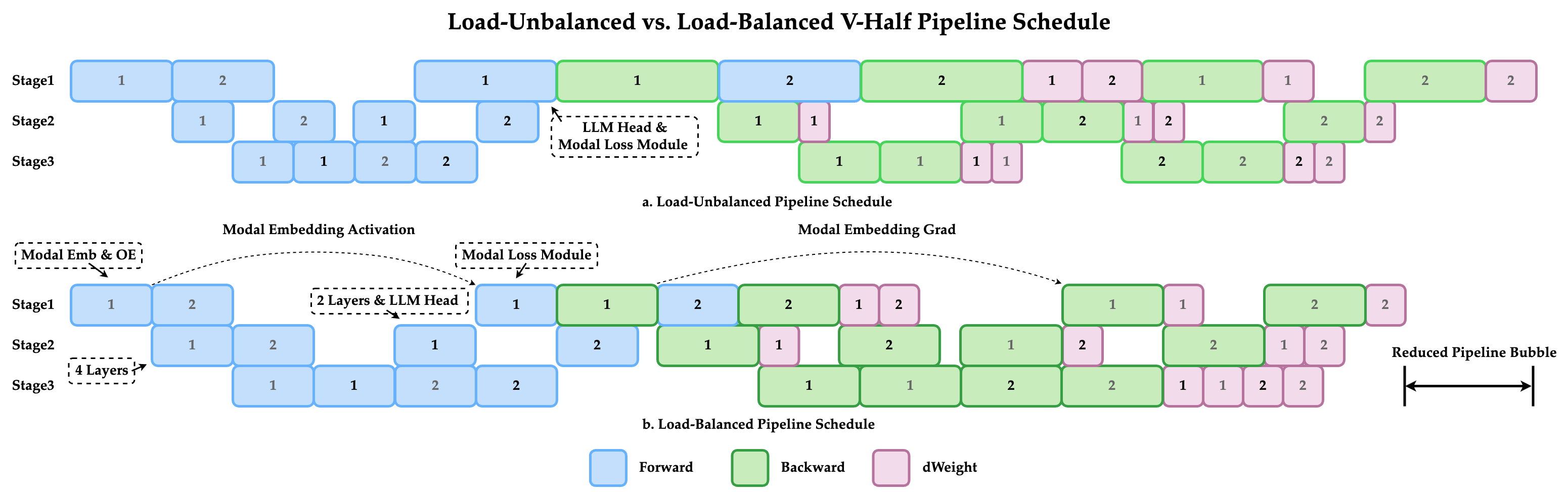}
    \caption{Infra training pipeline designed for LongCat-Next.}
    \label{fig:infra_training_pipeline}
    \vspace{-4pt}
\end{figure}

Specifically, we first profile the exact execution latency of the embedding layer, modality-specific loss modules, the LLM head, and a single LLM transformer layer. Based on these empirical measurements, we implement the V-Shaped stage assignment as follows (illustrated in Fig.~\ref{fig:infra_training_pipeline}): \textbf{(1) Co-locating Embedding and Modality Loss:} The embedding layer and the modality loss modules are deployed on the same pipeline stage as independent chunks. Although the execution time of the modality loss modules varies with input data, our profile-guided assignment ensures that the combined latency of these heterogeneous components remains strictly bounded within the ideal uniform chunk latency. \textbf{(2) Decoupling the LLM Head:} To prevent overloading the V-shape's anchor device, we strategically separate the LLM head from the modality-specific loss modules. The LLM head, bundled with a small number of LLM transformer layers, is assigned to a separate pipeline stage. \textbf{(3) Adaptive LLM Layer Distribution:} The remaining standard LLM transformer layers are evenly distributed across the rest of the chunks. This ensures that the baseline computational cost of each pipeline stage is approximately uniform across the entire cluster.

This V-shaped configuration yields two significant system-level benefits: \textbf{(1) Effective Bubble Mitigation:} By folding the pipeline and carefully isolating the heterogeneous workloads, the dynamic computational overhead of the embedding and modality loss modules is effectively absorbed by the schedule. This prevents these volatile modules from becoming systemic bottlenecks, substantially mitigating pipeline bubbles, and achieving near-perfect load balance. \textbf{(2) Elimination of Cross-Stage Communication:} Because the embedding layer and its corresponding modality loss module are co-located on the exact same device, both the forward activation passing and the backward gradient transmission between them are resolved entirely via zero-copy intra-device memory access. This design fundamentally eliminates the high cross-stage communication overhead that is typically required for these boundary modules.

\section{Advantages and Future Work}
Under the DiNA framework, multimodal modeling naturally inherits the advantages of discrete training, including the unification of generation and understanding, as well as infrastructure-friendly deployment. Furthermore, in the post-training stage, discrete visual representations can be optimized using modeling strategies analogous to those used in language, providing preliminary evidence for the potential of a unified discrete modeling paradigm.

\subsection{Reinforcement Learning}
Discrete representations in reinforcement learning (RL) offer inherent advantages that make them particularly suitable for multimodal tasks. The discrete visual latent space naturally serves as an action space, enabling seamless compatibility with existing language model RL methods such as Group Relative Policy Optimization (GRPO)~\cite{shao2024deepseekmath}. This avoids the need for sampling process conversion (\textit{e.g.}, ODE-to-SDE in flow models~\cite{liu2025flow, xue2025dancegrpo}) while maintaining a finite Markov Decision Process (MDP) structure for efficient policy optimization. In this work, we leverage these advantages by applying GRPO-based optimization to both image understanding and generation tasks, demonstrating the versatility of discrete representations across multimodal applications.

\textbf{RL for Text-to-Image Generation.} RL demonstrates superior adaptability and generalization capability in text-to-image generation tasks, particularly when facing data scarcity, complex scenarios, or ambiguous requirements~\cite{blip3oNext2025,liu2025flow}. 
The reward-driven optimization of RL provides significant advantages in text-to-image precise alignment and specialized capability enhancement. 

Given an input prompt, the policy generates a sequence of discrete tokens, which are subsequently decoded into an image by a frozen decoder. The generated image is then evaluated by a set of multi-dimensional reward models to compute a scalar reward signal. 
During the forward pass, the model outputs probability distributions $\pi_\theta^{(l)}$ for each quantization level $l$ by depth transformer. For policy optimization, we compute a weighted objective that considers the GRPO loss from all levels, allowing us to balance their contributions during training. 
The modified GRPO objective is formulated as:
\begin{align}
r_t^{(l)} &= \frac{\pi_\theta^{(l)}(a_t^{(l)}|s_t)}{\pi_{\text{old}}^{(l)}(a_t^{(l)}|s_t)}, \\
\mathcal{L}_{GRPO} &= \mathbb{E} \left[ \sum_{l=1}^{L} w_l \cdot
    \min \left(
        r_t^{(l)} \cdot \hat{A}_t,\;
        \mathrm{clip}(r_t^{(l)}, 1-\epsilon, 1+\epsilon) \cdot \hat{A}_t
    \right)
\right],
\end{align}
where $L$ denotes the number of quantization levels in RVQ, $w_l$ is the weight coefficient for level $l$.  $\pi_\theta^{(l)}(a_t^{(l)}|s_t)$ and $\pi_{\text{old}}^{(l)}(a_t^{(l)}|s_t)$ are the probabilities assigned by the current and previous policies, respectively. $\hat{A}_t$ is the advantage estimate, $\epsilon$ is the clipping parameter. 

Our reward models are designed across four key dimensions: comprehensive capability evaluation, OCR capability enhancement, text-image semantic alignment, and image quality assessment. 

\textbf{(1) Comprehensive Capability Enhancement. }
To holistically improve generation quality, we evaluate object count, color accuracy, spatial positioning, and attribute consistency between generated images and prompts. We first detect objects, then identify their attributes, and infer spatial relationships. The dataset includes prompts of varying lengths, from simple phrases to detailed descriptions, enabling quantitative evaluation across these dimensions.

\textbf{(2) OCR Capability Enhancement.} For visual text rendering, we use the GOT-OCR 2.0 \cite{wu2023human} model to recognize text in rendered images, including plain text, formulas, tables, and geometric figures. By quantifying the edit distance between rendered and ground truth text, we objectively assess textual fidelity. Prompts contain 2 to 20 texts, and we apply bucket sampling for different text lengths. 

\textbf{(3) Text-Image Semantic Alignment.} We utilize vision-language models (VLMs) as reward model to assess whether generated images accurately represent prompt descriptions. The VLM-based reward signal effectively guides the model toward producing images that maintain deeper conceptual alignment with input prompts. 

\textbf{(4) Image Quality Assessment.}
For comprehensive quality evaluation, we employ HPS~\cite{wu2023human, ma2025hpsv3}, Aesthetic Score~\cite{Laion2022} and Unified Reward~\cite{wang2025unified} for human preference alignment and visual quality assessment. These complementary metrics capture both subjective aesthetic preference and objective quality metrics.

This multi-dimensional reward framework enables systematic improvement across several critical aspects of text-to-image generation, with each component addressing specific challenges, and effectively mitigating reward hacking issues associated with single reward optimization.

\textbf{RL for Image Understanding.} We further conduct RL training on image understanding tasks. We face a severe challenge related to entropy explosion. 

As training progresses, both policy entropy and the degree of training-inference mismatch increase simultaneously, leading to the generation of noisy and garbled tokens during rollouts. When low-probability tokens enter the training data, their sampling probability is further amplified, creating a positive feedback loop that exacerbates the explosive growth of entropy. This phenomenon is particularly severe for long rollout sequences due to the cumulative effect of mismatch across tokens.

We identify that the core issue is the influx of a large number of noisy and garbled sequences into the training process. To address this issue, we propose sequence-level filtering mechanisms, including the entropy-based filter and the training-inference difference filter. 

\textbf{Entropy based Filter.} During training, we compute the mean and variance of the sequence-wise entropy for each minibatch. Any sequence whose entropy $H_{\text{seq}}$ exceeds the batch mean $\mu_H$ by $n$ standard deviations $\sigma_H$ is considered an outlier and filtered by 
$H_{\text{seq}} > \mu_H + n \sigma_H$. 

\textbf{Training-Inference Difference based Filter.} 
For each rollout, we monitor the per-token probability difference between the sampling policy and the actor policy. If the absolute difference for \textbf{any} token within a sequence surpasses a predefined threshold $\delta$, i.e., $|\pi_{\text{sampler}}(a_t|s_t) - \pi_{\text{actor}}(a_t|s_t)| > \delta$, the entire sequence is considered to have a severe training-inference mismatch and is consequently discarded.

Integrating these two sequence-level filters into the GRPO objective results in a stabilized loss formulation. The modified objective, $\mathcal{L}_{\text{GRPO-Filtered}}$, applies the filters via indicator functions that mask out problematic sequences before the standard clipped surrogate loss is computed:

\begin{equation}\label{eq:grpo_filtered}
\begin{aligned}
\mathcal{L}_{\text{GRPO-Filtered}} = \mathbb{E}_{(s_t, a_t) \sim \mathcal{D}} \Bigg[
& \underbrace{\mathbb{I} \left\{ H_{\text{seq}}(s_t) \le \mu_H + n \sigma_H \right\}}_{\text{(1) Entropy Filter}} \;
\cdot \;
\underbrace{\mathbb{I} \left\{ \max_{t' \in \mathcal{T}_y} |\pi_{\text{sampler}}(a_{t'}|s_{t'}) - \pi_{\text{actor}}(a_{t'}|s_{t'})| \le \delta \right\}}_{\text{(2) Prob. Diff. Filter}} \\
& \cdot \;
\min \Big(
r_t(\theta) \cdot \hat{A}_t, \;
\mathrm{clip}\big(r_t(\theta), 1-\epsilon, 1+\epsilon\big) \cdot \hat{A}_t
\Big)
\Bigg],
\end{aligned}
\end{equation}
where $\mathbb{I}\{\cdot\}$ act as a hard binary mask; a sequence contributes to the gradient \textbf{only if} it passes both filters.

In summary, this integrated filtering approach provides a diagnostic and prophylactic solution. It diagnostically identifies sequences containing the critical failure mode and prophylactically removes them from the gradient update. By doing so at the sequence level, it directly addresses the root cause, moving beyond token-level approximations to provide the stability necessary for large training-inference mismatch. More analysis is presented in the Appendix ~\ref{rl_appendix}.

\begin{figure}[t]
    \centering
    \begin{subfigure}[b]{0.19\linewidth}
        \centering
        \includegraphics[width=\linewidth]{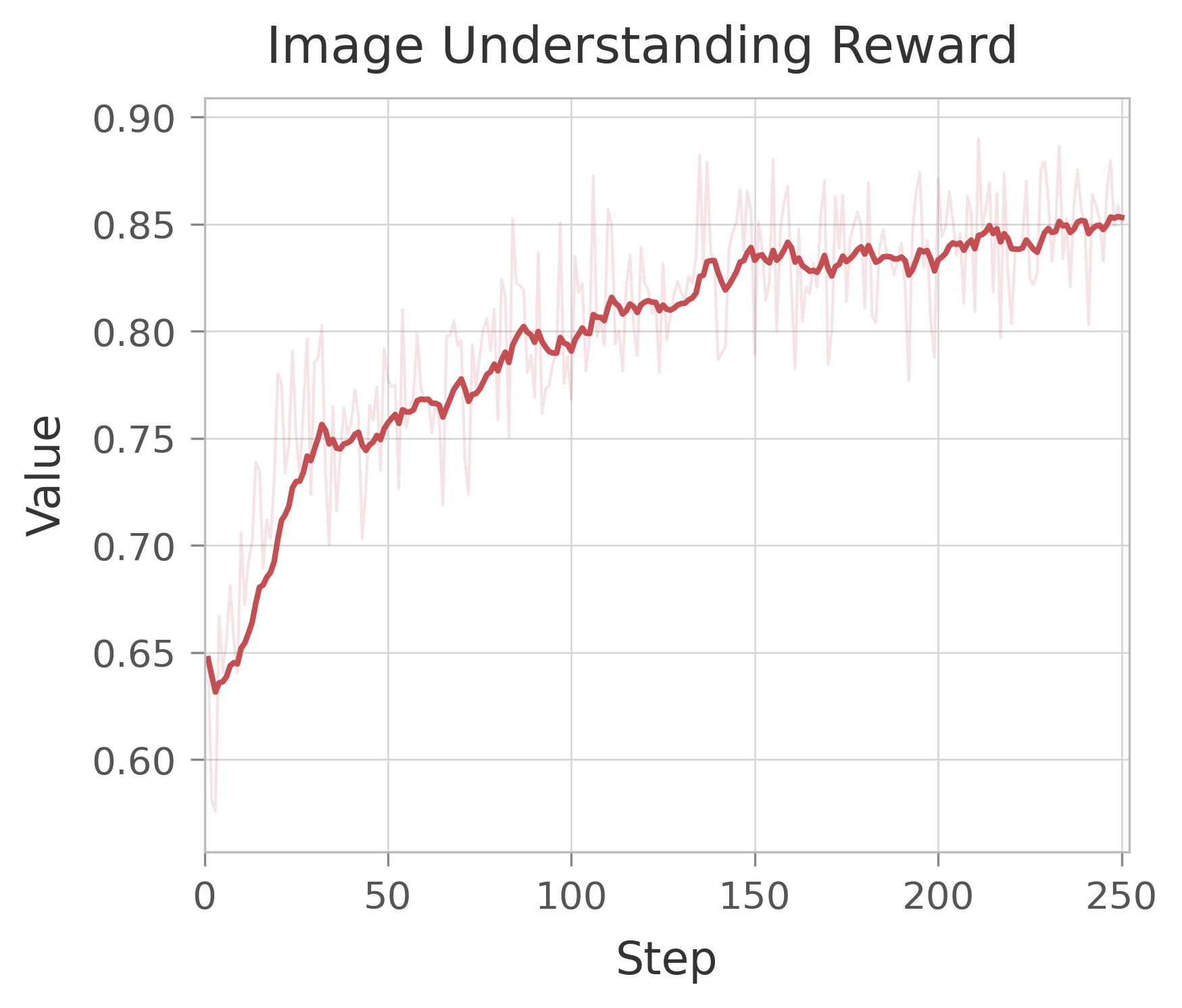}
        \label{fig:metrics:a}
    \end{subfigure}
    \begin{subfigure}[b]{0.19\linewidth}
        \centering
        \includegraphics[width=\linewidth]{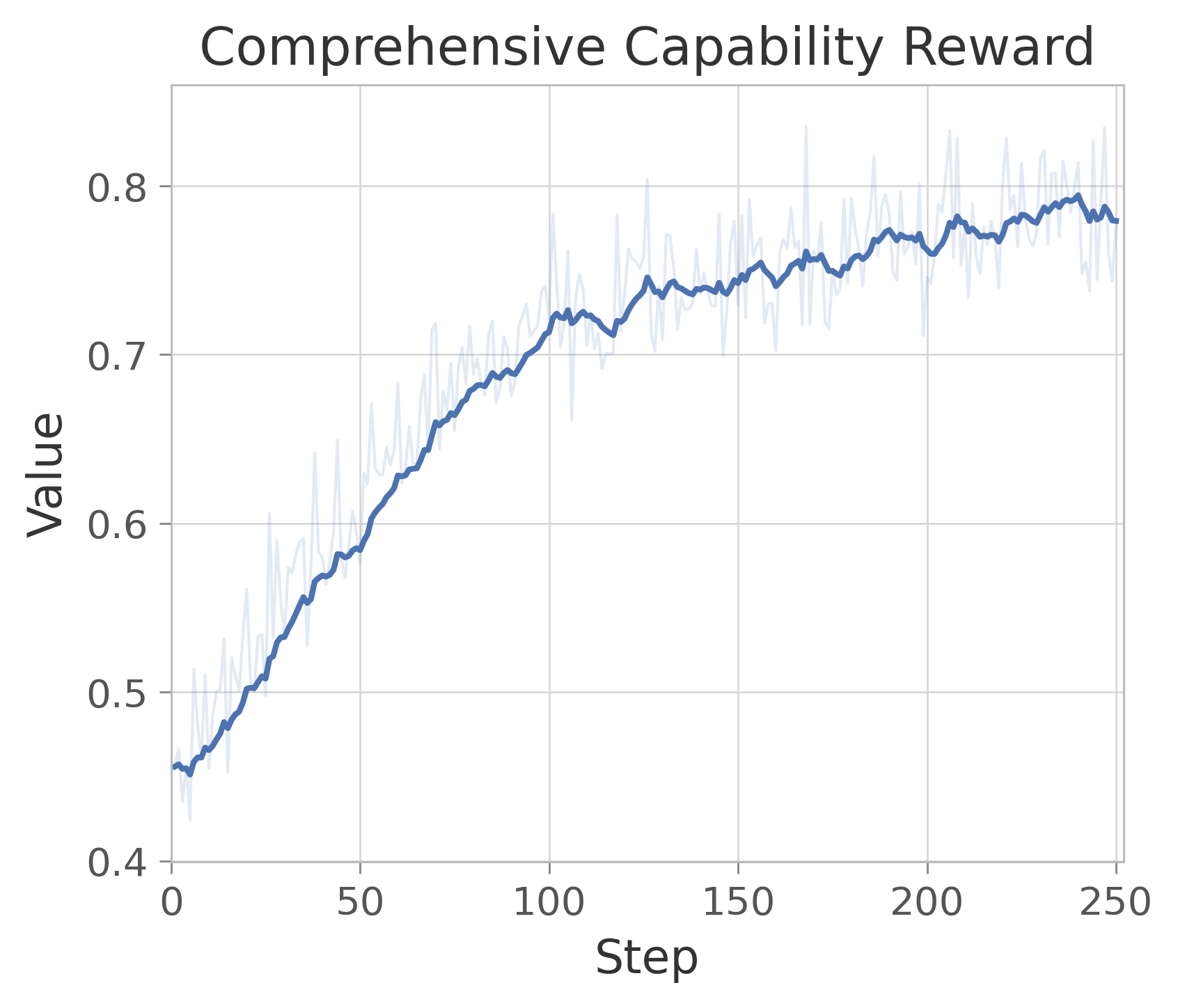}
        \label{fig:metrics:b}
    \end{subfigure}
    \begin{subfigure}[b]{0.19\linewidth}
        \centering
        \includegraphics[width=\linewidth]{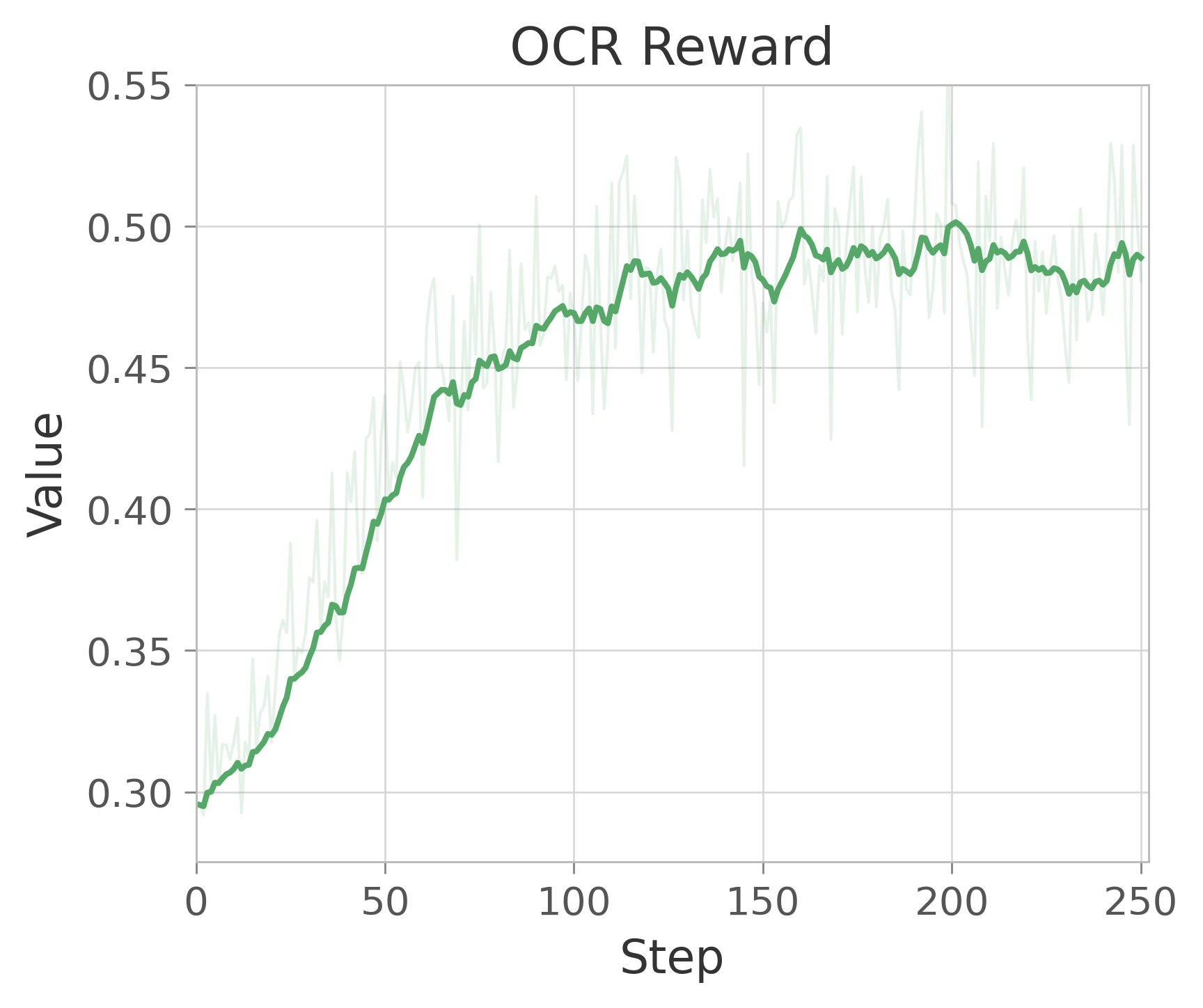}
        \label{fig:metrics:c}
    \end{subfigure}
    \begin{subfigure}[b]{0.19\linewidth}
        \centering
        \includegraphics[width=\linewidth]{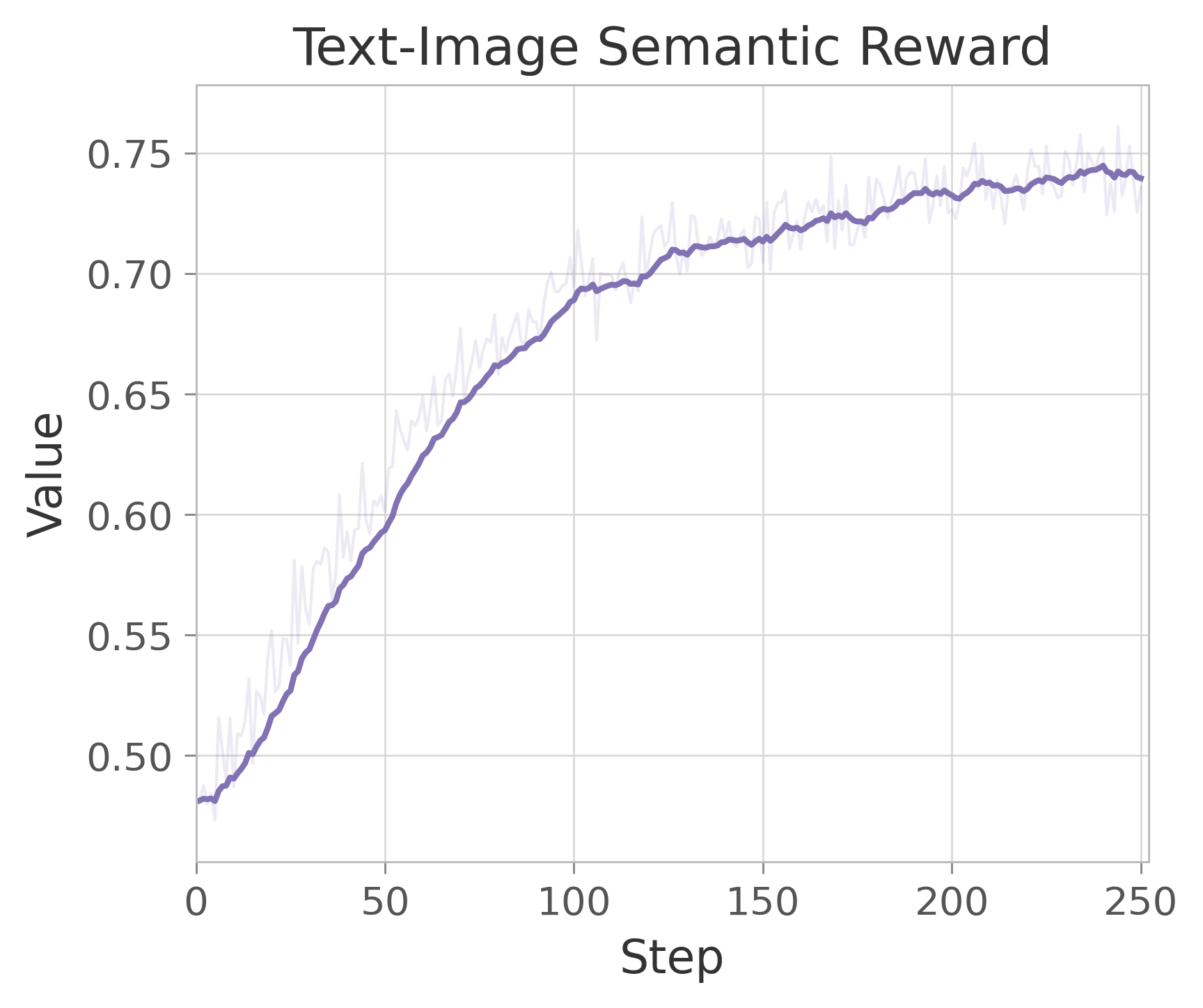}
        \label{fig:metrics:d}
    \end{subfigure}
    \begin{subfigure}[b]{0.19\linewidth}
        \centering
        \includegraphics[width=\linewidth]{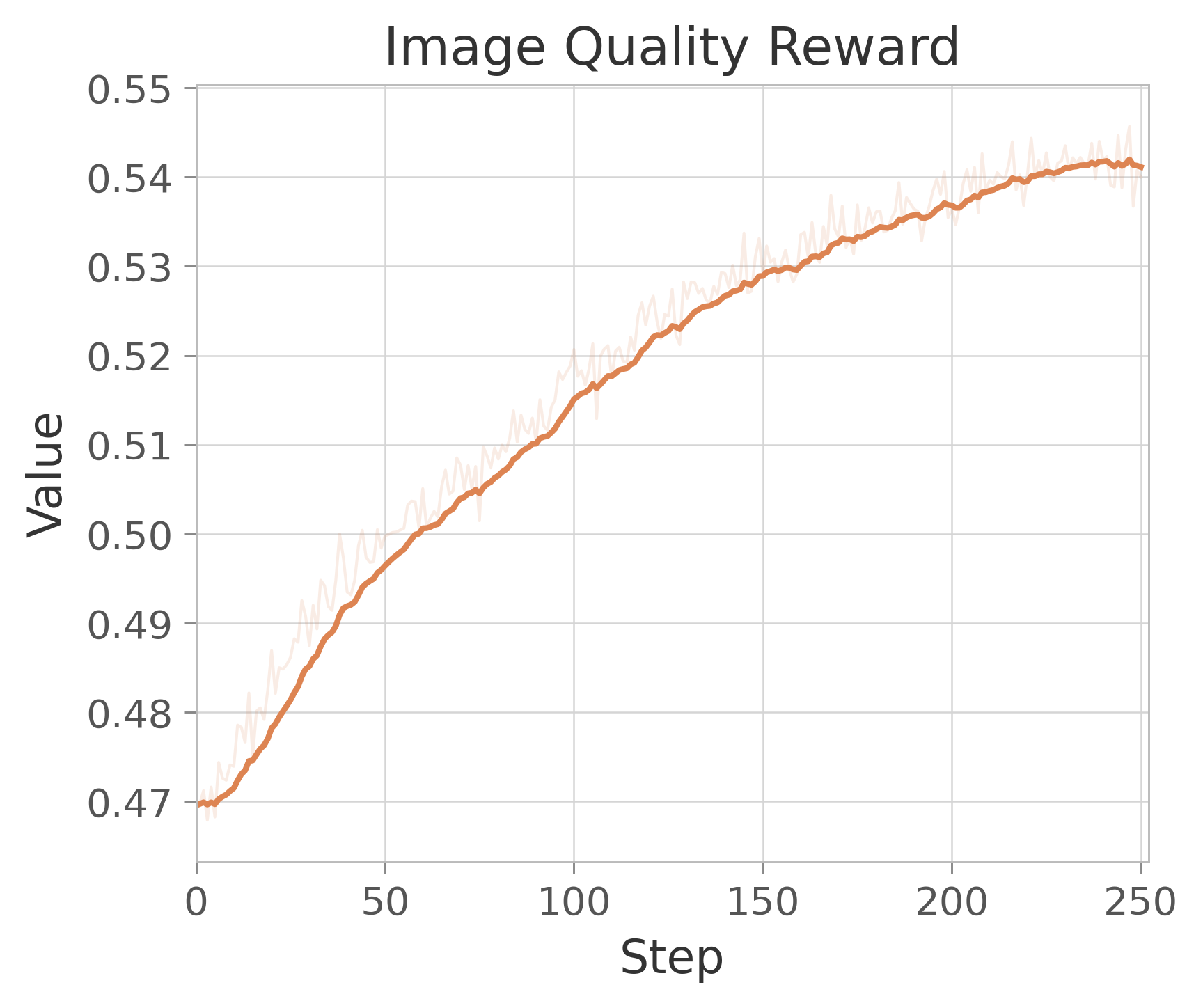}
        \label{fig:metrics:e}
    \end{subfigure}

    \caption{Reward Scores in RL.}
    \label{fig:metrics}
\end{figure}

Fig.~\ref{fig:metrics} shows the reward curves for image understanding and image generation. The rewards of image generation cover comprehensive capability, OCR performance, semantic alignment, and image quality. Each curve demonstrates the model’s progressive improvement across different tasks, reflecting steady increases in reward values throughout the training steps.

\subsection{Discussion and Future Work}

This work represents an initial step toward native multimodal modeling within a unified discrete autoregressive framework. Constrained by computational resources and data availability, current study remains limited in scope, and several important aspects are not yet fully explored. We view these limitations as opportunities for future investigation and welcome discussion and collaboration from the community. There are some potential future works as below:

\textbf{Vision Tokenizer.} The primary task of the current version is to realize the process of discrete encoding and decoding with dNaViT. As the model is currently focused on image understanding and generation tasks, the tokenizer is not yet fully optimized. Notably, the current de-tokenizer focuses primarily on ensuring semantic decoding consistency, rather than pixel fidelity. However, despite not being the most advanced version of the tokenizer, we have achieved strong performance in understanding and generation tasks, validating the effectiveness of the approach. This demonstrates that the model's capabilities can be further enhanced by scaling the data. We will update the tokenizer in future iterations to better meet the goals of the next version.

\textbf{Beyond Understanding and Generation.}
Our current evaluation emphasizes semantic integrity through tasks such as image-to-text and text-to-image. However, a truly native multimodal system should not be confined to these canonical directions. A natural next step is to generalize toward any-to-any generation and interleaved multimodal reasoning, where inputs and outputs span arbitrary combinations of text, vision, and audio. This includes long-context multimodal interaction, multi-turn grounded dialogue, and compositional generation where modalities dynamically condition one another. Enabling such flexible, unified interaction will be key to moving from task-specific competence to general multimodal intelligence.

\textbf{Data Scaling and Representation Learning.} A central open question is whether multimodality can introduce capabilities beyond those already captured by language. While language encodes a highly compressed abstraction of human knowledge, perceptual modalities provide complementary signals grounded in the physical world. Effectively leveraging this complementarity requires not only scaling data, but also improving its structure and alignment. In particular, the interaction between large-scale perceptual pretraining and discrete token modeling remains underexplored. Our preliminary results suggest that conventional pretraining paradigms do not directly translate into consistent gains, indicating a potential mismatch between continuous representations and discrete modeling. Future work may involve co-designing data, pretraining objectives, and discretization strategies to learn semantically aligned representations that better support unified multimodal reasoning.

Overall, we regard this work as a preliminary exploration, and many components remain to be validated and scaled. Future efforts will expand along these directions with more comprehensive experiments, stronger models, and broader evaluations, aiming to move closer to truly unified multimodal intelligence.

\section{Conclusion}

In this work, we revisit a fundamental question in native multimodality: how to represent and model the diverse modalities of the world within a unified learning paradigm. By introducing LongCat-Next, we explore the possibility that language-style discrete autoregressive modeling can naturally extend beyond text to encompass vision and audio within a shared token interface. Our results suggest that, with carefully designed tokenizers and training strategies, continuous perceptual signals can be effectively discretized while maintaining strong capabilities. As an initial step in this direction, we hope this work offers a different perspective on multimodal modeling and provides insights for the community toward building truly unified multimodal foundation models.

\section{Contributions and Acknowledgments}

Contributors are defined as individuals who held primary responsibilities in data curation, model design, model training, and infrastructure support throughout the full development lifecycle of LongCat-Next. The Acknowledgments section recognizes those who contributed to specific tasks such as data collection, annotation, model evaluation, and technical discussions. All names are listed in alphabetical order by first name; names marked with an asterisk (*) indicate former team members.

\section*{Contributors}

\begin{multicols}{4}
\small
Bin Xiao\\
Chao Wang\\
Chengjiang Li\\
Chi Zhang\\
Chong Peng\\
Hang Yu\\
Hao Yang\\
Haonan Yan\\
Haoze Sun\\
Haozhe Zhao\\
Hong Liu\\
Hui Su\\
Jiaqi Zhang\\
Jiawei Wang\\
Jing Li\\
Kefeng Zhang\\
Manyuan Zhang\\
Minhao Jing\\
Peng Pei\\
Quan Chen\\
Taofeng Xue\\
Tongxin Pan\\
Xiaotong Li\\
Xiaoyang Li\\
Xiaoyu Zhao\\
Xing Hu\\
Xinyang Lin\\
Xunliang Cai\\
Yan Bai\\
Yan Feng\\
Yanjie Li\\
Yao Qiu\\
Yerui Sun\\
Yifan Lu\\
Ying Luo\\
Yipeng Mei\\
Yitian Chen\\
Yuchen Xie\\
Yufang Liu\\
Yufei Chen\\
Yulei Qian\\
Yuqi Peng\\
Zhihang Yu\\
Zhixiong Han
\end{multicols}

\vspace{0.3em}
\section*{Acknowledgments}

\begin{multicols}{4}
\small
Changran Wang\\
Chen Chen\\
Dian Zheng\\
Fengjiao Chen\\
Ge Yang\\
Haowei Guo\\
Haozhe Wang\\
Hongyu Li\\
Huicheng Jiang\\
Jiale Hong\\
Jialv Zou\\
Jiamu Li\\
Jianping Lin\\
Jiaxing Liu\\
Jie Yang\\
Jing Jin\\
Jun Kuang\\
Juncheng She$^*$\\
Kunming Luo\\
Kuofeng Gao\\
Lin Qiu\\
Linsen Guo\\
Mianqiu Huang\\
Qi Li\\
Qian Wang\\
Rumei Li\\
Siyu Ren\\
Wei Wang\\
Wenlong He\\
Xi Chen\\
Xiao Liu\\
Xiaoyu Li\\
Xu Huang\\
Xuanyu Zhu\\
Xuezhi Cao\\
Yaoming Zhu\\
Yifei Cao\\
Yimeng Jia\\
Yizhen Jiang\\
Yufei Gao\\
Zeyang Hu\\
Zhenlong Yuan\\
Zijian Zhang\\
Ziwen Wang
\end{multicols}

\clearpage
\printbibliography



\clearpage
\section{Appendix}
\subsection{Training-Inference Mismatch Analysis in RL}
\label{rl_appendix}




\begin{figure}[h]
    \centering
    \begin{subfigure}[b]{0.24\linewidth}
        \centering
        \includegraphics[width=\linewidth]{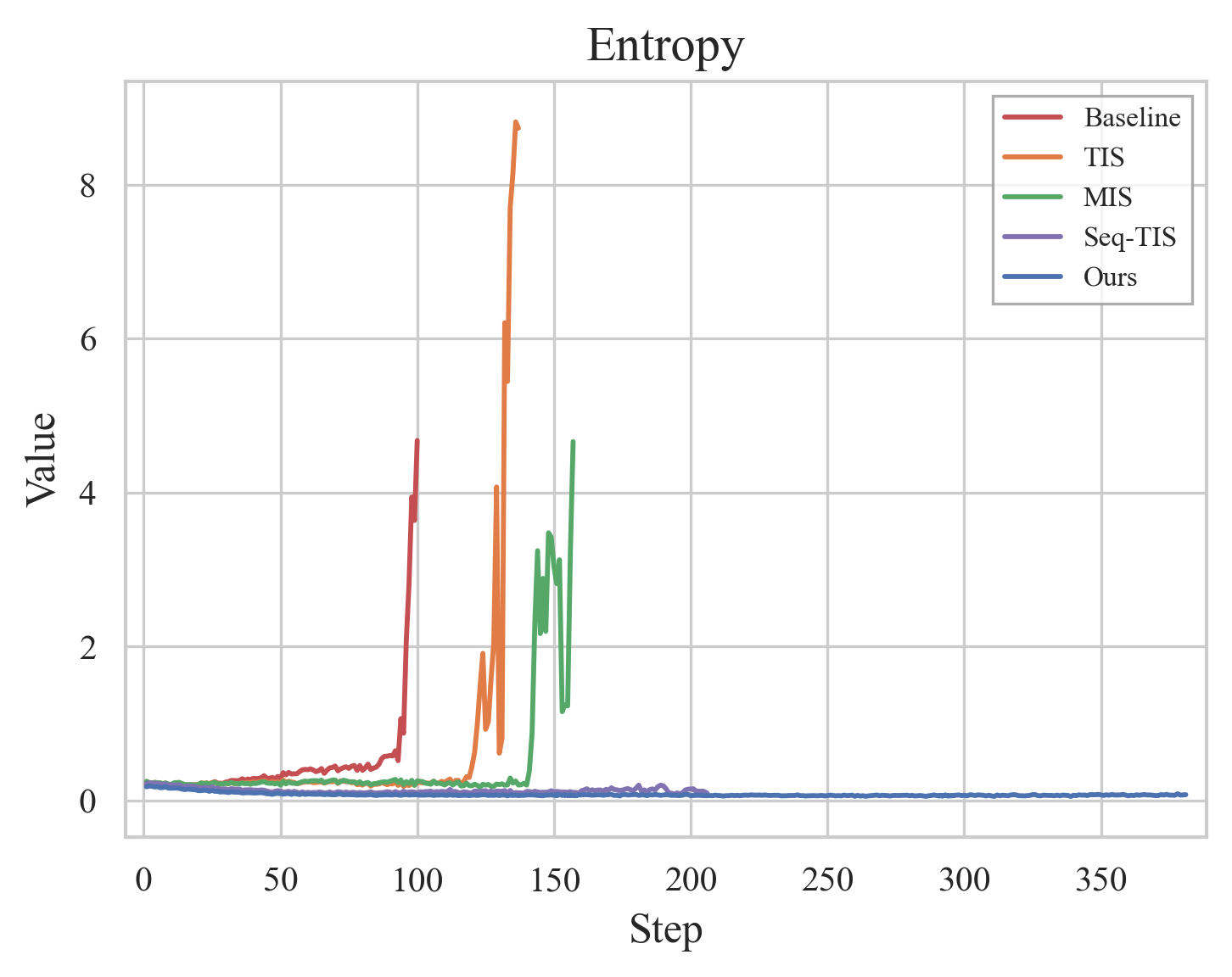}
        \label{fig:metrics:a}
    \end{subfigure}
    \hfill  
    \begin{subfigure}[b]{0.24\linewidth}
        \centering
        \includegraphics[width=\linewidth]{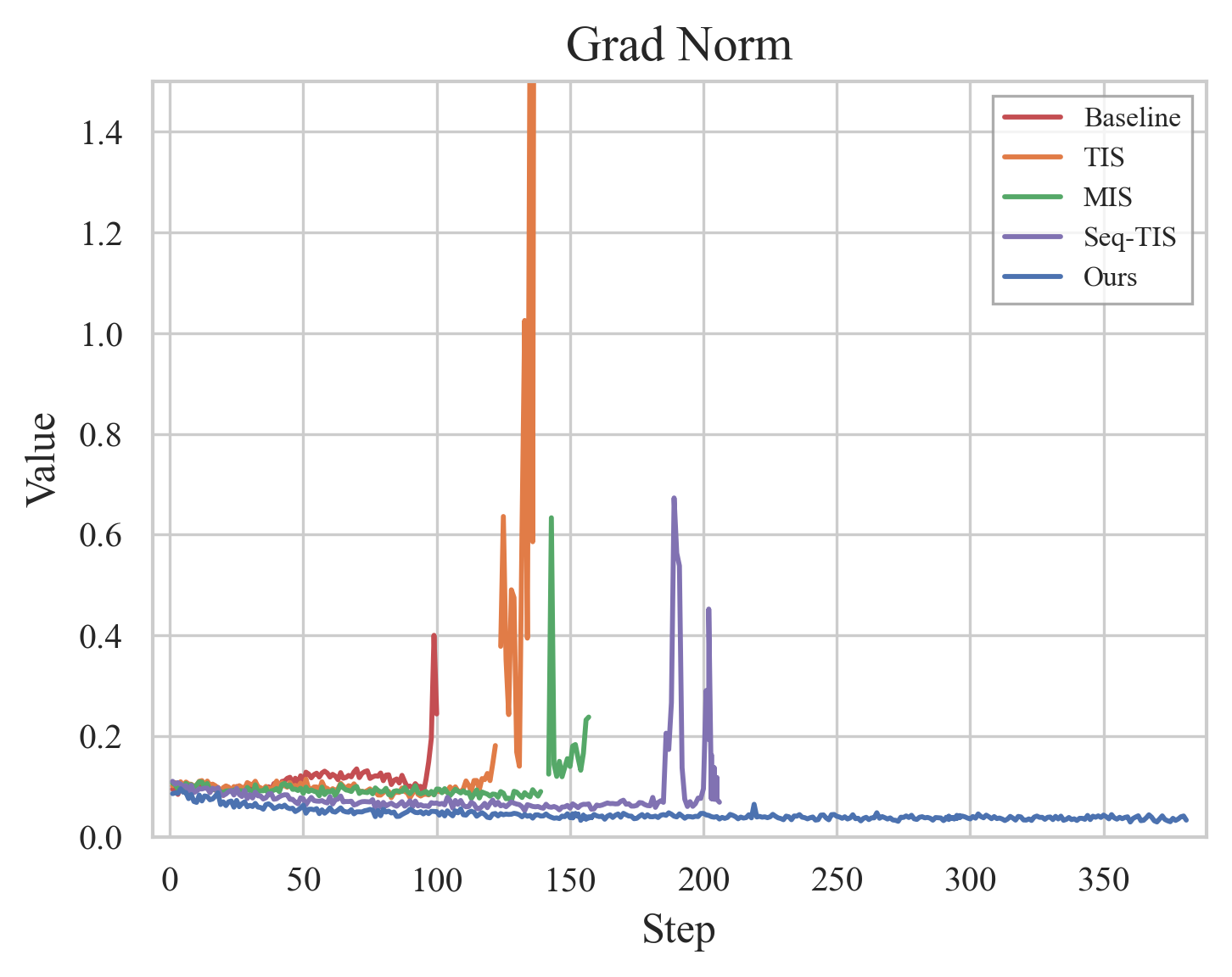}
        \label{fig:metrics:b}
    \end{subfigure}
    \hfill
    \begin{subfigure}[b]{0.24\linewidth}
        \centering
        \includegraphics[width=\linewidth]{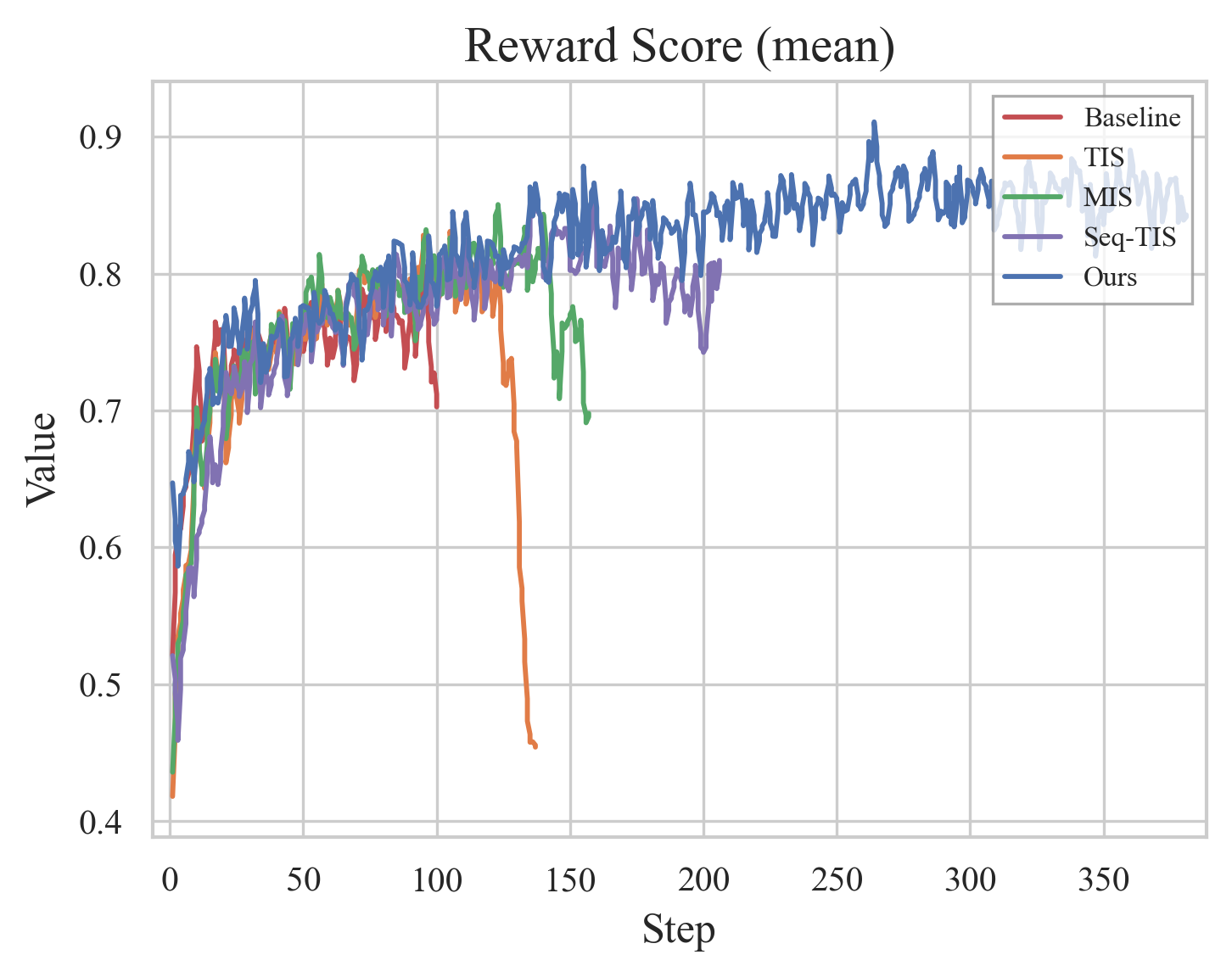}
        \label{fig:metrics:c}
    \end{subfigure}
    \hfill
    \begin{subfigure}[b]{0.24\linewidth}
        \centering
        \includegraphics[width=\linewidth]{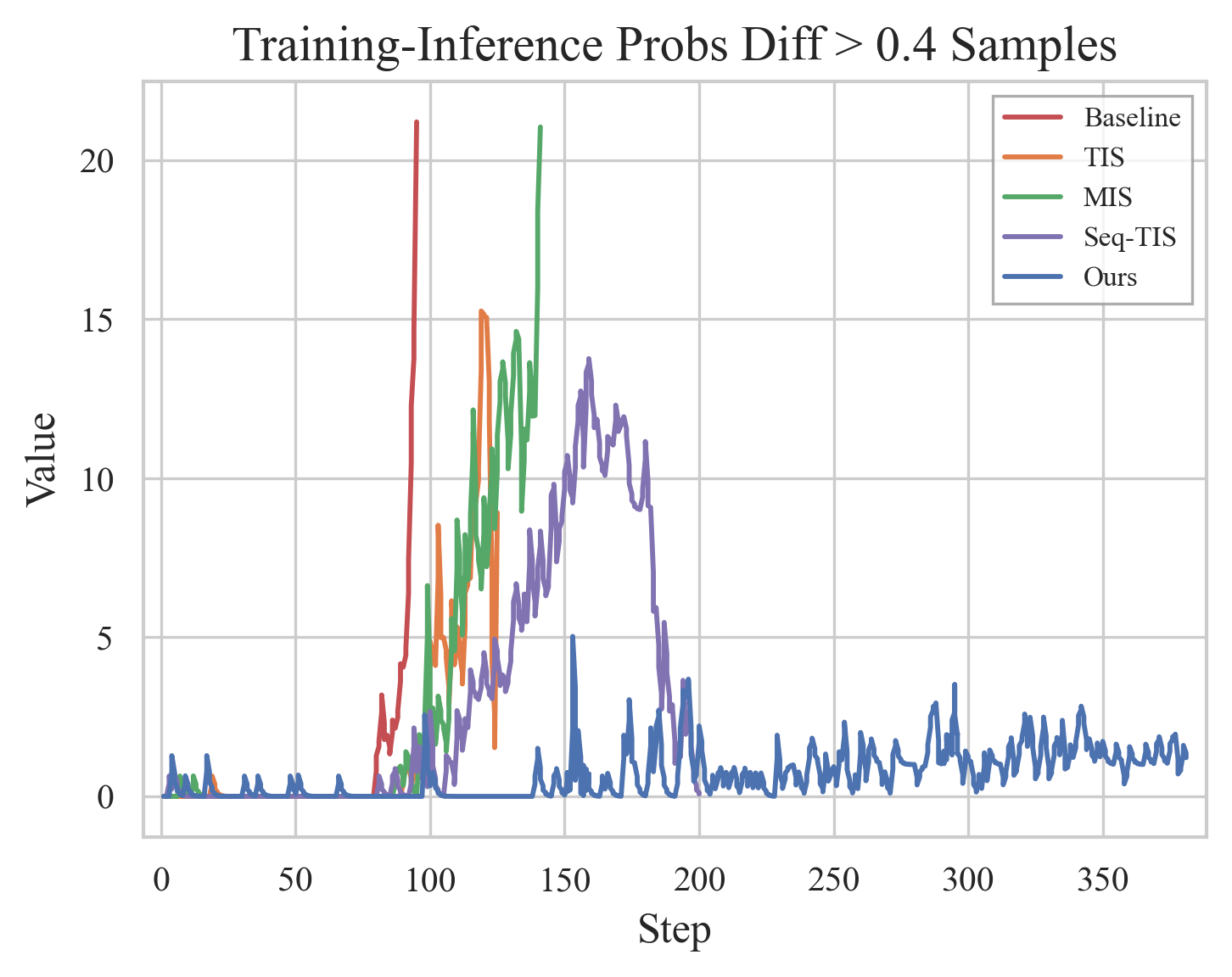}
        \label{fig:metrics:d}
    \end{subfigure}

    \caption{Training metrics comparison in RL.}
    \label{fig:metrics_rl}
\end{figure}

During RL training, we identified a critical phenomenon: sequence collapse is frequently triggered by severe training-inference mismatch in only a few or even a single token within a sequence. A token may have a probability exceeding 
$0.4$ under the sampling policy $\pi_{\text{sampler}}$, while its estimated probability under the training policy $\pi_{\text{actor}}$ can be as low as $10^{-2}$, whereas the average probability difference for the remaining tokens in the same sequence is on the order of $10^{-3}$. Such anomalous tokens have a catastrophic effect on overall sequence quality. They represent extremely low-probability noise that would normally be difficult to rollout but infiltrate the data due to hardware or numerical precision mismatches. More critically, subsequent generations contingent on these noisy tokens become unreliable, leading to error propagation throughout the sequence. Fig.~\ref{fig:metrics_rl} shows the number of samples in a batch where the training-inference probability divergence exceeds $0.4$ for at least one token. As illustrated, entropy explosion, gradient norm surges, and increasing training-inference policy divergence exhibit a positive correlation. This chain of factors ultimately leads to a decline in reward. 

To mitigate this issue, we experimented with several importance sampling correction techniques, including Truncated Importance Sampling (TIS)\cite{yao2025offpolicy} and Multiple Importance Sampling (MIS)\cite{liu2025speed}. As shown in Fig.~\ref{fig:metrics_rl}, these approaches merely delayed the onset of entropy explosion by a few training steps without addressing the fundamental problem. 

Some sequence-level correction methods\cite{liu-li-2025-rl-collapse} may also fail in this scenario and lead to a decrease in reward. The extreme divergence of a few anomalous tokens can be averaged out by the ratios of the many normal tokens in the sequence. 
Therefore, we adopt a more fundamental and direct intervention strategy, the threshold $\delta$ directly to the per-token absolute probability difference $|\pi_{\text{sampler}} - \pi_{\text{actor}}|$. This enables our filter to directly detect and pinpoint the finest-grained, token-level inconsistencies. By applying these strict filtering strategies, we effectively prevented noisy data from entering the training process.

\subsection{Experimental Analysis for RL}
\label{rl_exp}
In this experiment, we use Qwen-7B as the language backbone to validate the effectiveness of discretized RL. Except for LLM backbone, all other model architectures and configurations remain consistent with those used in the pretrained model. 
We evaluate the performance on both image understanding and image generation tasks. 

For the image understanding task, we conduct RL training using the open-source datasets VIRL39K~\cite{DBLP:journals/corr/abs-2504-08837} and Orsta-Data-47k~\cite{ma2025one}, along with in-house data. 
We merge and deduplicate the datasets based on image hash and prompt, perform four rollouts using the base model, and remove instances that are either entirely correct or entirely incorrect. Ultimately, we retain approximately 30K RL data samples. 
For the image generation task, we collect approximately 40K data samples, with each prompt potentially associated with multiple reward scores. 
Detailed performance results are presented in Table~\ref{rl_und}.

\begin{table}[htbp]
\vspace{-4pt}
\begin{center}
\small
\setlength{\tabcolsep}{8pt} 
\renewcommand{\arraystretch}{0.9}
\caption{Image Understanding and image generation performance based on Qwen-7B.}
\resizebox{0.95\textwidth}{!}{
\begin{tabular}{lccccccc}
    \toprule
    \multicolumn{1}{c}{} 
    & \multicolumn{4}{c}{\textbf{STEM}} 
    & \multicolumn{3}{c}{\textbf{General}} \\
    \cmidrule(lr){2-5} \cmidrule(lr){6-8}
    Dataset & MMMU & MMMU-Pro & MathVista & MathVision & MMStar  & RealWorldQA & MMVP \\
    \midrule
    baseline & 64.22 & 51.27 & 80.30 & 49.28 & 66.33 & 66.01  & 73.33 \\
    RL       & \textbf{66.45} & \textbf{53.58} & 81.90 & \textbf{53.52} & \textbf{71.13} &\textbf{72.54}  & 74.66 \\
    \midrule[\heavyrulewidth]
    \multicolumn{8}{c}{\textbf{OCR \& Doc}} \\
    \midrule
    Dataset & CharXiv$_\text{DQ}$  & CharXiv$_\text{RQ}$ & InfoVQA  & ChartQA & AI2D & \multicolumn{2}{c}{OmniDocBench(zh / en)$\downarrow$}    \\
    baseline & 85.45  & 51.8 & 81.69 & 88.56 & 81.57 & 0.187 & 0.256   \\
    RL   & \textbf{87.35}  & \textbf{56.7} & \textbf{83.85} & \textbf{92.08} & \textbf{85.13} & \textbf{0.169} & 0.266 \\
    \midrule[\heavyrulewidth]
    \multicolumn{8}{c}{\textbf{GenEval}} \\
    \midrule
    Dataset & Overall & single\_obj & two\_obj & counting & colors & position & color\_attr   \\
    baseline & 83.94 & 100.00 & 92.71  &  71.25 & 92.74  &  74.75 & 72.19    \\
    RL  &  \textbf{87.33} &  99.69 & 93.18 & \textbf{78.75}  &   94.15 & \textbf{81.50} &  \textbf{76.75}   \\
    \bottomrule
    \label{rl_und}
\end{tabular}
}
\vspace{-8pt}
\end{center}
\end{table}



RL exhibits significant improvements across the STEM, General VQA, and OCR benchmarks. For example, scores on datasets such as MMMU and MathVision show notable gains compared with the baseline model, demonstrating a substantial enhancement in the model's multi-step reasoning and problem-solving capabilities. Furthermore, for general VQA and OCR tasks such as RealWorldQA, MathStar, and ChartQA, the model also achieves significant improvements of around 5\%, indicating a comprehensive boost in its generalization capability. On the GenEval evaluation set for image generation, our model also achieves notable performance gains: counting accuracy improves by 7.50\%, position accuracy by 6.75\%, and color attribute accuracy by 4.56\%. These results highlight the effectiveness of reinforcement learning driven by reward feedback.

\subsection{Discrete Quantization Strategy}

\noindent
\begin{minipage}[c]{0.35\linewidth} 
    \centering
    \includegraphics[width=\linewidth]{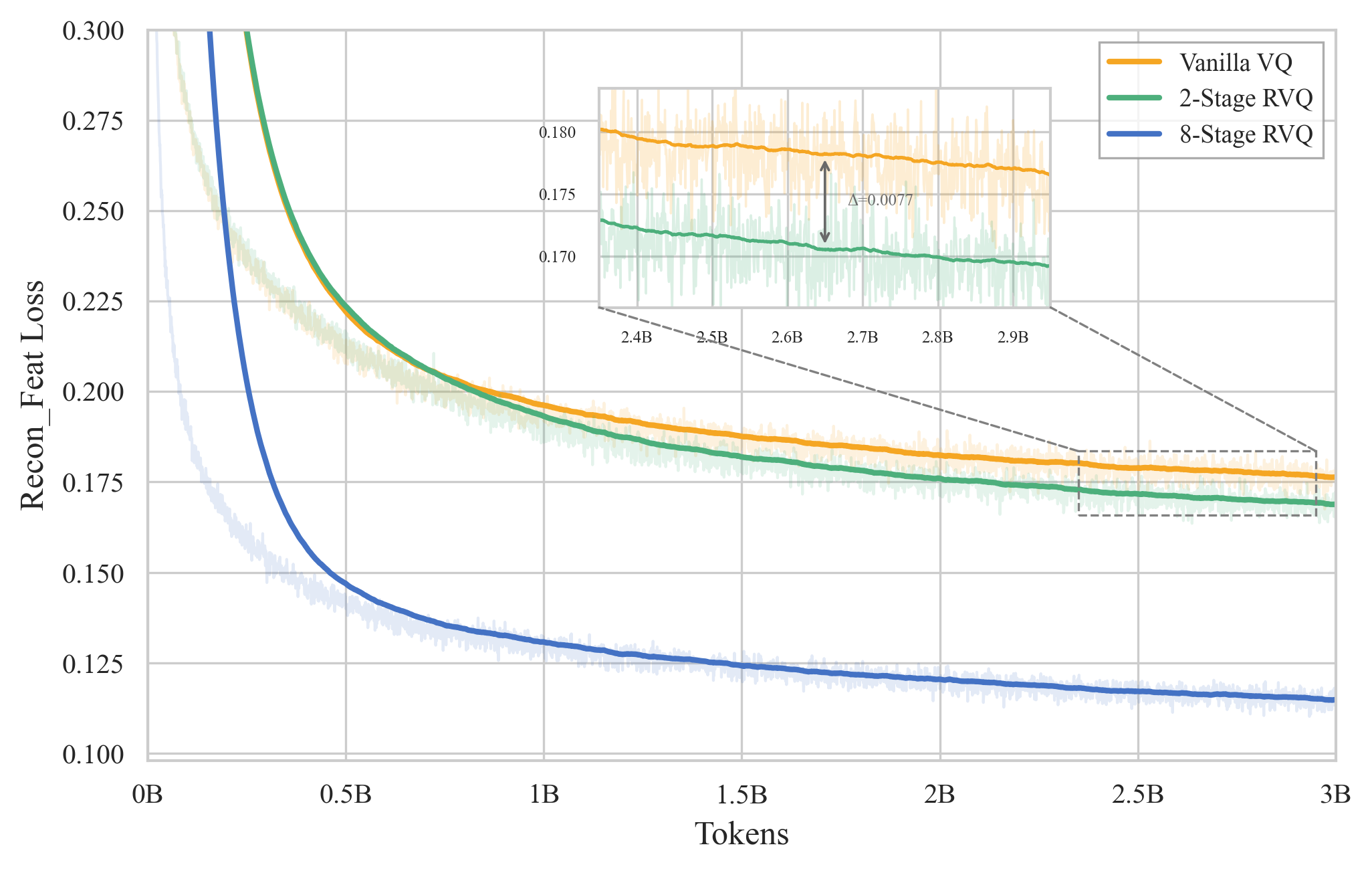} 
    \captionof{figure}{Comparison of feature reconstruction loss across different discrete quantization strategies.}
    \label{fig:rvq_loss}
\end{minipage}
\hfill
\begin{minipage}[c]{0.60\linewidth} 
\small 
To select a discrete quantization strategy that minimizes information loss, we utilize feature reconstruction loss as a proxy task to evaluate the method's information retention capabilities. As shown in Fig.~\ref{fig:rvq_loss}, the two-stage RVQ outperforms the vanilla VQ slightly. Furthermore, when scaling the RVQ to eight stages, the feature reconstruction loss decreases significantly. This demonstrates that the residual mechanism and compositionality of RVQ are essential for achieving discrete quantization with minimal information loss. Empirical validation confirms that the eight-stage RVQ achieves sufficiently low information loss without imposing excessive computational overhead, fully meeting our requirements. Consequently, we adopt it as the default setting for our model.
\vspace{-3pt}
\end{minipage}

\vspace{1em} 

\subsection{Vision Understanding and Generation under DiNA}

In the autoregressive modeling process under DiNA, as in Fig.~\ref{fig:rvq_loss}, we effectively balance performance and efficiency through multi-level encoding and decoding. During the understanding phase, visual signals are encoded into hierarchical discrete tokens, which are then additive over the levels before being fed into the large language model (LLM), ensuring that information from all levels is fully utilized. During the generation phase, hidden representations are passed to the DepthTransformer, which decodes them into multi-level tokens. This entire process is completed in a single autoregressive step, enabling efficient parallel decoding while maintaining high-quality generation. This approach allows visual understanding and generation to proceed seamlessly within a unified framework, overcoming the information loss and computational bottlenecks seen in traditional models.

\begin{figure}[hbtp]
    \centering
    \includegraphics[width=0.8\linewidth]{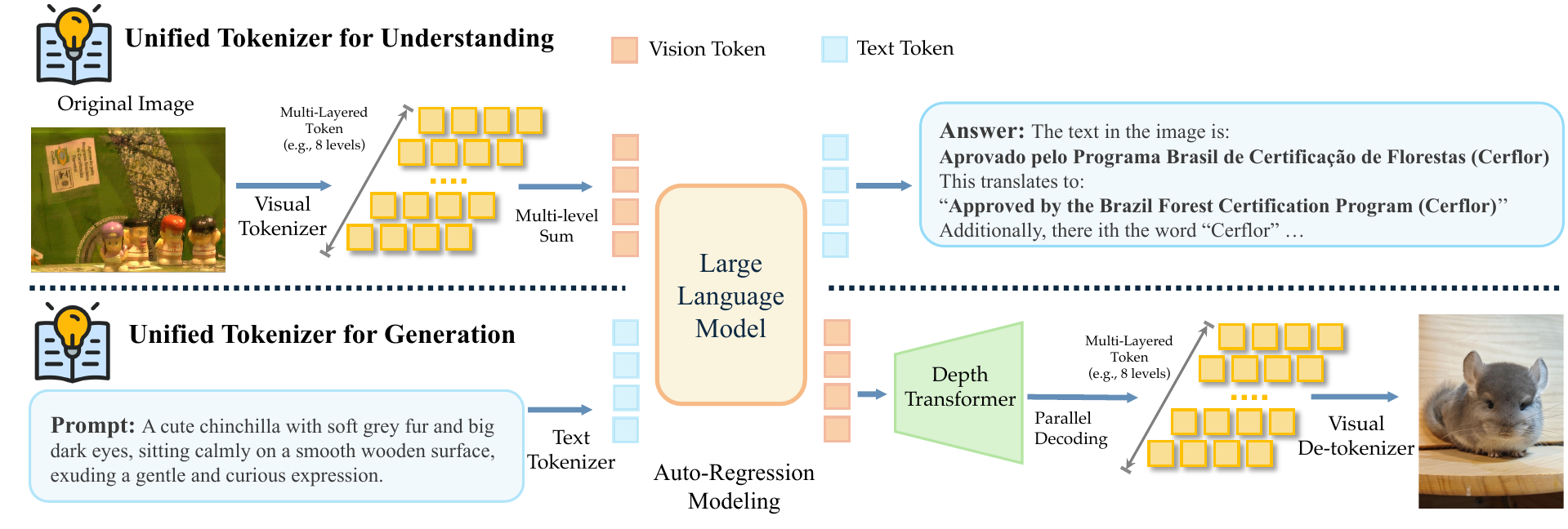}
    \caption{A unified tokenizer and detokenizer for both understanding and generation within the DiNA paradigm.}
    \label{fig:dnavit_umm}
\end{figure}

\vspace{1em} 

\noindent\begin{minipage}{\textwidth}
\subsection{Qualitative Examples}
\vspace{1em}
\centering

\begin{tcolorbox}[
    enhanced, 
    colback=gray!2, 
    colframe=gray!50!black, 
    title=\textbf{Dense Mathematical OCR}, 
    fonttitle=\bfseries,
    attach boxed title to top left={yshift=-2mm, xshift=5mm},
    boxed title style={colback=gray!70!black},
    boxsep=2mm
]

\scriptsize 
\linespread{1.4}\selectfont 

\begin{CJK*}{UTF8}{gbsn}

\begin{minipage}[t]{0.48\textwidth}
    \vspace{0pt} 
    
    \centering
    \includegraphics[width=\textwidth, height=0.45\textheight, keepaspectratio]{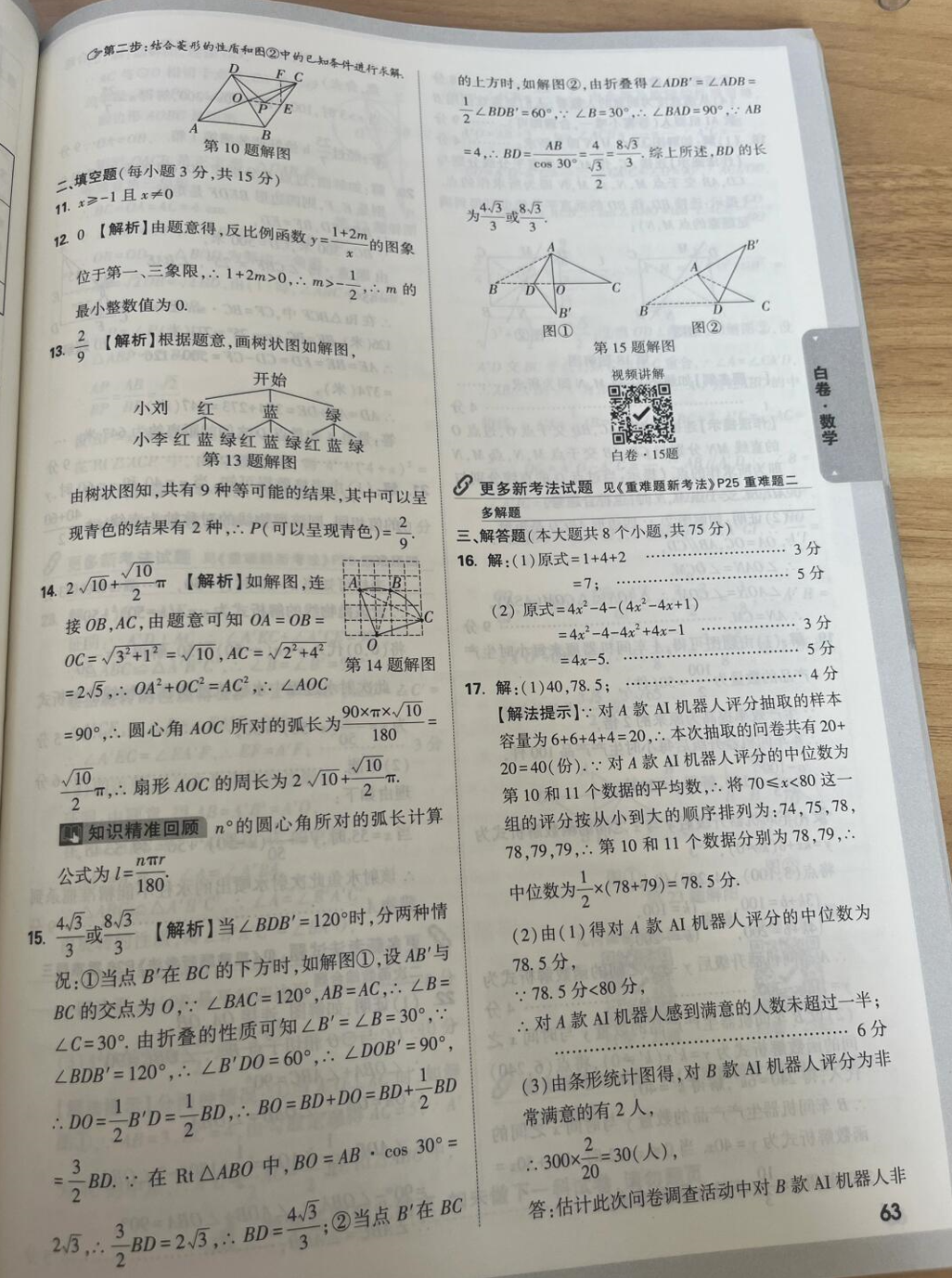}
    
    \vspace{2ex} 
    
    \raggedright 
    
    \hangindent=1.5em
    \hangafter=1
    \textbf{Q: }提取文档图片中正文的所有信息用markdown格式表示，表格用html格式表达，文档中公式用latex格式表示，按照阅读顺序组织进行解析。
    
    \vspace{1.5ex} 
    
    \vspace{1.5ex}
    
    \textbf{A: } \textit{[Model Output]} \\[0.5ex]
    \setlength{\leftskip}{1.5em} 
    
    第二步：结合菱形的性质和图2中的已知条件进行求解 \\[1ex]
    第10题解图 \\[1ex]
    \verb|##| 二、填空题(每小题3分，共15分) \\[0.5ex]
    11. $ x \ge -1 $ 且 $ x \neq 0 $ \\[0.5ex]
    12. 【解析】由题意得，反比例函数 $ y = \frac{1+2m}{x} $ 的图象位于第一、三象限， $ \therefore 1+2m > 0 $, $ \therefore m > -\frac{1}{2} $, $ \therefore m $ 的最小整数值为 0. \\[0.5ex]
    13. $ \frac{2}{9} $ \\
    【解析】根据题意，画树状图如解图， \\
    小李红蓝绿红蓝绿红蓝绿 \\
    第13题解图 \\
    由树状图知，共有9种等可能的结果，其中可以呈现青色（绿色）的结果有2种， $ \because P $ （可以呈现青色）= $ \frac{2}{9} $ .
    
    \vspace{1em} 
\end{minipage}%
\hfill 
\begin{minipage}[t]{0.48\textwidth}
    \vspace{0pt} 
    \raggedright 
    \setlength{\leftskip}{1.5em} 
    
    14. $ 2\sqrt{10}+\frac{\sqrt{10}}{2}\pi $ \\
    【解析】如解图，连接OB，AC，由题意可知 $ OA=OB=OC=\sqrt{3^{2}+1^{2}}=\sqrt{10} $ ， $ AC=\sqrt{2^{2}+4^{2}}=2\sqrt{5} $ ， $ OA^{2}+OC^{2}=AC^{2} $ ， $ \therefore \angle AOC=90^{\circ} $ ， $ \therefore $ 圆心角AOC所对的弧长为 $ \frac{90\times\pi\times\sqrt{10}}{180}=\frac{\sqrt{10}}{2}\pi $ ，扇形AOC的周长为 $ 2\sqrt{10}+\frac{\sqrt{10}}{2}\pi $ . \\
    知识精准回顾 $ n^{\circ} $ 的圆心角所对的弧长计算公式为 $ l = \frac{n\pi r}{180} $ . \\[1.5ex]
    
    15. $ \frac{4\sqrt{3}}{3} $ 或 $ \frac{8\sqrt{3}}{3} $ \\
    【解析】当 $ \angle B D B^{\prime}=120^{\circ} $ 时，分两种情况： \\
    ①当点 $ B^{\prime} $ 在BC的下方时，如解图①，设AB与 $ B C^{\prime} $ 的交点为O， $ \because \angle BAC=120^{\circ} $ ，$ AB=AC $， $ \therefore \angle B=\angle C=30^{\circ} $ . 由折叠的性质可知 $ \angle B^{\prime}=\angle B=30^{\circ} $ ， $ \angle D B B^{\prime}=120^{\circ} $ ， $ \angle B^{\prime} D O=60^{\circ} $ ， $ \angle D O B^{\prime}=90^{\circ} $ ， $ \therefore D O=\frac{1}{2}B^{\prime}D=\frac{1}{2}B D $ ， $ \therefore B O=B D+D O=B D+\frac{1}{2}B D=\frac{3}{2}B D $ . $ \therefore $ 在Rt$\triangle$ABO中，$ BO=AB\cdot\cos 30^{\circ}=\frac{3}{2}BD $ ； \\
    ②当点 $ B^{\prime} $ 在BC上方时，如解图②，由折叠得 $ \angle ADB^{\prime}=\angle ADB=\frac{1}{2}\angle BDB^{\prime}=60^{\circ} $ ， $ \therefore \angle B=30^{\circ} $ ， $ \therefore \angle BAD=90^{\circ} $ ， $ \therefore AB=4 $ ， $ \therefore BD=AB\cos 30^{\circ}=\frac{4}{3}\times\frac{8\sqrt{3}}{3}=\frac{4\sqrt{3}}{3} $ 或 $ \frac{8\sqrt{3}}{3} $ . \\[1.5ex]
    
    第15题解图 \\
    视频讲解 \\
    白卷·数学 \\
    更多新考法试题见《重难题新考法P25重难题二》 \\[1.5ex]
    
    \verb|##| 三、解答题(本大题共8小题，共75分) \\[0.5ex]
    
    16.解:(1)原式=$1+4+2=7$; \\
    (2)原式=$ 4x^{2}-4-(4x^{2}-4x+1)=4x^{2}-4-4x^{2}+4x-1=4x-5 $. \\[1.5ex]
    
    17. 解: (1)40,78.5; \\
    【解题提示】:对A款AI机器人评分抽取的样本容量为$6+6+4+4+20$,本次抽取的问卷共有$20+6+6+4+4+20=40$(份). 对A款AI机器人评分的中位数为20,将$70 \le x < 80$这一组的评分按从小到大的顺序排列为:$74,75,78,78,79,79$,第10和11个数据分别为$78,79$.中位数为$\frac{1}{2}\times(78+79)=78.5$分. \\
    (2)由(1)得对A款AI机器人评分的中位数为78.5分, $ \because 78.5 $分$ < 80 $分, 对A款AI机器人感到满意的人数未超过一半; \\
    (3)由条形统计图得,对B款AI机器人评为非常满意的有2人, $ \therefore 300\times\frac{2}{20}=30 $ (人), \\
    答:估计此次问卷调查活动中对B款AI机器人非。

    63

\end{minipage}

\end{CJK*}
\end{tcolorbox}
\vspace{1em}
\captionof{figure}{Qualitative results of mathematical text and formula recognition. The layout explicitly places the source image and initial text on the left, continuing the long derivation on the right.}
\label{fig:math_ocr_case}

\end{minipage} 

\vspace{1em} 


\begin{figure}[h]
    \centering
    \includegraphics[width=0.9\linewidth]{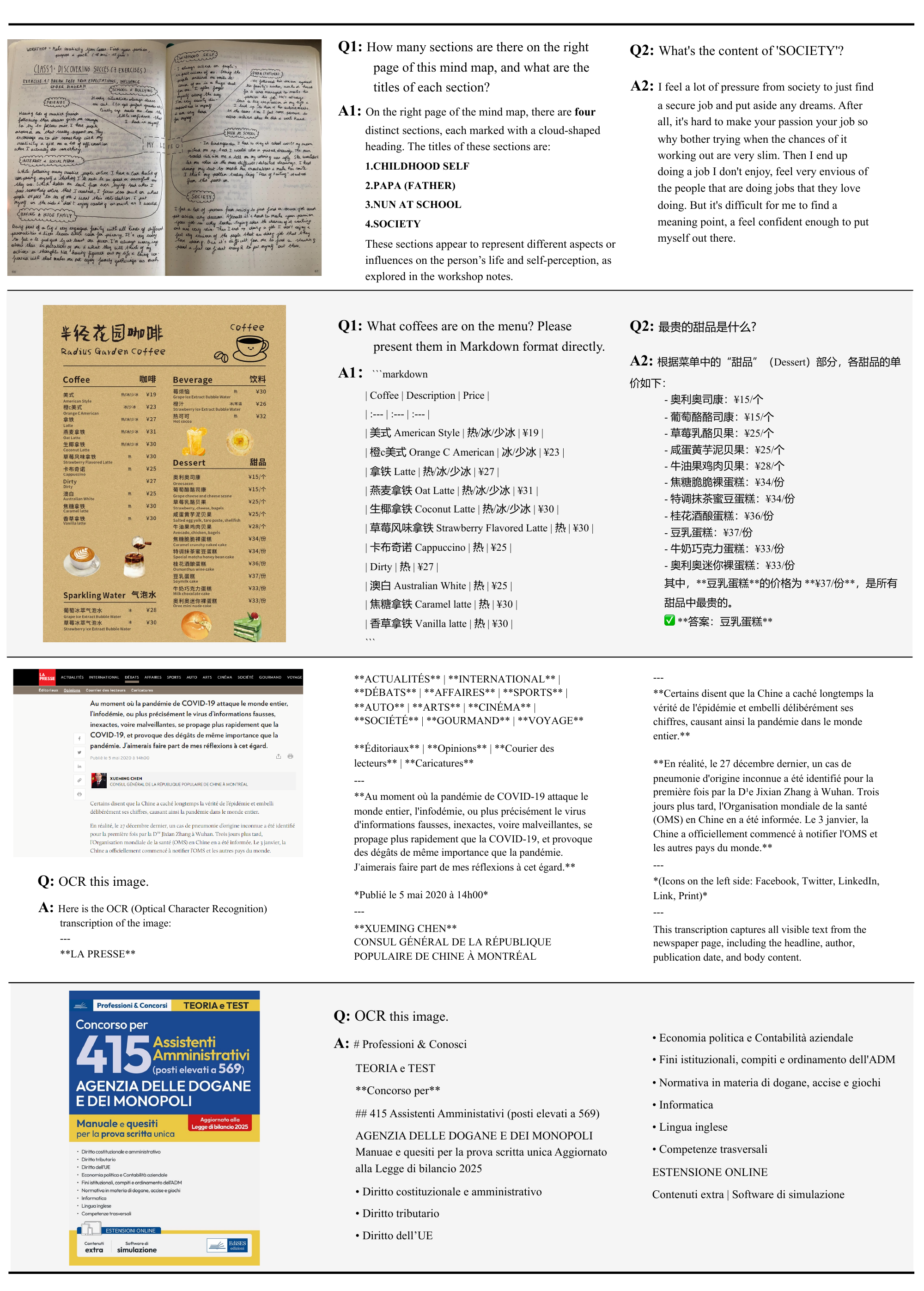}
    \caption{Our model is capable of producing formatted outputs and performing numerical comparisons in the OCR recognition phase, thereby enhancing the accuracy and interpretability of the recognition results.}
\end{figure}

\begin{figure}
    \centering
    \includegraphics[width=0.9\linewidth]{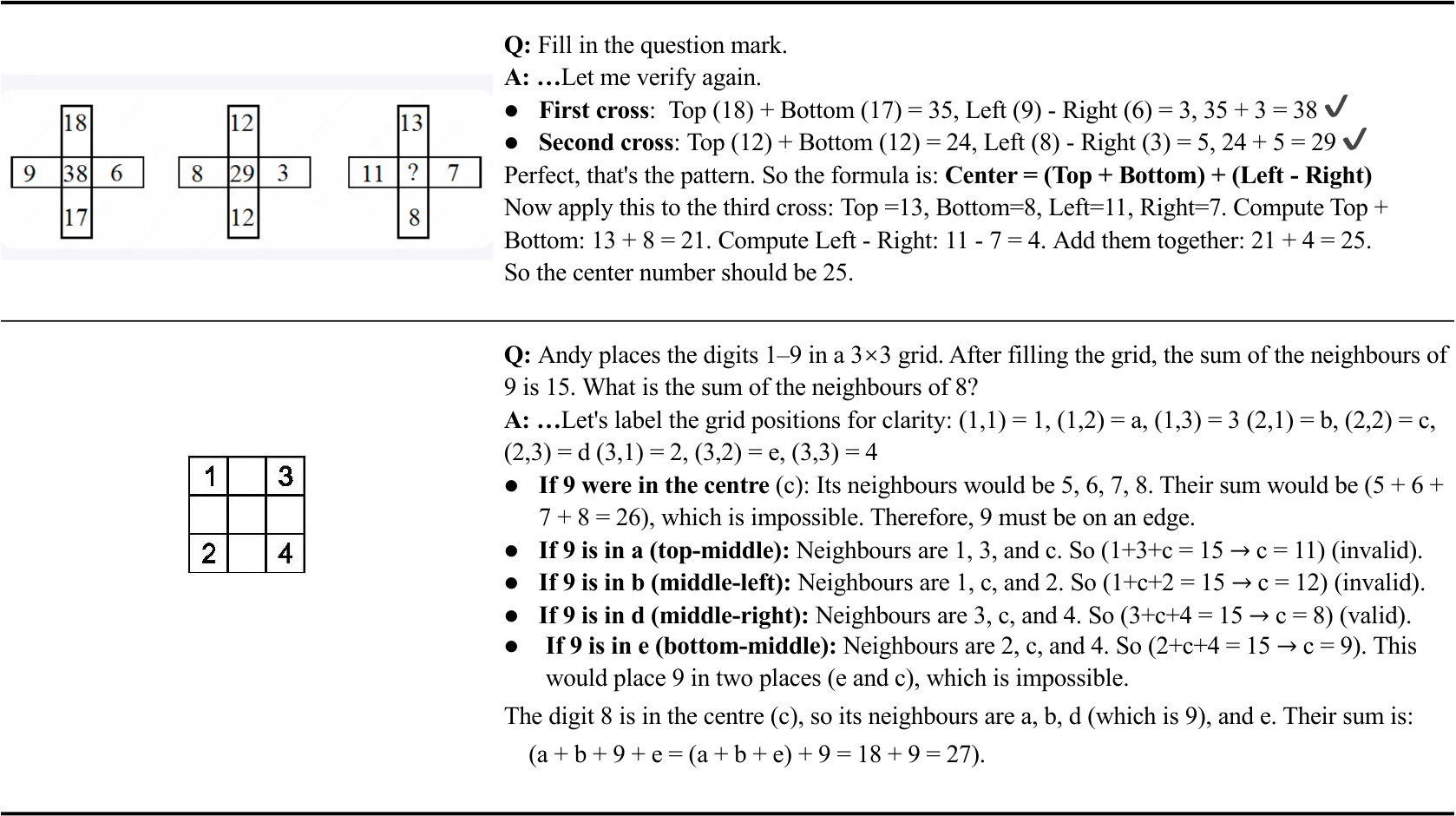}
    \vspace{1em}
    \caption{Two examples of \longcat solving complex mathematical logic puzzles through multi-step reasoning. The top panel showcases the model's ability to identify arithmetic patterns within a cross-shaped visual structure to deduce a missing number. The bottom panel illustrates advanced spatial reasoning and constraint satisfaction, where the model systematically tests different hypotheses to deduce the correct placement of digits in a 3x3 grid and calculate the required sum.}
    \label{fig:stem_case}
\end{figure}

\begin{figure}
    \centering
    \includegraphics[width=0.9\linewidth]{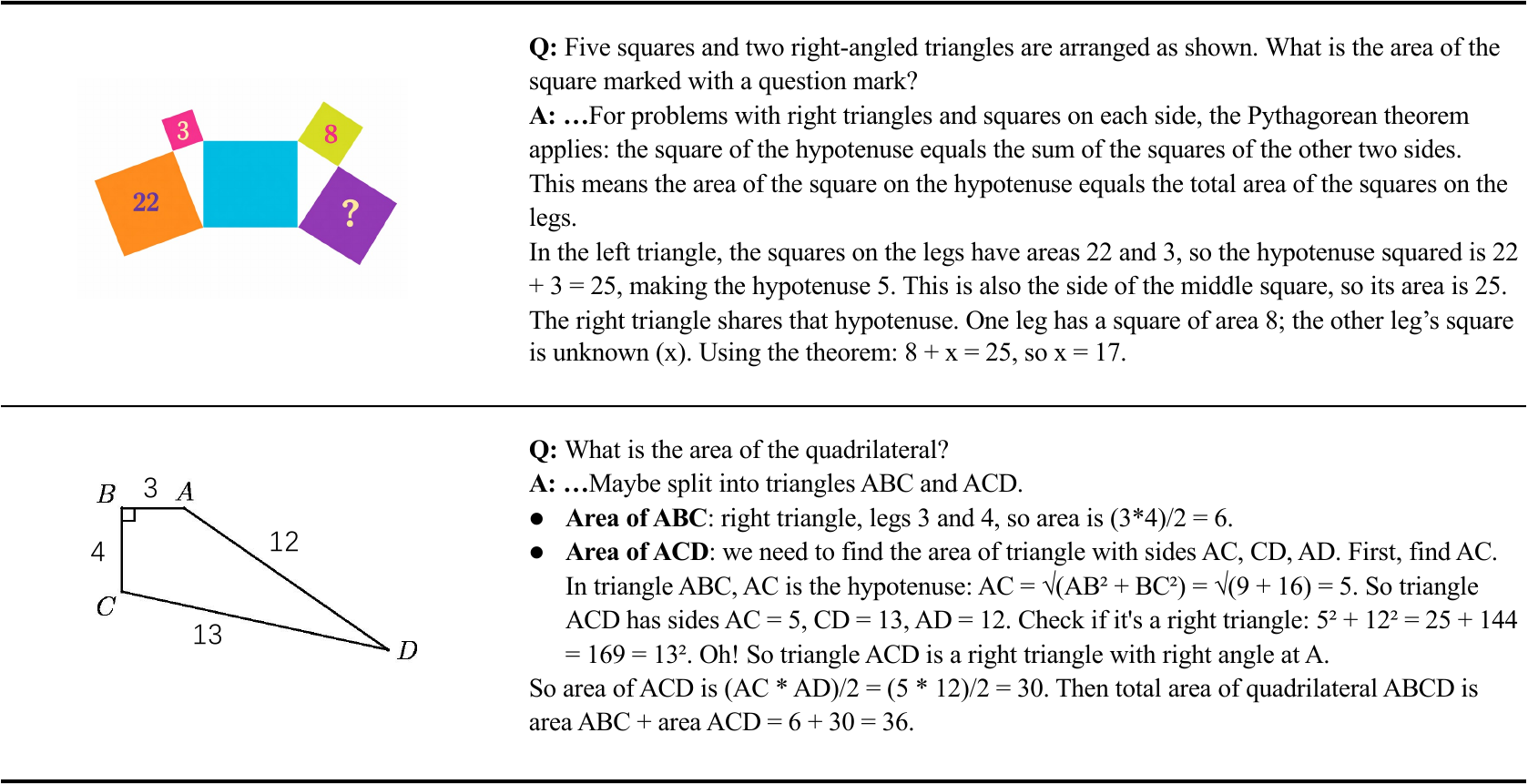}
    \vspace{1em}
    \caption{Two examples of \longcat solving geometric problems. The top panel demonstrates the model's ability to deduce square areas through nested geometric relationships, while the bottom panel highlights its capacity to decompose a complex quadrilateral into triangles and verify right-angle properties to calculate the total area.}
    \label{fig:stem_case_geometry}
\end{figure}

\clearpage
\subsection{The Analysis of Visual De-tokenizer}
\begin{figure}[hbtp]
    \centering
    \includegraphics[width=\linewidth]{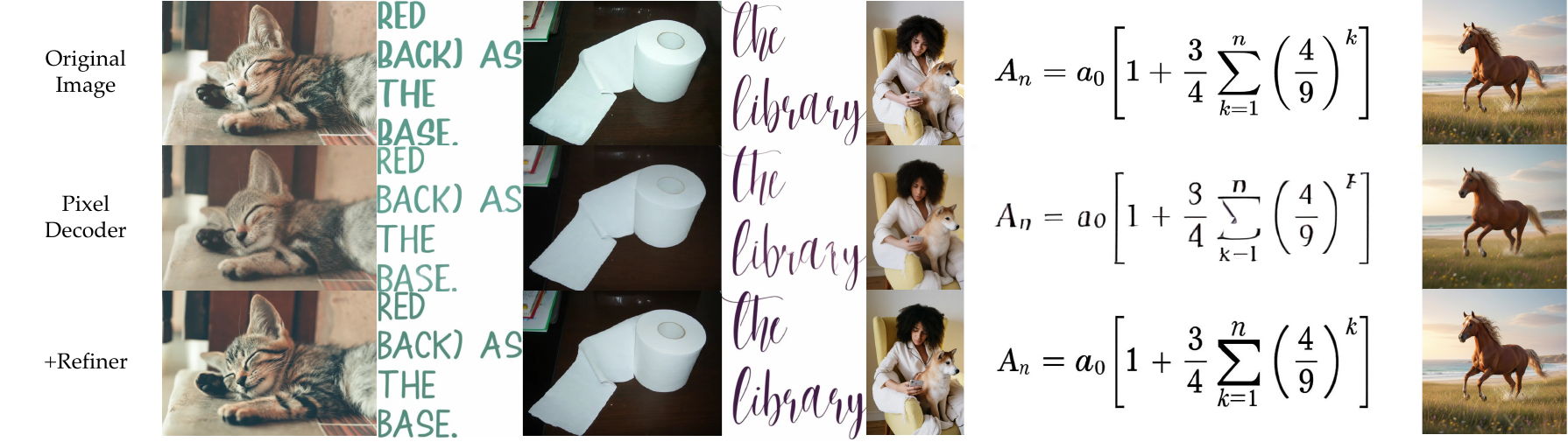}
    \caption{The effect of visual de-tokenizer about the pixel decoder and refiner module.}
    \label{fig:image_rec}
\end{figure}

In the current version, we focus primarily on generation quality, with reconstruction only required to maintain semantic consistency. The de-tokenizer component of our model, which includes the Vision Transformer (ViT) pixel decoder and the diffusion refiner module, collaboratively reconstructs semantic information from images. The visualization is shown in ~\ref{fig:image_rec}. While the ViT decoder alone is capable of reconstructing the semantic content of the image, the frozen SAE encoder limits the capacity to capture high-frequency and fine-grained details. In this context, the refiner, which is designed to focus on detail restoration, plays a critical role in faithfully recovering the original image at the semantic level. Furthermore, within the framework of LLM autoregression, the predicted discrete tokens inherently encode semantic content, such as the layout and structural elements of the image. Notably, these discrete tokens demonstrate superior performance in OCR tasks, as they inherently contain semantically complete information, thereby enabling faithful reconstruction of textual information.

\end{document}